%% file: neurips_2026.tex
\documentclass{article}

\PassOptionsToPackage{numbers,sort&compress}{natbib}
\usepackage[main,final]{neurips_2026}
\usepackage[utf8]{inputenc}
\usepackage[T1]{fontenc}
\usepackage{hyperref}
\hypersetup{hidelinks}
\usepackage{url}
\usepackage{booktabs}
\usepackage{amsfonts}
\usepackage{nicefrac}
\usepackage{microtype}
\usepackage{xcolor}

\usepackage{array}
\usepackage{listings}
\usepackage{setspace}
\usepackage{algorithmic}
\usepackage{pifont}
\usepackage{colortbl}
\usepackage{graphicx} %
\usepackage{amssymb}
\usepackage{fontenc} %

\usepackage{algorithm}
\usepackage{amsthm}
\usepackage{subfig}
\usepackage{colortbl}
\usepackage{amsmath}
\usepackage{mathtools}
\usepackage{multirow}
\usepackage{enumitem}
\usepackage{makecell}
\usepackage{pifont}
\usepackage{wrapfig}
\usepackage{adjustbox}
\usepackage{needspace}
\usepackage{placeins}
\usepackage{float}

\theoremstyle{plain}
\newtheorem{theorem}{Theorem}[section]

\newtheorem{corollary}[theorem]{Corollary}
\theoremstyle{definition}

\theoremstyle{remark}

\title{Sample-Grained Approximate Unlearning with Provable Per-Sample Bounds}

\input{authors}

\begin{document}

\maketitle

\begin{abstract}
	Machine unlearning aims to erase the influence of a designated forget set from a trained model while preserving strong utility on the retain set. Approximate unlearning methods efficiently achieve this goal without full retraining, but their effectiveness is primarily assessed using dataset-level aggregate metrics. These aggregate metrics can mask deviations from retraining at the level of individual samples.
	In this work, we focus on per-sample unlearning effectiveness. We introduce \textbf{SALA} (\textbf{Sa}mple-grained \textbf{L}ikelihoods \textbf{A}Lignment), a plug-in optimization framework that enables approximate unlearning methods to operate at the level of individual samples by steering each example toward its correct target distribution. To quantitatively assess this behavior, we further propose the sample-grained distribution gap, a  metric that captures per-sample discrepancies between approximate unlearning and exact retraining output distributions. We provide theoretical guarantees showing that SALA contracts this gap and yields a bounded per-sample discrepancy to retraining. To make this framework practical, we develop a low-cost estimation method for the member and non-member target distributions. Extensive experiments on classification and generation tasks demonstrate that SALA reduces the sample-grained distribution gap by up to $86.67\%$ when paired with existing approximate unlearning baselines, while also improving robustness to unlearning hyperparameters.
\end{abstract}

\input{1_intro}
\input{2_problem_definition}
\input{3_methodology}
\input{4_sample_logits_estimation}
\input{5_experiments}
\input{6_related_works}

\vspace{-0.285cm}
\section{Conclusion}
\label{sec:conclusion}
\vspace{-0.31cm}
Approximate unlearning can reduce dataset-level gap to retraining, but our experiments show that small dataset-level discrepancy can still hide large sample-level deviations. We address this issue with SALA, a plug-in framework that uses per-sample likelihood alignment to steer each sample toward its role-dependent target distribution. Under the InDist criterion, SALA gives a computable upper bound on the sample-grained distribution gap measured by DNO. We further introduce SMSE, a low-overhead estimator for the per-sample member and non-member target distributions required by SALA. Across classification and generation tasks, SALA reduces DNO by up to $86.67\%$ and improves robustness to unlearning hyperparameters.

\input{acknowledgments}
\bibliographystyle{plainnat}
\bibliography{reference}

\newpage
\appendix
\input{appendix}

\newpage
\input{checklist.tex}
\end{document}

%% file: authors.tex
\let\AuthorOriginalMaketitle\maketitle
\renewcommand{\maketitle}{%
  \begingroup
  \setlength{\tabcolsep}{1.5pt}%
  \AuthorOriginalMaketitle
  \endgroup
}
\author{%
  Yaxin Xiao \\
  The Hong Kong \\
  Polytechnic University \\
  \texttt{samin-yaxin.xiao@polyu.edu.hk}
  \And
  Yulin Jin \\
  The Hong Kong \\
  Polytechnic University \\
  \texttt{yulin.jin@connect.polyu.hk}
  \And
  Li Tang\thanks{Corresponding author: \texttt{tangli@ynu.edu.cn}.} \\
  Yunnan University \\
  \strut \\
  \texttt{tangli@ynu.edu.cn}
  \AND
  \makebox[0.45\textwidth][c]{Haoyang LI} \\
  The Hong Kong Polytechnic University \\
  \texttt{hao-yang9905.li@connect.polyu.hk}
  \And
  \makebox[0.45\textwidth][c]{Huadi Zheng} \\
  Huawei Technologies Co., Ltd. \\
  \texttt{zhenghuadi@huawei.com}
  \AND
  \makebox[0.45\textwidth][c]{Qingqing Ye} \\
  The Hong Kong Polytechnic University \\
  \texttt{qqing.ye@polyu.edu.hk}
  \And
  \makebox[0.45\textwidth][c]{Haibo Hu} \\
  The Hong Kong Polytechnic University \\
  Research Centre for Privacy and Security \\
  Technologies in Future Smart Systems \\
  \texttt{haibo.hu@polyu.edu.hk}
}

%% file: 1_intro.tex
\vspace{-0.24cm}
\section{Introduction}
\vspace{-0.29cm}
\label{sec:intro}
Motivated by the legal data-deletion requirement known as the ``right to be forgotten''~\cite{gdpr2016}, machine unlearning is introduced to remove the influence of a designated forget set from a trained model~\cite{cao2015forget,ginart2019making,bourtoule2021machine} while preserving utility on the retain set. The most direct way to achieve this goal is to retrain the model from scratch using only the retain set. This exact retraining procedure is the gold standard for unlearning. However, because exact retraining is often too expensive for deployment, recent work has increasingly shifted to approximate unlearning (AU)~\cite{golatkar2020eternal,1modelsparisty23jia}, which updates a pretrained model with small compute overhead to closely match the result of exact retraining. Accordingly, the efficacy of AU is commonly evaluated by gap-to-retrain~\cite{cadet2024deep}, which measures how close the AU model distribution is to the exact-retraining (ER) distribution under a chosen observable metric. 

\begin{wrapfigure}{r}{0.51\columnwidth}
	\centering
	\includegraphics[width=0.235\columnwidth]{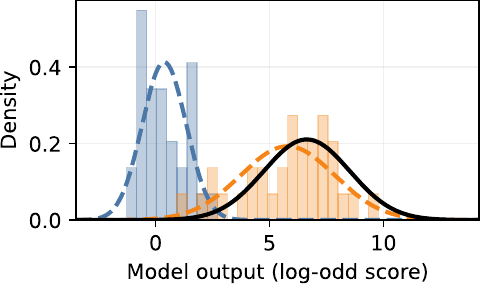}
	\includegraphics[width=0.235\columnwidth]{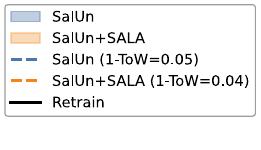}
	\vspace{-0.18cm}
	\caption{Output log-odds distributions for a selected forgotten sample. We compare retraining with AU (\emph{i.e.}, SalUn~\cite{fan2024salun}) with and without SALA under the score in Eq.~\eqref{eq:logodds}. SALA moves the distribution closer to the retraining reference.}
	\label{fig:per_sample_gap}
	\vspace{-0.168cm}
\end{wrapfigure}

\begin{wrapfigure}{r}{0.51\columnwidth}
	\centering
	\vspace{-0.65cm}
	\subfloat[DNO vs ToW]{
		\label{fig:tow_vs_dno}
		\includegraphics[width=0.235\columnwidth]{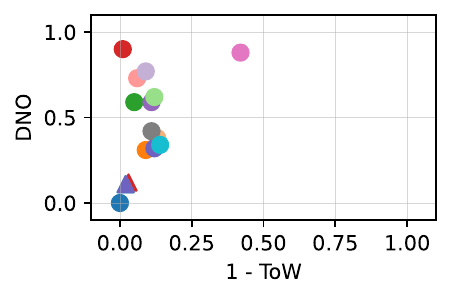}}
	\vspace{0.03cm}
	\subfloat[DNO vs RULI]{
		\label{fig:ruli_vs_dno}
		\includegraphics[width=0.235\columnwidth]{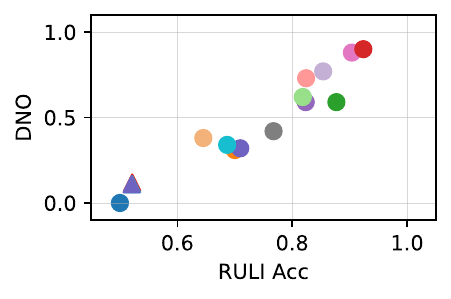}}\\
	\vspace{-0.07cm}
	\includegraphics[width=0.505\columnwidth]{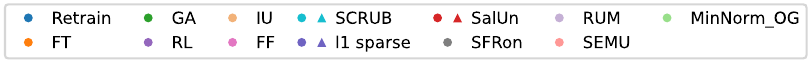}
	\vspace{-0.5cm}
	\caption{Gap-to-retrain evaluations for AU. ToW measures comprehensive dataset-average gap (see \eqref{eq:tow}). Triangles denote AU results assisted by SALA.}
	\label{fig:dno_tow_ruli}
	\vspace{-0.2cm}
\end{wrapfigure}

However, existing works usually rely on dataset-level average observables, such as prediction accuracy over a dataset, which can mask unlearning failures at the sample level.
To capture per-sample unlearning performance, we measure distribution non-overlap (DNO\footnote{If DNO is large for $x$, approximate unlearning (AU) produces outputs that differ from exact retraining (ER) on $x$, and $x$ is not in the desired unlearned state, even when dataset-level averages appear good.}), defined as the discrepancy between the output distributions estimated over multiple runs of unlearning a given sample $x$ and exact retraining without $x$, as illustrated in Figure~\ref{fig:per_sample_gap}. Figure~\ref{fig:tow_vs_dno} shows that state-of-the-art AU methods can achieve a small dataset-level gap-to-retrain, measured by $1-\mathrm{ToW}$\footnote{A smaller $1-\mathrm{ToW}$ indicates a smaller dataset-level gap to retrain.}, while still exhibiting a large per-sample gap-to-retrain, \emph{i.e.,} DNO. Figure~\ref{fig:ruli_vs_dno} further shows that this per-sample gap is associated with greater privacy leakage on forget samples, as evaluated here by the representative attack RULI~\cite{naderloui2025rectifying}.

\vspace{-0.25cm}
These findings highlight a key research gap: \textbf{sample-grained approximate unlearning remains insufficiently studied and far from well achieved}. We address this gap with \textbf{SALA} (\textbf{Sa}mple-grained \textbf{L}ikelihoods \textbf{A}lignment), a plug-in framework that equips AU with explicit per-sample targets. SALA assigns each sample a role-specific target distribution: retain samples align to $Q_{\mathrm{in}}(\cdot\mid x)$, which represents member behavior, while forget samples align to $Q_{\mathrm{out}}(\cdot\mid x)$, which represents non-member behavior. For each sample $x$, SALA minimizes the sample's $z$-score to its assigned target distribution, driving the output toward the prescribed distribution.

Beyond empirically reducing the per-sample gap-to-retrain of AU methods, SALA provides a provable upper bound on this gap measured by DNO. Moreover, it reduces the strong hyperparameter sensitivity of existing AU methods, which are often tuned heuristically and can degrade sharply under poor configurations. SALA mitigates this issue through its $z$-score, which serves as an explicit per-sample alignment signal. From this signal, we define \textbf{InDist}, the fraction of samples whose outputs lie within a preset $z$-band around their assigned targets. InDist supports two complementary automated mechanisms: \textbf{hyperparameter selection} and a \textbf{stop criterion}. For the former, SALA runs a fast probe with only a few unlearning steps and chooses the configuration that maximizes InDist. For the latter, SALA stops once InDist exceeds a high threshold, such as $0.95$. This improves AU robustness to hyperparameters and avoids unnecessary updates. Finally, since InDist imposes an explicit distributional constraint, it also yields a theoretical upper bound on SALA's DNO.

However, constraining InDist in SALA requires access to the full original training set, which may be unrealistic in some settings. We therefore introduce a forget-target-only variant that aligns only forget samples to $Q_{\mathrm{out}}(\cdot\mid x)$, for AU methods with limited access to the retain set. When the underlying AU method satisfies the utility preservation condition in Eq.~\ref{eq:retain_consistency_loss}, this variant still yields an upper bound on DNO under a single InDist constraint on forget samples.

The main bottleneck of SALA is estimating $Q_{\mathrm{in}}(\cdot\mid x)$ and $Q_{\mathrm{out}}(\cdot\mid x)$ for each training sample $x$. These targets can be estimated once during model development and reused across unlearning requests. One direct solution uses shadow-model estimation~\cite{feldman2020neural}, which trains models under different forget-retain splits and aggregates their outputs. This pipeline resembles the memorization-estimation step used by AU methods such as SUM~\cite{zhao2024what}. If compatible member and non-member score statistics already exist, SALA can reuse them for target estimation. However, shadow-model estimation is expensive and costs tens of standard training runs. To avoid this overhead, \textbf{we propose an efficient estimator that obtains both targets and the model in a single training run}. It uses a split--merge training scheme. In the middle stage, two models train on complementary subsets, so each sample is in-training for one model and out-of-training for the other. In the late stage, Federated Averaging merges the two models~\cite{kairouz2021advances}. Finally, checkpoint ensembles collected during training then estimate $Q_{\mathrm{in}}(\cdot\mid x)$ and $Q_{\mathrm{out}}(\cdot\mid x)$.

In summary, this work makes the following contributions:
\begin{itemize}
	\item We propose a sample-grained approximate unlearning (AU) framework. It monitors and guides alignment to retraining at the individual-sample level.
	\item SALA provides a theoretical upper bound on the sample-grained gap-to-retrain metric (\emph{i.e.}, DNO), yielding guarantees for AU at the sample level.
	\item We introduce SMSE, which estimates per-sample member and non-member distributions with split--merge training and segment ensembles. SMSE reduces preparation from multiple shadow-model trainings to a single model training run.
	\item We evaluate SALA on classification and generative tasks with diverse AU methods. Figure~\ref{fig:dno_tow_ruli} shows that SALA reduces DNO by up to $86.67\%$, keeps a small dataset-level gap to retraining (\emph{i.e.}, 1-ToW), and remains robust without manual hyperparameter tuning.
\end{itemize}

%% file: 2_problem_definition.tex
\vspace{-0.25cm}
\section{Problem Definition}
\label{sec:problem_definition}
\vspace{-0.2cm}
\subsection{Machine Unlearning Notations}
\vspace{-0.22cm}
\label{subsec:pd_notations}
Machine unlearning studies how to update a trained model so that the influence of a designated subset of its training data is removed. Let $f_\theta$ denote a model with parameters $\theta$, trained on a dataset $D\sim \mathcal{D}$. Upon a deletion request supported by the \emph{right to be forgotten} (Article 17 of the GDPR~\cite{gdpr2016}), a forget set $D_\mathrm{f} \subseteq D$ is specified and the remaining retain set is denoted as $D_\mathrm{r} = D \setminus D_\mathrm{f}$. Let $\mathcal{A}: D \rightarrow \mathcal{R}$ be a learning algorithm that maps a dataset to a model in the output space $\mathcal{R}$ (\emph{e.g.}, neural network parameters), and let $\mathcal{U}$ be an approximate unlearning (AU) algorithm. The retrained model is defined as $f_{\theta^*}$, with $\theta^*=\mathcal{A}(D_\mathrm{r})$. The unlearned model is $f_{\theta_u} = \mathcal{U}\bigl(\mathcal{A}(D),\, D_\mathrm{f},\, D\bigr)$. AU aims to make $\theta_u$ behave similarly to $\theta^*$ while being substantially more efficient than rerunning $\mathcal{A}$ from scratch on $D_\mathrm{r}$. 
\vspace{-0.25cm}
\subsection{Evaluation for Approximate Unlearning (AU)}
\vspace{-0.2cm}
\label{subsec:pd_evaluation}
We evaluate approximate unlearning (AU) along four dimensions~\cite{li2025mubox,cadet2024deep}: (i) \textbf{efficacy}, \emph{i.e.}, how close the unlearned model is to the exact-retraining result and how much utility it keeps; (ii) \textbf{privacy}, \emph{i.e.}, whether deleted samples can be detected by inference attacks; and (iii) \textbf{efficiency}, \emph{i.e.}, the compute cost of unlearning, commonly summarized by run-time. In addition, we propose (iv) \textbf{robustness}, which captures AU sensitivity to hyperparameters and run-to-run variance.
\vspace{-0.15cm}
\subsubsection{Efficacy Metrics for AU}
\vspace{-0.2cm}
Efficacy is assessed at two levels: dataset-level and sample-grained. Most existing metrics are dataset-level and quantify the gap-to-retrain under a chosen observable. For an AU unlearned model $f_{\theta_u}$, smaller gaps to the retrained model $f_{\theta^*}$ indicate higher unlearning efficacy.

\noindent\textbf{Dataset-level metrics.}
This gap is measured between $f_{\theta_u}$ and $f_{\theta^*}$ on three datasets: test ${D}_\mathrm{t}$, retain ${D}_\mathrm{r}$, and forget ${D}_\mathrm{f}$. For classification, the metrics are accuracy: test accuracy (\textbf{TA}), retain accuracy (\textbf{RA}), and forget accuracy (\textbf{FA}). For text generative tasks, the metrics are text perplexity (PPL): \textbf{T-PPL}, \textbf{R-PPL}, and \textbf{F-PPL}. These gaps are aggregated by the Tug-of-War (\textbf{ToW}) score~\cite{zhao2024what}:
    \vspace{-0.07cm}
\begin{equation}
	\footnotesize
	\label{eq:tow}
	\text{ToW}(M_1, M_2, M_3) = \prod\nolimits_{i=1}^3\left(1 - \frac{d(M_i(\theta_u), M_i(\theta^*))}{M_i(\theta^*)}\right),
\end{equation}
where $M_i(\theta)$ is a metric value of model $f_\theta$ (e.g., TA/RA/FA or T-PPL/R-PPL/F-PPL), and $d(\cdot,\cdot)$ is the absolute difference between two metric values.

\noindent\textbf{Sample-grained metrics reveal finer gaps.}
Sample-grained metrics compare AU with exact retraining at the level of individual samples. For a fixed sample $x$, they measure the discrepancy between the distribution of an observable score $s$ induced by the AU model output and the corresponding distribution under exact retraining. This paper introduces Distribution Non-Overlap (\textbf{DNO}) to quantify individual-sample distributional deviations of AU from retraining under both the retain and forget roles. Let $P_{\mathrm{r}}(s \mid x)$ and $P_{\mathrm{f}}(s \mid x)$ denote the estimated score distributions produced by AU when $x$ is treated as a retain sample (member) and a forget sample (non-member), respectively. $P_{\mathrm{r}}(s \mid x)$ and $P_{\mathrm{f}}(s \mid x)$ are estimated over $N$ independent runs in which $x$ repeatedly takes the corresponding role. Let $Q_{\mathrm{in}}(s \mid x)$ and $Q_{\mathrm{out}}(s \mid x)$ be their exact-retraining counterparts. For reporting, DNO summarizes these individual-sample discrepancies as 
\vspace{-0.2cm}
\begin{equation}
	\footnotesize
	\label{eq:dno}
	\mathrm{DNO}
	=\mathbb{E}_{x}\frac{1}{2}\Bigg[
	\left(1-\int \min\!\bigl(P_{\mathrm{r}}(s\mid x),\,Q_{\mathrm{in}}(s\mid x)\bigr)\,ds\right)
	+\left(1-\int \min\!\bigl(P_{\mathrm{f}}(s\mid x),\,Q_{\mathrm{out}}(s\mid x)\bigr)\,ds\right)
	\Bigg].
\end{equation}

\vspace{-0.29cm}
Smaller DNO indicates smaller sample-level distributional discrepancies from retraining on average. 
\vspace{-0.2cm}
\subsubsection{Privacy Metrics for AU}
\vspace{-0.22cm}
This paper considers two AU privacy evaluation cases that differ in the attacker’s knowledge of the unlearning procedure: a legacy attacker that ignores unlearning, and an unlearning-aware attacker that knows the unlearning algorithm~\cite{hayes2025inexact}. Both cases are evaluated via membership inference attack (MIA)~\cite{sula2024silver}, which takes forget samples as positives and held-out test samples as negatives. The legacy setting is reported as \textbf{MIA efficacy}~\cite{1modelsparisty23jia}. The unlearning-aware setting is instantiated with \textbf{RULI}~\cite{naderloui2025rectifying}, a recent unlearning-specific MIA evaluation framework. For both settings, we use the best MIA accuracy as the primary metric. 
In both cases, the attacker estimates score densities from shadow models and applies the same likelihood-ratio rule; the hypotheses $(H_1,H_0)$ change with attacker knowledge. Specifically, the legacy case compares the member state ($x\in D$) against the non-member state ($x\notin D$), as in LiRA~\cite{carlini2022membership}, while the unlearning-aware case compares the trained-then-unlearned state ($x\in D_\mathrm{f}$) against the never-trained state ($x\notin D$), using scores computed on the unlearned model, as in RULI~\cite{naderloui2025rectifying}. Concretely,
\vspace{-0.1cm}
\begin{equation}
\footnotesize
	\Lambda(x)=\frac{p\bigl(s(x)\mid H_1\bigr)}{p\bigl(s(x)\mid H_0\bigr)},
	\text{predict member if }\Lambda(x)>\tau_\mathrm{MI}.
\end{equation}

\vspace{-0.15cm}
For the legacy case, $(H_1,H_0)=(x\in D,\;x\notin D)$; for the unlearning-aware case, $(H_1,H_0)=(x\in D_f,\;x\notin D)$ with score $s(x)$ computed on the unlearned model.

\vspace{-0.20cm}
\subsubsection{Robustness Criteria}
\vspace{-0.22cm}
Robustness measures sensitivity to hyperparameter choices. Let $\mathcal{K}\triangleq\{\mathrm{arg}_i\}_{i=1}^{n}$ denote the set of evaluated hyperparameter settings. For an efficacy metric $M$ (\emph{e.g.}, ToW or DNO), robustness is measured by the worst--best \textbf{range} (\emph{i.e.}, fluctuation) over $\mathcal{K}$:
\vspace{-0.02cm}
\begin{equation}
	\footnotesize
	\mathrm{Range}(M)
	=\max\nolimits_{\mathrm{arg}\in\mathcal{K}} M_{\mathrm{arg}}-\min\nolimits_{\mathrm{arg}\in\mathcal{K}} M_{\mathrm{arg}}.
\end{equation}

\vspace{-0.2cm}
Smaller $\mathrm{Range}(M)$ indicates higher robustness.

%% file: 3_methodology.tex
\vspace{-0.24cm}
\section{Sample-Grained Unlearning Framework}
\begin{figure*}
\centering
\includegraphics[width=1.0\columnwidth]{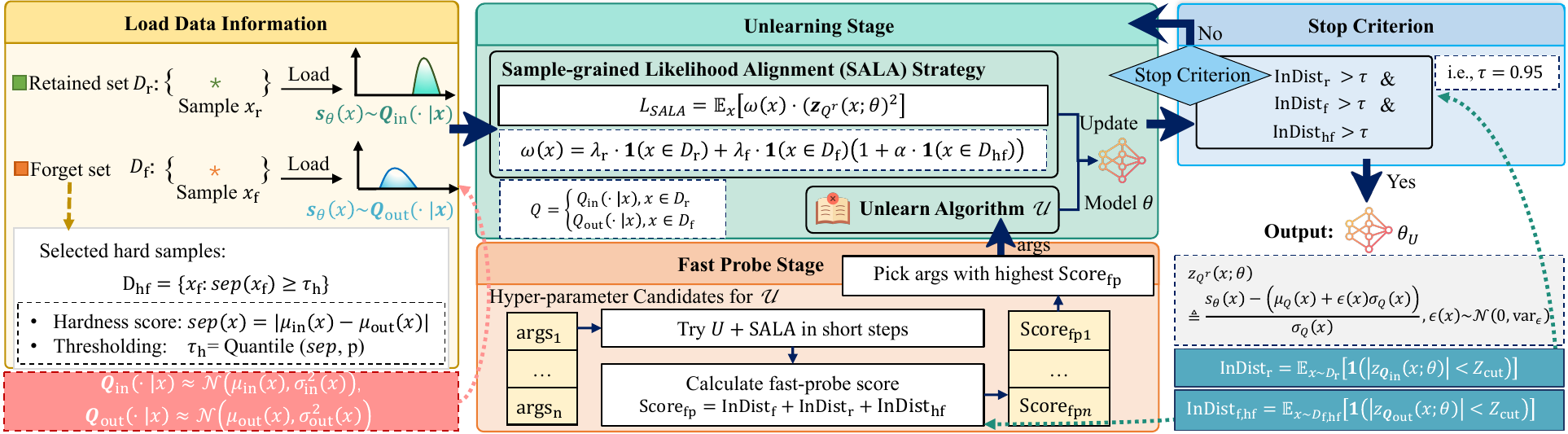}
	\vspace{-0.57cm}
\caption{Overview of SALA. SALA augments a base unlearning $\mathcal{U}$ with sample-wise distribution-alignment guidance losses, fast-probe hyperparameter selection, and sample-level stop criterion.}
	\vspace{-0.42cm}
\label{fig:sala_framework}
\end{figure*}

\vspace{-0.29cm}
\label{sec:sala_framework}
This section introduces sample-grained likelihood alignment (SALA), an auxiliary framework that improves approximate unlearning (AU) by explicitly aligning the predicted likelihood of each sample with its target distribution, as shown in Figure~\ref{fig:sala_framework}. For a forget sample $x \in D_\mathrm{f}$, SALA aligns the prediction with the non-member distribution $Q_{\mathrm{out}}(\cdot\mid x)$. For a retain sample $x \in D_\mathrm{r}$, SALA aligns the prediction with the member distribution $Q_{\mathrm{in}}(\cdot\mid x)$. SALA measures this alignment with a per-sample $z$-score, which both defines the guidance loss and serves as a sample-level stopping criterion (Section~\ref{subsec:sala_loss}). When a sufficiently large fraction of samples satisfy this criterion, SALA bounds the per-sample gap to retraining, as measured by DNO (Section~\ref{subsec:theory}). We also provide a related SALA variant that constrains only forget samples and admits a DNO upper bound under additional conditions on the base unlearning algorithm (Section~\ref{subsec:single_side_sala}).

\noindent\textbf{SALA Framework.}
Figure~\ref{fig:sala_framework} shows the four-step SALA workflow. First, SALA prepares per-sample reference statistics for $Q_{\mathrm{in}}(\cdot\mid x)$ and $Q_{\mathrm{out}}(\cdot\mid x)$. It also identifies a hard-forget subset $D_{\mathrm{hf}}\subseteq D_\mathrm{f}$ whose member and non-member distributions are more separable and thus harder to erase. Second, a fast probe screens candidate hyperparameters $\mathcal{K}\triangleq\{\mathrm{arg}_i\}_{i=1}^{n}$ with a few unlearning steps. The probe selects the candidate that best satisfies InDist (Eq.~\ref{eq:indist}), which measures the fraction of samples whose outputs fall within the target $z$-band. Third, SALA uses the selected $\mathrm{arg}$ in the full unlearning stage and augments the base update rule $\mathcal{U}$ with a per-sample guidance loss. Finally, SALA monitors full-data InDist online and returns $\theta_u$ once InDist exceeds a preset threshold. Appendix~\ref{app:algorithm} provides the pseudo-code.
	\vspace{-0.1cm}
\subsection{Definitions}
	\vspace{-0.1cm}
\noindent\textbf{Score $s_\theta(x)$.} SALA uses an output score $s_\theta(x)$ for distribution alignment, abbreviated as $s(x)$ or $s$ when clear from context. For classification with logits $g_\theta(x)\in\mathbb{R}^C$ and label $y$, we define
$g_\theta(x)\in\mathbb{R}^C$ and label $y$, define
\vspace{-0.05cm}
\begin{equation}
	\label{eq:logodds}
	\footnotesize
	s_\theta(x)
	\triangleq \log\frac{p_\theta(y\mid x)}{1-p_\theta(y\mid x)},
\end{equation}
where $p_\theta(y\mid x)=\mathrm{softmax}(g_\theta(x))_y$ is the predicted probability of the true class. The log-odds score is unbounded and retains distinctions when the probability is close to 0 or 1. Prior empirical evidence~\cite{carlini2022membership} shows that the score distribution for a fixed sample across runs is approximately Gaussian, motivating Gaussian modeling of the per-sample reference scores. For generation, we define the score as the per-token cross-entropy loss. We further validate this Gaussian approximation experimentally in Appendix~\ref{app:rebuttal-gaussian}.

\noindent\textbf{Per-sample reference distributions.} 
Following prior empirical findings~\cite{carlini2022membership}, we define the member and non-member target output distributions for each sample $x$ as
\begin{equation}
	\footnotesize
	\label{eq:q_in_out}
	Q_{\mathrm{in}}(\cdot\mid x)\approx \mathcal{N}\bigl(\mu_{\mathrm{in}}(x),\,\sigma^2_{\mathrm{in}}(x)\bigr),\qquad
	Q_{\mathrm{out}}(\cdot\mid x)\approx \mathcal{N}\bigl(\mu_{\mathrm{out}}(x),\,\sigma^2_{\mathrm{out}}(x)\bigr),
\end{equation}

\vspace{-0.1cm}
\noindent\textbf{$z$-score with shifted mean for distribution match.}
Let $\mu_Q(x)$ and $\sigma_Q(x)$ denote the mean and standard deviation of the target distribution for $x$, where $Q=Q_{\mathrm{in}}$ for retain samples and $Q=Q_{\mathrm{out}}$ for forget samples. To avoid the degenerate solution where $s_\theta(x)$ matches $\mu_Q(x)$ and the effective variance collapses, for each unlearning run, we compute the $z$-score with a shifted mean:
\begin{equation}
	\label{eq:zscore}
	\footnotesize
	z_{Q^r}(x;\theta)\triangleq \frac{s_\theta(x)-\mu^r_Q(x)}{\sigma_Q(x)},\qquad
	\mu^r_Q(x)\triangleq \mu_Q(x)+\epsilon(x)\,\sigma_Q(x),
\end{equation}

\vspace{-0.25cm}
where $\epsilon(x)\sim\mathcal{N}(0,\mathrm{var}_\epsilon)$ is a fixed random offset sampled once per $x$ for each run. Small $\bigl|z_{Q^r}(x;\theta)\bigr|$ means the current output $s_\theta(x)$ lies in a high-density region of the (regularized) target.

\noindent\textbf{InDist metric.}
To explicitly monitor distribution alignment at the sample level, we define InDist as the fraction of samples whose outputs lie within a fixed $z$-band:
\vspace{-0.02cm}
\begin{equation}
	\label{eq:indist}
	\footnotesize
	\mathrm{InDist}(S;\theta)
	=\mathbb{E}_{x\sim S}\left[\mathbf{1}\bigl(|z_{Q^r}(x;\theta)|<Z_{\mathrm{cut}}\bigr)\right],
\end{equation}

\vspace{-0.21cm}
where $S\in\{D_\mathrm{r},\;D_\mathrm{f},\;D_{\mathrm{hf}}\}$ and $Z_{\mathrm{cut}}$ is a fixed cutoff. InDist takes values in $[0,1]$, and a larger value indicates that more samples fall within their target distributions.

\vspace{-0.2cm}
\subsection{SALA Hyper-parameter Fast Probe, Unlearning Guidance, and Stop Criterion}
\label{subsec:sala_loss}
	\vspace{-0.2cm}
\noindent\textbf{Hyper-parameter fast probe.}
Each candidate $\mathrm{arg}\in\mathcal{K}$ is evaluated by running only a few unlearning steps from the same pre-unlearning state and scoring coverage, \emph{e.g.}, 
$\mathrm{InDist}(D_\mathrm{r};\theta)+\mathrm{InDist}(D_\mathrm{f};\theta)+\mathrm{InDist}(D_{\mathrm{hf}};\theta)$. 
The best candidate with the highest score is then used for the unlearning, while all probe updates are discarded.

\noindent\textbf{SALA guidance in unlearning.}
SALA augments the unlearning objective with the following sample-wise distribution-alignment loss:
\vspace{-0.135cm}
\begin{equation}
	\label{eq:Lsala}
	\footnotesize
	L_{\mathrm{SALA}}(\theta)
	=\mathbb{E}_{x}\!\left[\omega(x)\,z_{Q^r}(x;\theta)^2\right].
\end{equation}

\vspace{-0.2cm}
The $z_{Q^r}(x;\theta)^2$ term pulls outputs toward the target shifted mean. Sample weights are
\begin{equation}
	\label{eq:omega}
	\footnotesize
	\omega(x)=\lambda_\mathrm{r}\mathbf{1}[x\in D_\mathrm{r}]
	+\lambda_\mathrm{f}(1+\alpha\mathbf{1}[x\in D_{\mathrm{hf}}])\,\mathbf{1}[x\in D_\mathrm{f}],
\end{equation}

\vspace{-0.25cm}
where $\lambda_\mathrm{r},\lambda_\mathrm{f}>0$ balance retain and forget guidance, and $\alpha\ge 0$ emphasizes on hard forget samples.

\noindent\textbf{InDist-guided stop criterion.}
Unlearning stops once coverage exceeds a target threshold $\tau$:
\begin{equation}
	\label{eq:stop}
	\footnotesize
	\mathrm{InDist}(S;\theta)\ge\tau, \forall S\in\{D_\mathrm{r},\;D_\mathrm{f},\;D_{\mathrm{hf}}\}
\end{equation}

\vspace{-0.25cm}
This makes the stopping decision explicit at the sample level and avoids  ad-hoc step budgets.
	\vspace{-0.1cm}
	
\vspace{-0.1cm}
\subsection{Theoretical Guarantee for SALA's DNO Upper Bound}
\label{subsec:theory}
\vspace{-0.2cm}
This section derives an upper bound on DNO in~\eqref{eq:dno} from the InDist constraint imposed by the SALA stop criterion. We first bound the overlap gap for each sample and then average over the set.

\noindent\textbf{Setup.} 
Fix a sample $x$ with target distribution $Q$. Let the SALA-unlearned model distribution induce
{\footnotesize\[
	X\sim Q^{\mathrm{SALA}}(\cdot\mid x)=\mathcal{N}\!\bigl(\mu^{\mathrm{SALA}}(x),(\sigma^{\mathrm{SALA}}(x))^2\bigr).
	\]}
Define 
{\footnotesize$
	\Delta(x)\triangleq \frac{\mu^{\mathrm{SALA}}(x)-\mu_Q(x)}{\sigma_Q(x)},~
	\rho(x)\triangleq \frac{\sigma^{\mathrm{SALA}}(x)}{\sigma_Q(x)},~
	\eta(x)\triangleq \frac{X-\mu^r(x)}{\sigma_Q(x)}.
	$}
\begin{theorem}[DNO upper bound from SALA]
	\label{thm:dno_bound}
	Assume that SALA makes the InDist constraint hold for $x$:
	\begin{equation}
		\footnotesize
		\label{eq:indist_constraint}
		\mathbb{P}_{X\sim Q^{\mathrm{SALA}}(\cdot\mid x)}\!\left(\left|\frac{X-\mu_Q^r(x)}{\sigma_Q(x)}\right|<Z_{\mathrm{cut}}\right)\ \ge\ \tau.
	\end{equation}
	The mean shift in~\eqref{eq:zscore} induces variance expansion, which gives a non-degenerate lower bound on the effective scale:
	{\footnotesize\[
		\rho_{\min}\triangleq \max\!\Bigl(0,\ \mathrm{var}_\epsilon-\mathrm{Cov}\bigl(\epsilon(x),\eta(x)\bigr)\Bigr).
		\]}
	If $\rho(x)\in[\rho_{\min},\frac{Z_{\mathrm{cut}}}{q}]$ with $\rho_{\min}>0$, then the \emph{per-sample} overlap gap satisfies
	\begin{equation}
		\footnotesize
		\label{eq:dno_x_ub}
		1-\int \min\!\bigl(Q^{\mathrm{SALA}}(s\mid x),\,Q(s\mid x)\bigr)\,ds
		\ \le\
		\sqrt{2\bigl(1-\mathrm{BC}_{\min}\bigr)},
            	\vspace{-0.02cm}
	\end{equation}

    \vspace{-0.25cm}
	where $\mathrm{BC}_{\min}$ is the worst-case Bhattacharyya coefficient~\cite{kailath1967divergence} over the feasible $(\Delta,\rho)$ region:
	\begin{equation}
		\footnotesize
		\label{eq:bcmin}
		\mathrm{BC}_{\min}
		=\min_{\rho\in[\rho_{\min},\frac{Z_{\mathrm{cut}}}{q}]}
		\left[
		\sqrt{\frac{2\rho}{1+\rho^2}}\;
		\exp\!\left(
		-\frac{(Z_{\mathrm{cut}}-\rho q)^2}{4(1+\rho^2)}
		\right)
		\right],\qquad
		q\triangleq \Phi^{-1}\!\left(\frac{1+\tau}{2}\right).
        	\vspace{-0.03cm}
	\end{equation}

    \vspace{-0.35cm}
	Consequently, the sample-averaged DNO in~\eqref{eq:dno} is bounded by the same right-hand side as \eqref{eq:dno_x_ub}:
	\begin{equation}
		\label{eq:dno_set_ub}
		\footnotesize
		\mathrm{DNO}
		\ \le\
		\sqrt{2\bigl(1-\mathrm{BC}_{\min}\bigr)}.
        	\vspace{-0.2cm}
	\end{equation}
\end{theorem}
The proof of Theorem~\ref{thm:dno_bound} is provided in Appendix~\ref{app:dno_bound_proof}.

\noindent\textbf{Instantiation.} 
For a sanity check, a simplifying assumption $\mathrm{Cov}(\epsilon,\eta)=0$ is adopted. Under an easy-to-achieve setting ($\tau=0.95,\ Z_{\mathrm{cut}}=2$,\ $\epsilon=\mathrm{clip}(\mathcal{N}(0,1),\pm1.8)$), and assuming exact access to $Q_{\mathrm{in}}(\cdot\mid x)$ and $Q_{\mathrm{out}}(\cdot\mid x)$, Theorem~\ref{thm:dno_bound} yields a closed-form DNO upper bound of about $0.10$. This bound matches the empirical DNO of SALA, about $0.11$, in Section~\ref{sec:baseline_comparison}.

\vspace{-0.18cm}
\subsection{SALA with Access Only to the Forget Set and Its DNO Guarantees}
\vspace{-0.25cm}
\label{subsec:single_side_sala}

To accommodate unlearning settings with no access to retain samples, we further relax SALA by removing its dependence on the retain set in all stages where it is used, including the fast probe, the unlearning guidance, and the stop criterion. In other words, \textbf{in this variant, SALA no longer explicitly optimizes any term involving $x\in D_\mathrm{r}$}.

Even under this one-sided constraint, SALA still admits DNO guarantees as long as the base unlearning algorithm $\mathcal{U}$ provides a condition on $D_\mathrm{r}$ that can equivalently replace $\mathrm{InDist}(D_\mathrm{r};\theta)$ criterion. Corollary~\ref{cor:dno_bound_single} summarizes this claim, with the proof given in Appendix~\ref{app:dno_bound_proof_single}.

\begin{corollary}[DNO upper bound for SALA with forget-set access only]
	\label{cor:dno_bound_single}
	Assume that the base unlearning algorithm $\mathcal{U}$ satisfies the following retain-consistency condition:
	\vspace{-0.0cm}
	\begin{equation}
		\label{eq:retain_consistency_loss}
		\footnotesize
		L_r(\theta)
		\triangleq
		\mathbb E_{x\sim D_\mathrm{r}}
		\left[
		\left(
		\frac{s_\theta(x)-s_{\theta_0}(x)}{\sigma_{\mathrm{in}}(x)}
		\right)^2
		\right]
		\le
		(1-\tau) Z_{\mathrm{cut}}^2,
	\end{equation}

\vspace{-0.2cm}
	where $\theta_0$ denotes the model before unlearning. Then the DNO upper bound in Theorem~\ref{thm:dno_bound} continues to hold for SALA with access only to the forget set.
\end{corollary}

%% file: 4_sample_logits_estimation.tex
\vspace{-0.25cm}
\section{Efficient Estimation of Per-Sample (Non-)Member Distributions}
\vspace{-0.29cm}
\label{sec:estimation}
SALA requires the per-sample member and non-member output distributions in Eq.~\eqref{eq:q_in_out}. A straightforward baseline is the shadow-model estimation~\cite{carlini2022membership}, which trains multiple models on different forget-retain splits and aggregates their outputs. Compatible member and non-member score statistics retained from a shadow-model memorization pipeline, as used in SUM~\cite{zhao2024what}, can be reused to estimate these distributions. However, the shadow-model approach itself is generally too expensive. Therefore, the technical challenge for SALA is to \textbf{estimate the per-sample distributions for all training samples at a cost comparable to a single training run, while also producing the target model itself}.
\vspace{-0.18cm}
\subsection{Method: Split--Merge Training with Segment Ensemble (SMSE) Estimation}
\vspace{-0.23cm}
This paper proposes SMSE to estimate the distributions for every $x$ using one structured training run that also produces the final model. In standard training, estimating $Q_{\mathrm{out}}(\cdot\mid x)$ is challenging: every sample is always in-training, so no true non-member state exists. SMSE estimates non-member behavior by splitting the training set into two complementary halves and training two branches from shared warmup weights; each sample is excluded from one branch during half-data training. A merge stage then recovers a single final model via Federated Averaging~\cite{kairouz2021advances}. To capture training randomness and approximate training-induced variability, SMSE uses an ensemble of late-stage checkpoints rather than many independent full retrains.

\noindent\textbf{Overview.} 
SMSE consists of four stages: (1) \textbf{Split and warmup}: the training set $D$ is split into two disjoint subsets $A$ and $B$, followed by a warmup training on $D$ that initializes shared weights; (2) \textbf{Half-data training}: two models are trained independently, producing $\theta_A$ from $A$ and $\theta_B$ from $B$; (3) \textbf{Merge}: $\theta_A$ and $\theta_B$ are merged using Federated Averaging~\cite{kairouz2021advances}, and the merged model is then fine-tuned on $A\cup B$; (4) \textbf{Segment ensemble}: checkpoints from the $A$-model and $B$-model are aggregated as an ensemble to estimate per-sample score moments. Appendix~\ref{app:algorithm} provides the SMSE pseudo-code.

\noindent\textbf{Segment-ensemble estimation of per-sample moments.} 
Let $s_\theta(x)$ denote the scalar score in~\eqref{eq:logodds}. Let
$\{\theta_A^{(m)}\}_{m=1}^M$ and $\{\theta_B^{(m)}\}_{m=1}^M$
be $M$ late-stage checkpoints from the two half-data trainings. For each sample $x$, define role-specific checkpoints $\theta_t^{(m)}(x)$ for $t\in\{\mathrm{in},\mathrm{out}\}$: if $x\in A$, then $\theta_{\mathrm{in}}^{(m)}(x)=\theta_A^{(m)}$ and $\theta_{\mathrm{out}}^{(m)}(x)=\theta_B^{(m)}$; if $x\in B$, the two roles are reversed. SMSE then estimates
\vspace{-0.18cm}
\begin{equation}
	\footnotesize
	\hat{\mu}_{t}(x)=\frac{1}{M}\sum\nolimits_{m=1}^{M} s_{\theta_t^{(m)}(x)}(x),\qquad
	\hat{\sigma}^2_{t}(x)=\frac{1}{M}\sum\nolimits_{m=1}^{M}\bigl(s_{\theta_t^{(m)}(x)}(x)-\hat{\mu}_{t}(x)\bigr)^2,
	\label{eq:smse_moments}
\end{equation}
for $t\in\{\mathrm{in},\mathrm{out}\}$. These estimates define $\hat{Q}_{\mathrm{in}}(\cdot\mid x)$ and $\hat{Q}_{\mathrm{out}}(\cdot\mid x)$ in~\eqref{eq:q_in_out} for all $x$, at a cost close to a single training run.

\vspace{-0.2cm}
\subsection{Effect of Distribution Estimation Error on SALA's DNO Bound in Theorem~\ref{thm:dno_bound}}
\vspace{-0.2cm}
\label{subsec:est_error_bound}
The DNO bound in Theorem~\ref{thm:dno_bound} assumes exact target distributions from multiple retrained shadow models. In practice, SALA uses lightweight SMSE estimates instead, which introduces distribution estimation error and thereby affects the bound. Next, we quantify that effect and derive the corresponding adjusted DNO bound.

The key point is that Theorem~\ref{thm:dno_bound} still controls the discrepancy induced by SALA, but now with respect to the estimated target distribution $\hat Q(\cdot\mid x)$ by SMSE rather than the exact target distribution $Q(\cdot\mid x)$. Consequently, the DNO with respect to $\hat Q(\cdot\mid x)$ can be decomposed into two parts: an estimation error term and a SALA-induced term. This is summarized in the following corollary.

\begin{corollary}[Adjusted DNO Bound under Distribution Estimation Error]
	\label{cor:est_error_bound}
	For a fixed sample $x$ and a fixed role (member or non-member), let $Q(\cdot\mid x)$ denote the true target distribution, let $\hat Q(\cdot\mid x)$ denote its estimate, and let $Q^{\mathrm{SALA}}(\cdot\mid x)$ denote the distribution induced by SALA. Then
	\begin{equation}
		\label{eq:tv_triangle}
		\footnotesize
		\mathrm{DNO}(x)
		\triangleq
		\mathrm{TV}\!\bigl(Q^{\mathrm{SALA}}(\cdot\mid x),Q(\cdot\mid x)\bigr)
		\le
		\underbrace{\mathrm{TV}\!\bigl(Q(\cdot\mid x),\hat Q(\cdot\mid x)\bigr)}_{\mathrm{DNO}_{\mathrm{est}}(x)}
		+
		\underbrace{\mathrm{TV}\!\bigl(Q^{\mathrm{SALA}}(\cdot\mid x),\hat Q(\cdot\mid x)\bigr)}_{\mathrm{DNO}_{\hat Q}(x)}.
	\end{equation}
	
	\vspace{-0.2cm}
	Here, $\mathrm{DNO}_{\mathrm{est}}(x)$ is the distribution estimation error, which is explicitly bounded in Appendix~\ref{app:smse_error_bound}. $\mathrm{DNO}_{\hat Q}(x)$ is the per-sample DNO component with respect to the estimated ${\hat Q}(x)$. By Theorem~\ref{thm:dno_bound} applied to $\hat Q(\cdot\mid x)$, we have {\small $\mathrm{DNO}_{\hat Q}(\cdot\mid x)
	\le
	\sqrt{2\bigl(1-\widehat{\mathrm{BC}}_{\min}\bigr)}$}.
\end{corollary}

%% file: 5_experiments.tex
    \vspace{-0.25cm}
\section{Experiments}
\label{sec:ex}
\vspace{-0.28cm}
We evaluate whether SALA enables sample-grained approximate unlearning. 
Specifically, we test whether dataset-level aggregate metrics hide per-sample deviation from exact retraining, whether SALA reduces this deviation under DNO while preserving utility, privacy, and efficiency, and whether SMSE provides accurate target estimates for SALA. 
Code is available at \url{https://github.com/SALA-Unlearning/Sala-unlearning}.
\vspace{-0.25cm}
\subsection{Experimental Setup}
\vspace{-0.2cm}
\label{sec:ex_setup}

\noindent\textbf{Datasets and models.}
We evaluate SALA on classification and generation tasks. For classification, we use CIFAR-10~\cite{krizhevsky2009learning} with ResNet-18~\cite{he16deep}, and ImageNet-1000~\cite{russakovsky2015imagenet} with ResNet-50. For generation, we use WikiText-103~\cite{merity2016pointer} with GPT-2~\cite{radford2019language}. Appendix~\ref{app:setup} provides implementation details.

\noindent\textbf{Unlearning setup and metrics.}
We evaluate three unlearning scenarios: random sample unlearning~\cite{bourtoule2021machine}, class-wise unlearning~\cite{fan2024salun} for classification, and sequence unlearning~\cite{naderloui2025rectifying} for generation. Sequence unlearning uses contiguous text segments as forget sets. We evaluate unlearning from efficacy, privacy, and runtime perspectives, following Section~\ref{subsec:pd_evaluation}. DNO is computed over multiple unlearning tasks, with 64 tasks on CIFAR-10 and 16 tasks on ImageNet-1000 and WikiText-103. We compare against 12 approximate unlearning baselines on CIFAR-10. Due to memory constraints, we include five representative baselines on ImageNet-1000 and WikiText-103. All baselines use manually tuned hyperparameters, and Appendix~\ref{app:setup} provides the baseline configurations and search protocol.

\noindent\textbf{SALA and SMSE configuration.}
SALA is an auxiliary objective for iterative unlearning, so we pair it with SalUn~\cite{fan2024salun} in the main experiments. Results on other AU backbones are consistent and are reported in Appendix~\ref{app:other_backbones}. We evaluate three variants: \textbf{SALA} with shadow-model targets, \textbf{SALA-SMSE} with SMSE-estimated targets, and \textbf{SALA w/ $D_\mathrm{f}$ only}, which constrains only the forget set. Appendix~\ref{app:setup} gives the full SALA and SMSE configurations, and Appendices~\ref{sec:indist_ablation} and~\ref{app:smse_ablation} report the corresponding parameter ablations.

\vspace{-0.1cm}
\subsection{Evaluation of SMSE Estimation}
\vspace{-0.1cm}
\label{sec:estimation_accuracy}

We evaluate SMSE by comparing its estimates of 
$(\mu_{\mathrm{in}},\sigma_{\mathrm{in}},\mu_{\mathrm{out}},\sigma_{\mathrm{out}})$
in~\eqref{eq:q_in_out} against expensive shadow-model references and a lightweight single-run baseline. 
The shadow-model reference trains $64$ models on CIFAR-10 and $16$ models on ImageNet-1000 and WikiText-103 under different forget-retain splits, which incurs high computational cost.
The single-run baseline uses a Laplace approximation to estimate $(\mu_{\mathrm{in}},\sigma_{\mathrm{in}})$~\cite{daxberger2021laplace}, sets 
$\mu_{\mathrm{out}}(x)\approx \mu_{\mathrm{in}}(x)-\Delta\mu_{\mathrm{SI}}(x)$
using the self-influence shift $\Delta\mu_{\mathrm{SI}}(x)$, and obtains $\sigma_{\mathrm{out}}(x)$ through the same uncertainty propagation~\cite{brophy2023adapting}.

\begingroup
\setlength{\intextsep}{6pt}
\begin{table}[H]
	\centering
	\scriptsize
	\setlength{\tabcolsep}{2.08pt}
	\renewcommand{\arraystretch}{1.05}
	\caption{Estimation evaluation of per-sample member and non-member distribution parameters.}
	\label{tab:estimation_accuracy}
	\begin{tabular}{llcccccccc cc}
		\toprule
		\multirow{3}{*}{\textbf{Dataset}} &
		\multirow{3}{*}{\textbf{Method}} &
		\multicolumn{2}{c}{$\boldsymbol{\mu_{\mathrm{in}}}$} &
		\multicolumn{2}{c}{$\boldsymbol{\mu_{\mathrm{out}}}$} &
		\multicolumn{2}{c}{$\boldsymbol{\sigma_{\mathrm{in}}}$} &
		\multicolumn{2}{c}{$\boldsymbol{\sigma_{\mathrm{out}}}$} & 
			\multirow{3}{*}{\textbf{$\mathrm{DNO}_{\mathrm{est}}$}$\downarrow$} &
		\multirow{3}{*}{\textbf{Cost}$\downarrow$ } \\
		\cmidrule(lr){3-4} \cmidrule(lr){5-6} \cmidrule(lr){7-8} \cmidrule(lr){9-10}
		& & \textbf{MSE}$\downarrow$ & \textbf{Pearson r}$\uparrow$
		& \textbf{MSE}$\downarrow$ & \textbf{Pearson r}$\uparrow$
		& \textbf{MSE}$\downarrow$ & \textbf{Pearson r}$\uparrow$
		& \textbf{MSE}$\downarrow$ & \textbf{Pearson r}$\uparrow$ & & \\
		\midrule
		\multirow{2}{*}{\textbf{CIFAR-10}} 
		& Laplace + SI
		& 2.80 & 0.48
		& 9.48 & 0.41
		& 0.78 & 0.04
		& 2.45 & 0.05
		&0.4580
		& 1.24$\times$ \\
		& SMSE (Ours)
		& \textbf{1.29} & \textbf{0.79}
		& \textbf{1.90} & \textbf{0.92}
		& \textbf{0.67} & \textbf{0.58}
		& \textbf{0.52} & \textbf{0.63}
		&\textbf{0.1801}
		& 1.00$\times$ \\
		\midrule
		\multirow{2}{*}{\textbf{ImageNet-1000}} 
		& Laplace + SI
		& 0.74 & 0.90
		& 4.23 & 0.45
		& 0.92 & 0.03
		& 2.72 & 0.04
		& 0.4835
		& 1.08$\times$ \\
		& SMSE (Ours)
		& \textbf{0.28} & \textbf{0.98}
		& \textbf{0.31} & \textbf{0.98}
		& \textbf{0.59} & \textbf{0.62}
		& \textbf{0.67} & \textbf{0.52}
		& \textbf{0.1064}
		& 1.00$\times$ \\
		\midrule
		\multirow{2}{*}{\textbf{WikiText-103}} 
		& Laplace + SI
		& 0.15 & 0.92
		& 1.56 & 0.48
		& 0.85 & 0.05
		& 1.14 & -0.08
		& 0.3913
		& 1.16$\times$ \\
		& SMSE (Ours)
		& \textbf{0.0065} & \textbf{0.99}
		& \textbf{0.0067} & \textbf{0.99}
		& \textbf{0.47} & \textbf{0.65}
		& \textbf{0.56} & \textbf{0.68}
		& \textbf{0.0935}
		& 1.00$\times$ \\
		\bottomrule
	\end{tabular}
	\vspace{-0.10cm}
\end{table}
\endgroup

\begin{figure}[!htbp]
	\centering
	\subfloat[$\mu_{\text{in}}$]{
		\label{fig:mu_in_esti}
		\includegraphics[width=0.23\columnwidth]{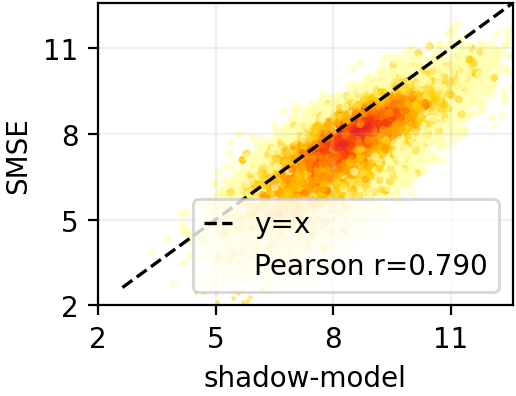}}
	\subfloat[$\mu_{\text{out}}$]{
		\label{fig:mu_out_esti}
		\includegraphics[width=0.23\columnwidth]{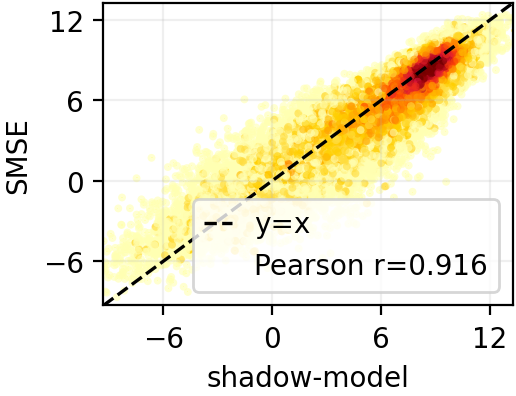}}
	\subfloat[$\sigma_{\text{in}}$]{
		\label{fig:sigma_in_esti}
		\includegraphics[width=0.23\columnwidth]{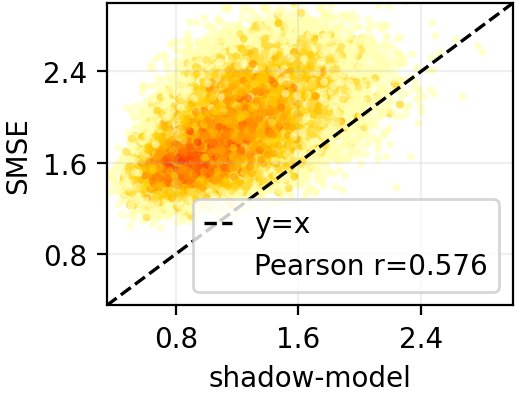}}
	\subfloat[$\sigma_{\text{out}}$]{
		\label{fig:sigma_out_esti}
		\includegraphics[width=0.23\columnwidth]{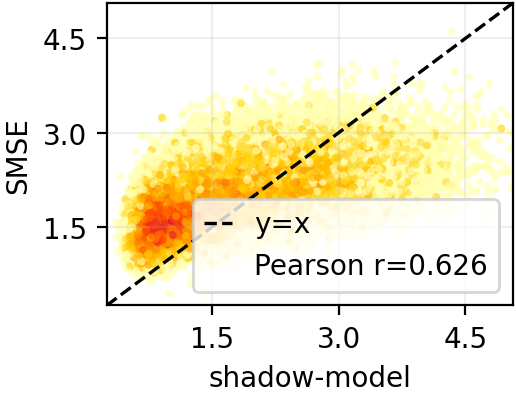}}
	\includegraphics[width=0.042\columnwidth]{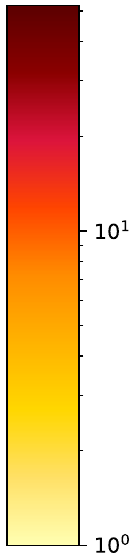}
	\caption{Per-sample agreement between SMSE-estimated moments and shadow-model references on CIFAR-10.}
	\label{fig:segment_ensemble_quality}
	\vspace{0cm}
\end{figure}

\noindent\textbf{Results.} 
We measure estimation quality by MSE, Pearson correlation, and $\mathrm{DNO}_{\mathrm{est}}$ against shadow-model references, and report training cost relative to a standard training run. Table~\ref{tab:estimation_accuracy} shows that SMSE improves over Laplace+SI across all datasets and moments with $1.00\times$ nominal training process. Appendix~\ref{app:rebuttal-cost} supplements the experiments with comprehensive computational costs. The gains are largest for non-member and variance estimates: on CIFAR-10, SMSE raises the correlation of $\mu_{\mathrm{out}}$ from $0.41$ to $0.92$ and lowers $\mathrm{DNO}_{\mathrm{est}}$ from $0.4580$ to $0.1801$. 
Figure~\ref{fig:segment_ensemble_quality} visualizes the per-sample agreement between SMSE estimates and shadow-model references, and Table~\ref{tab:smse_utility} in Appendix~\ref{app:smse_utility} reports the utility of SMSE-trained models. It indicates that SMSE preserves utility, with at most a $0.18\%$ accuracy drop on vision tasks and a $0.99$ PPL increase on WikiText-103 compared with standard training.

\vspace{-0.25cm}
\subsection{Comprehensive Evaluation for Unlearning Methods}
\label{sec:baseline_comparison}
\vspace{-0.28cm}

Table~\ref{tab:baseline_comparison} compares SALA with representative approximate unlearning baselines on random sample unlearning and sequence unlearning. Class-wise unlearning is reported in Appendix~\ref{sec:ex_classwise}. The results show that aggregate gap-to-retraining metrics alone are insufficient to characterize approximate unlearning. On CIFAR-10, SalUn reaches ToW = 0.95, but its DNO remains 0.9009 and its RULI remains 92.38\%. The same pattern appears on ImageNet-1000, where SalUn reaches ToW = 0.95 but still has DNO = 0.9256 and RULI = 92.30\%. Thus, dataset-level similarity to retraining does not necessarily eliminate sample-level deviation or unlearning-aware privacy leakage.

\Needspace{9\baselineskip}
SALA consistently reduces the sample-grained distributional gap while preserving utility and efficiency. Across CIFAR-10, ImageNet-1000, and WikiText-103, SALA achieves the lowest DNO among approximate methods, with DNO = 0.1365, 0.1268, and 0.1048, respectively. It also keeps RULI close to random guessing, with 52.15\%, 52.35\%, and 50.92\% on the three datasets. Across all three datasets, SALA reduces runtime by more than an order of magnitude relative to retraining. SALA-SMSE and SALA w/ ${D}_\mathrm{f}$ only have higher DNO than full SALA, which is consistent with estimation error and relaxed retain-side constraints, but they still keep DNO substantially below the corresponding SalUn baseline.
\begin{table*}
	\centering
	\scriptsize
	\setlength{\tabcolsep}{3pt}
	\renewcommand{\arraystretch}{1.30}
	\caption{Comprehensive comparison of approximate unlearning methods.}
	\vspace{-0.08cm}
	\label{tab:baseline_comparison}
	\begin{tabular*}{\linewidth}{@{\extracolsep{\fill}}lcccccccc@{}}
		\toprule
		\raisebox{0.5\normalbaselineskip}{\textbf{Method}} &
		\shortstack{\textbf{TA(\%)}\\\textbf{(T-PPL)}} &
		\shortstack{\textbf{FA(\%)}\\\textbf{(F-PPL)}}&
		\shortstack{\textbf{RA(\%)}\\\textbf{(R-PPL)}}&
		\shortstack{\textbf{ToW}\\$\uparrow$} &
		\shortstack{\textbf{DNO}\\$\downarrow$} &
		\shortstack{\textbf{MIA efficacy}\\(\%)$\downarrow$} &
		\shortstack{\textbf{RULI}\\(\%)$\downarrow$} &
		\shortstack{\textbf{Time}\\(s)$\downarrow$} \\
		\hline
		\rowcolor{blue!7}
		\multicolumn{9}{c}{\textbf{CIFAR-10 (ResNet-18)}} \\ \hline
		Retrain&
		93.84\scalebox{0.7}{$\!\pm\!$0.21}& 
		93.68\scalebox{0.7}{$\!\pm\!$0.28} &
		100.00\scalebox{0.7}{$\!\pm\!$0.01} &
		1.00\scalebox{0.7}{$\!\pm\!$0.00}&
        0.0646\scalebox{0.7}{$\!\pm\!$0.0022} &
		50.57\scalebox{0.7}{$\!\pm\!$0.25}& 
		50.00\scalebox{0.7}{$\!\pm\!$0.00}&
		1625.53\scalebox{0.7}{$\!\pm\!$5.33}\\
		\hline
		FT&
		89.04\scalebox{0.7}{$\!\pm\!$1.13} & 
		92.32\scalebox{0.7}{$\!\pm\!$1.20}& 
		95.84\scalebox{0.7}{$\!\pm\!$1.10} &
		0.90\scalebox{0.7}{$\!\pm\!$0.01} &
        0.3388\scalebox{0.7}{$\!\pm\!$0.0053} &
		52.09\scalebox{0.7}{$\!\pm\!$0.22} &
		70.01\scalebox{0.7}{$\!\pm\!$0.73}  &
		242.01\scalebox{0.7}{$\!\pm\!$42.25} \\
		GA&
		86.01\scalebox{0.7}{$\!\pm\!$1.42} & 
		85.93\scalebox{0.7}{$\!\pm\!$2.10} & 
		95.13\scalebox{0.7}{$\!\pm\!$1.20} &
		0.80\scalebox{0.7}{$\!\pm\!$0.01} &
        0.6228\scalebox{0.7}{$\!\pm\!$0.0054} &
		51.04\scalebox{0.7}{$\!\pm\!$0.52}  & 
		87.69\scalebox{0.7}{$\!\pm\!$0.86}  & 
		149.30\scalebox{0.7}{$\!\pm\!$2.02} \\
		RL&
		88.94\scalebox{0.7}{$\!\pm\!$0.71} & 
		87.63\scalebox{0.7}{$\!\pm\!$0.90} & 
		98.08\scalebox{0.7}{$\!\pm\!$0.60} &
		0.87\scalebox{0.7}{$\!\pm\!$0.02} &
        0.6875\scalebox{0.7}{$\!\pm\!$0.0059} &
		55.10\scalebox{0.7}{$\!\pm\!$0.21}  & 
		82.35\scalebox{0.7}{$\!\pm\!$14.31}  & 
		321.32\scalebox{0.7}{$\!\pm\!$7.70} \\
		IU&
		86.90\scalebox{0.7}{$\!\pm\!$5.61} & 
		93.17\scalebox{0.7}{$\!\pm\!$6.25} & 
		93.52\scalebox{0.7}{$\!\pm\!$6.26} &
		0.86\scalebox{0.7}{$\!\pm\!$0.03} &
        0.3982\scalebox{0.7}{$\!\pm\!$0.0090} &
		52.35\scalebox{0.7}{$\!\pm\!$2.13}  & 
		64.52\scalebox{0.7}{$\!\pm\!$5.17}  & 
		\textbf{41.59}\scalebox{0.7}{$\!\pm\!$0.51} \\
		FF&
		76.11\scalebox{0.7}{$\!\pm\!$7.42} & 
		80.85\scalebox{0.7}{$\!\pm\!$8.08} & 
		80.83\scalebox{0.7}{$\!\pm\!$8.11} &
		0.57\scalebox{0.7}{$\!\pm\!$0.03} &
        0.8947\scalebox{0.7}{$\!\pm\!$0.0234} &
		53.30\scalebox{0.7}{$\!\pm\!$0.15}  & 
		90.39\scalebox{0.7}{$\!\pm\!$8.41}  & 
		{2408.39\scalebox{0.7}{$\!\pm\!$105.19}} \\
		MinNorm-OG&
		88.12\scalebox{0.7}{$\!\pm\!$1.28} &
		94.90\scalebox{0.7}{$\!\pm\!$1.19} &
		94.91\scalebox{0.7}{$\!\pm\!$1.22} &
		0.88 \scalebox{0.7}{$\!\pm\!$0.02} &
        0.6355\scalebox{0.7}{$\!\pm\!$0.0075} &
		57.31\scalebox{0.7}{$\!\pm\!$0.49} &
		81.81\scalebox{0.7}{$\!\pm\!$1.75} &
		60.20 \scalebox{0.7}{$\!\pm\!$7.75} \\
		
		SEMU&
		92.66\scalebox{0.7}{$\!\pm\!$0.34} &
		95.56\scalebox{0.7}{$\!\pm\!$0.35} &
		99.82\scalebox{0.7}{$\!\pm\!$0.03} &
		0.96 \scalebox{0.7}{$\!\pm\!$0.01} &
        0.7390\scalebox{0.7}{$\!\pm\!$0.0050} &
		53.53\scalebox{0.7}{$\!\pm\!$0.49} &
		82.42\scalebox{0.7}{$\!\pm\!$0.86} &
		48.15 \scalebox{0.7}{$\!\pm\!$5.76} \\
		
		$\ell_1$ sparsity&
		88.18\scalebox{0.7}{$\!\pm\!$3.47} & 
		91.94\scalebox{0.7}{$\!\pm\!$4.48} & 
		95.80\scalebox{0.7}{$\!\pm\!$2.98} &
		0.88\scalebox{0.7}{$\!\pm\!$0.02}  &
        0.3376\scalebox{0.7}{$\!\pm\!$0.0098} &
		53.09\scalebox{0.7}{$\!\pm\!$2.72}  & 
		70.03\scalebox{0.7}{$\!\pm\!$5.24}  & 
		150.28\scalebox{0.7}{$\!\pm\!$28.45} \\
		SalUn&92.36\scalebox{0.7}{$\!\pm\!$0.45} &
		95.29\scalebox{0.7}{$\!\pm\!$0.57} &
		98.61\scalebox{0.7}{$\!\pm\!$0.36} &
		0.95\scalebox{0.7}{$\!\pm\!$0.01}&
        0.9009\scalebox{0.7}{$\!\pm\!$0.0015} &
		51.87\scalebox{0.7}{$\!\pm\!$0.21}  &
		92.38\scalebox{0.7}{$\!\pm\!$0.28}  &
		221.92\scalebox{0.7}{$\!\pm\!$96.37} \\
		RUM&
		91.66\scalebox{0.7}{$\!\pm\!$0.47} & 
		92.96\scalebox{0.7}{$\!\pm\!$0.49} &
		96.89\scalebox{0.7}{$\!\pm\!$0.39} &
		0.94\scalebox{0.7}{$\!\pm\!$0.01} &
        0.7826\scalebox{0.7}{$\!\pm\!$0.0062} &
		50.88\scalebox{0.7}{$\!\pm\!$0.22}  & 
		85.35\scalebox{0.7}{$\!\pm\!$1.37} &
		375.29\scalebox{0.7}{$\!\pm\!$83.05} \\
		SFRon&
		87.31\scalebox{0.7}{$\!\pm\!$1.64} & 
		90.39\scalebox{0.7}{$\!\pm\!$2.45} & 
		96.77\scalebox{0.7}{$\!\pm\!$1.05} &
		0.87\scalebox{0.7}{$\!\pm\!$0.01}  &
        0.4518\scalebox{0.7}{$\!\pm\!$0.0091} &
		54.23\scalebox{0.7}{$\!\pm\!$0.98}  & 
		76.75\scalebox{0.7}{$\!\pm\!$2.04}  & 
		
		67.97\scalebox{0.7}{$\!\pm\!$0.70} \\
		SCRUB&
		90.29\scalebox{0.7}{$\!\pm\!$2.31} &
		94.06\scalebox{0.7}{$\!\pm\!$3.15} &
		95.74\scalebox{0.7}{$\!\pm\!$3.00} &
		0.92\scalebox{0.7}{$\!\pm\!$0.03}  &
        0.3644\scalebox{0.7}{$\!\pm\!$0.0086} &
		52.23\scalebox{0.7}{$\!\pm\!$0.57}  & 
		68.67\scalebox{0.7}{$\!\pm\!$3.62}  & 
		101.93\scalebox{0.7}{$\!\pm\!$11.98} \\
		\hline
		SALA& 
		94.44\scalebox{0.7}{$\!\pm\!$0.28}&
		95.58\scalebox{0.7}{$\!\pm\!$0.21}&
		100.00\scalebox{0.7}{$\!\pm\!$0.00}&
		0.97\scalebox{0.7}{$\!\pm\!$0.02} &
        \textbf{0.1365}\scalebox{0.7}{$\!\pm\!$0.0055} &
		\textbf{52.42}\scalebox{0.7}{$\!\pm\!$0.16}&
		\textbf{52.15}\scalebox{0.7}{$\!\pm\!$1.30}&
		53.22\scalebox{0.7}{$\!\pm\!$3.21} \\
		SALA-SMSE&
		94.25\scalebox{0.7}{$\!\pm\!$0.27}&
		95.13\scalebox{0.7}{$\!\pm\!$0.17}&
		99.97\scalebox{0.7}{$\!\pm\!$0.01}&
		0.98\scalebox{0.7}{$\!\pm\!$0.01} &
        0.2508\scalebox{0.7}{$\!\pm\!$0.0056} &
		54.28\scalebox{0.7}{$\!\pm\!$0.36}&
		55.97\scalebox{0.7}{$\!\pm\!$0.53}&
		56.47\scalebox{0.7}{$\!\pm\!$5.35} 
		\\
		SALA w/ ${D}_\mathrm{f}$ only&
		93.75\scalebox{0.7}{$\!\pm\!$0.29}&
		94.49\scalebox{0.7}{$\!\pm\!$0.30}&
		99.75\scalebox{0.7}{$\!\pm\!$0.08}&
		\textbf{0.99}\scalebox{0.7}{$\!\pm\!$0.01} &
        0.2453\scalebox{0.7}{$\!\pm\!$0.0042} &
		53.37\scalebox{0.7}{$\!\pm\!$0.34}&
		56.75\scalebox{0.7}{$\!\pm\!$2.49}&
		59.35\scalebox{0.7}{$\!\pm\!$6.26} 
		\\
		\hline
		\rowcolor{green!7}
		\multicolumn{9}{c}{\textbf{ImageNet-1000 (ResNet-50)}} \\ \hline
		retrain
		& 71.26\scalebox{0.7}{$\!\pm\!$0.47}
		& 71.27\scalebox{0.7}{$\!\pm\!$0.52}
		& 89.45\scalebox{0.7}{$\!\pm\!$0.48}
		& 1.00\scalebox{0.7}{$\!\pm\!\mathrm{0.00}$}
		&
        0.0277\scalebox{0.7}{$\!\pm\!$0.0032} & 50.53\scalebox{0.7}{$\!\pm\!$0.09}
		& 50.01\scalebox{0.7}{$\!\pm\!$0.01}
		& 1.42e5\scalebox{0.7}{$\!\pm\!$5.47e3} \\ \hline
		FT
		& 68.36\scalebox{0.7}{$\!\pm\!$0.39}
		& 72.65\scalebox{0.7}{$\!\pm\!$0.64}
		& 80.89\scalebox{0.7}{$\!\pm\!$0.15}
		& 0.85\scalebox{0.7}{$\!\pm\!$0.02}
		&
        0.3497\scalebox{0.7}{$\!\pm\!$0.0090} & 52.31\scalebox{0.7}{$\!\pm\!\!$0.14}
		& 65.42\scalebox{0.7}{$\!\pm\!$0.42}
		& 1.93e4\scalebox{0.7}{$\!\pm\!$7.52e2} \\
		GA
		& 75.71\scalebox{0.7}{$\!\pm\!$0.40}
		& 87.11\scalebox{0.7}{$\!\pm\!$0.40}
		& 87.22\scalebox{0.7}{$\!\pm\!$0.38}
		& 0.71\scalebox{0.7}{$\!\pm\!$0.03}
		&
        0.3931\scalebox{0.7}{$\!\pm\!$0.0077} & 51.43\scalebox{0.7}{$\!\pm\!\!$0.25}
		& 67.66\scalebox{0.7}{$\!\pm\!$1.21}
		& 2.92e3\scalebox{0.7}{$\!\pm\!$2.12e2} \\
		RL
		& 74.64\scalebox{0.7}{$\!\pm\!$1.33}
		& 74.95\scalebox{0.7}{$\!\pm\!$0.25}
		& 76.28\scalebox{0.7}{$\!\pm\!$1.87}
		& 0.77\scalebox{0.7}{$\!\pm\!$0.05}
		&
        0.7048\scalebox{0.7}{$\!\pm\!$0.0052} & 54.96\scalebox{0.7}{$\!\pm\!\!$0.35}
		& 86.29\scalebox{0.7}{$\!\pm\!$1.15}
		& 2.76e4\scalebox{0.7}{$\!\pm\!$1.10e4} \\
		
		$\ell_1$ sparsity&
        67.92\scalebox{0.7}{$\!\pm\!$0.43}&
        71.50\scalebox{0.7}{$\!\pm\!$1.70}&
        80.65\scalebox{0.7}{$\!\pm\!$0.28}&
        0.86\scalebox{0.7}{$\!\pm\!$0.03}&
        0.3294\scalebox{0.7}{$\!\pm\!$0.0093} &
        52.20\scalebox{0.7}{$\!\pm\!$0.18}&
        69.40\scalebox{0.7}{$\!\pm\!$1.55}&
        9.65e3\scalebox{0.7}{$\!\pm\!$1.05e3} \\
		SalUn&
        70.95\scalebox{0.7}{$\!\pm\!$0.70}&
        73.35\scalebox{0.7}{$\!\pm\!$0.65}&
        88.15\scalebox{0.7}{$\!\pm\!$0.36}&
        0.95\scalebox{0.7}{$\!\pm\!$0.01}&
        0.9256\scalebox{0.7}{$\!\pm\!$0.0020} &
        52.05\scalebox{0.7}{$\!\pm\!$0.12}&
        92.30\scalebox{0.7}{$\!\pm\!$2.80}&
        6.02e3\scalebox{0.7}{$\!\pm\!$3.20e2} \\
		\hline
			SALA&
            71.13\scalebox{0.7}{$\!\pm\!$1.63}&
            72.33\scalebox{0.7}{$\!\pm\!$0.69}&
            88.36\scalebox{0.7}{$\!\pm\!$1.14}&
            \textbf{0.97}\scalebox{0.7}{$\!\pm\!$0.01}&
        \textbf{0.1268}\scalebox{0.7}{$\!\pm\!$0.0049} &
            \textbf{51.05}\scalebox{0.7}{$\!\pm\!$0.25}&
            \textbf{52.35}\scalebox{0.7}{$\!\pm\!$0.17}&
            3.08e3\scalebox{0.7}{$\!\pm\!$1.56e2}\\ 
			SALA-SMSE&
            71.79\scalebox{0.7}{$\!\pm\!$4.43}&
            73.92\scalebox{0.7}{$\!\pm\!$3.31}&
            88.06\scalebox{0.7}{$\!\pm\!$1.05}&
            0.94\scalebox{0.7}{$\!\pm\!$0.01}&
        0.1845\scalebox{0.7}{$\!\pm\!$0.0056} &
            52.35\scalebox{0.7}{$\!\pm\!$0.44}&
            54.33\scalebox{0.7}{$\!\pm\!$0.47}&
            \textbf{2.72e3}\scalebox{0.7}{$\!\pm\!$3.32e2}\\ 
            SALA w/ ${D}_\mathrm{f}$ only&
            71.49\scalebox{0.7}{$\!\pm\!$0.92}&
            73.31\scalebox{0.7}{$\!\pm\!$1.43}&
            88.25\scalebox{0.7}{$\!\pm\!$0.85}&
            0.95\scalebox{0.7}{$\!\pm\!$0.01}&
        0.2247\scalebox{0.7}{$\!\pm\!$0.0061} &
            52.12\scalebox{0.7}{$\!\pm\!$0.13}&
            54.42\scalebox{0.7}{$\!\pm\!$0.35}&
            {2.84e3}\scalebox{0.7}{$\!\pm\!$1.65e2}\\ 
            \hline
		\rowcolor{orange!9}
		\multicolumn{9}{c}{\textbf{WikiText-103 (GPT-2)}} \\ \hline
		Retrain&
		63.72\scalebox{0.7}{$\!\pm\!$5.47}&
		63.73\scalebox{0.7}{$\!\pm\!$5.50}&
		56.86\scalebox{0.7}{$\!\pm\!$6.14}&
		1.00\scalebox{0.7}{$\!\pm\!$0.00}&
        0.0178\scalebox{0.7}{$\!\pm\!$0.0027} &
		50.43\scalebox{0.7}{$\!\pm\!$0.13}&
		50.08\scalebox{0.7}{$\!\pm\!$0.04}&
		6.14e4\scalebox{0.7}{$\!\pm\!$4.35e3} \\ \hline
		
		FT&
		61.25\scalebox{0.7}{$\!\pm\!$0.47} &
		58.85\scalebox{0.7}{$\!\pm\!$0.53} &
		56.85\scalebox{0.7}{$\!\pm\!$0.48} &
		0.89\scalebox{0.7}{$\!\pm\!$0.02} &
        0.2496\scalebox{0.7}{$\!\pm\!$0.0052} &
		52.08\scalebox{0.7}{$\!\pm\!$0.52}&
		52.66\scalebox{0.7}{$\!\pm\!$0.11} &
		1.72e3\scalebox{0.7}{$\!\pm\!$2.15e2}\\
		
		GA&
		60.73\scalebox{0.7}{$\!\pm\!$0.44} &
		58.76\scalebox{0.7}{$\!\pm\!$0.51} &
		55.85\scalebox{0.7}{$\!\pm\!$0.45}&
		0.86\scalebox{0.7}{$\!\pm\!$0.04} &
        0.2421\scalebox{0.7}{$\!\pm\!$0.0046} &
		51.45\scalebox{0.7}{$\!\pm\!$0.13}&
		51.86\scalebox{0.7}{$\!\pm\!$0.09} &
		3.99e3\scalebox{0.7}{$\!\pm\!$3.54e2}\\
		
		RL&
		63.80\scalebox{0.7}{$\!\pm\!$4.10}&
		60.61\scalebox{0.7}{$\!\pm\!$4.28}&
		59.88\scalebox{0.7}{$\!\pm\!$3.88}&
		0.90\scalebox{0.7}{$\!\pm\!$0.03}&
        0.2783\scalebox{0.7}{$\!\pm\!$0.0061} &
		52.73\scalebox{0.7}{$\!\pm\!$0.52}&
		53.69\scalebox{0.7}{$\!\pm\!$0.63}&
		\textbf{6.02e2}\scalebox{0.7}{$\!\pm\!$9.47e1}\\
		
		$\ell_1$ sparsity&
		61.25\scalebox{0.7}{$\!\pm\!$0.47}&
		58.85\scalebox{0.7}{$\!\pm\!$0.53}&
		56.85\scalebox{0.7}{$\!\pm\!$0.48}&
		0.89\scalebox{0.7}{$\!\pm\!$0.03} &
        0.2567\scalebox{0.7}{$\!\pm\!$0.0083} &
		51.35\scalebox{0.7}{$\!\pm\!$0.43}&
		52.66\scalebox{0.7}{$\!\pm\!$0.11} &
		2.16e3\scalebox{0.7}{$\!\pm\!$1.49e2}\\
		
		SalUn&
		62.90\scalebox{0.7}{$\!\pm\!$0.49}&
		60.40\scalebox{0.7}{$\!\pm\!$0.93}&
		58.70\scalebox{0.7}{$\!\pm\!$0.85}&
		0.90\scalebox{0.7}{$\!\pm\!$0.03}&
        0.3014\scalebox{0.7}{$\!\pm\!$0.0070} &
		52.10\scalebox{0.7}{$\!\pm\!$0.20} &
		58.40\scalebox{0.7}{$\!\pm\!$1.42} &
		6.25e3\scalebox{0.7}{$\!\pm\!$3.10e2}\\ \hline
		
		SALA&
		62.18\scalebox{0.7}{$\!\pm\!$0.91}&
		61.18\scalebox{0.7}{$\!\pm\!$0.90}&
		57.24\scalebox{0.7}{$\!\pm\!$0.88}&
		0.93\scalebox{0.7}{$\!\pm\!$0.03}&
        \textbf{0.1048}\scalebox{0.7}{$\!\pm\!$0.0043} &
		\textbf{50.15}\scalebox{0.7}{$\!\pm\!$0.08} &
		\textbf{50.92}\scalebox{0.7}{$\!\pm\!$0.14} &
		{6.42e2}\scalebox{0.7}{$\!\pm\!$2.53e2}\\
		
		SALA-SMSE&
		62.70\scalebox{0.7}{$\!\pm\!$0.76} &
		60.90\scalebox{0.7}{$\!\pm\!$0.93} &
		57.40\scalebox{0.7}{$\!\pm\!$0.95}&
		0.93\scalebox{0.7}{$\!\pm\!$0.02} &
        0.1965\scalebox{0.7}{$\!\pm\!$0.0058} &
		51.20\scalebox{0.7}{$\!\pm\!$0.18} &
		51.96\scalebox{0.7}{$\!\pm\!$0.33} &
		7.90e2\scalebox{0.7}{$\!\pm\!$1.62e2}
		\\
		SALA w/ ${D}_\mathrm{f}$ only&
		63.63\scalebox{0.7}{$\!\pm\!$0.46} &
		61.29\scalebox{0.7}{$\!\pm\!$0.85} &
		58.12\scalebox{0.7}{$\!\pm\!$0.74}&
		\textbf{0.94}\scalebox{0.7}{$\!\pm\!$0.02} &
        0.2217\scalebox{0.7}{$\!\pm\!$0.0064} &
		51.72\scalebox{0.7}{$\!\pm\!$0.21} &
		51.79\scalebox{0.7}{$\!\pm\!$0.14} &
		7.69e2\scalebox{0.7}{$\!\pm\!$1.86e2}
		\\
		\bottomrule
	\end{tabular*}
        \vspace{-0.10cm}
\end{table*}

\vspace{-0.25cm}

\Needspace{15\baselineskip}
\subsection{Robustness Evaluation for SALA and Baselines}
\vspace{-0.1cm}
\label{sec:ex_robust}

Unlike Section~\ref{sec:baseline_comparison}, which reports tuned configurations, this subsection evaluates unlearning robustness under a broad learning-rate sweep. We test SalUn~\cite{fan2024salun}, SCRUB~\cite{kurmanji2023towards}, and $\ell_1$ sparsity~\cite{1modelsparisty23jia} with five log-spaced learning rates from $10^{-4}$ to $10^{-1}$ and fix the epoch budget to $10$.

\begin{figure}[!htbp]
	\centering
	\subfloat[SalUn]{
		\includegraphics[width=0.31\columnwidth]{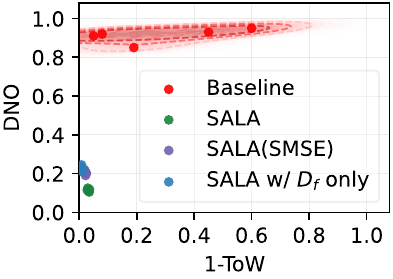}}\hfill
	\subfloat[SCRUB]{
		\includegraphics[width=0.31\columnwidth]{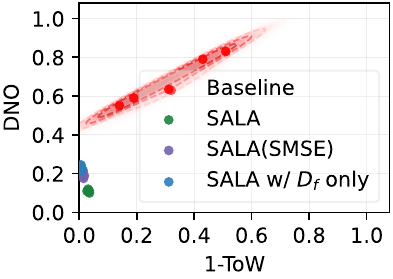}}\hfill
	\subfloat[$\ell_1$ sparsity]{
		\includegraphics[width=0.31\columnwidth]{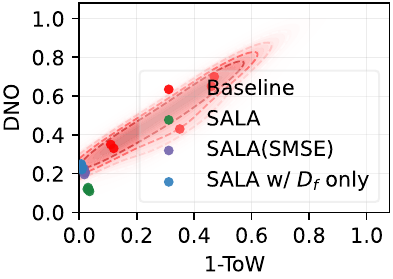}}
	\caption{Robustness under hyperparameter sweeps for three unlearning methods. SALA substantially reduces performance fluctuation.}
	\label{fig:robustness}
\end{figure}
\noindent\textbf{Results.}
Figure~\ref{fig:robustness} shows that the performance of base unlearning methods changes sharply across the sweep. SalUn has a ToW range of $0.55$, and SCRUB has a DNO range of $0.28$, which indicates unstable dataset-level and sample-level unlearning quality. SALA reduces this variation through fast-probe hyper-parameter selection and InDist-based stopping. The red clouds show kernel density estimates of the baseline evaluation points and darker regions indicate higher density. In detail, with SALA and its variants, the ToW range drops to $0.03$ for SalUn and $0.04$ for SCRUB and $\ell_1$ sparsity. Furthermore, the DNO range drops to $0.04$ for SalUn and SCRUB and $0.06$ for $\ell_1$ sparsity. Overall, SALA stabilizes unlearning quality through automatic selection among the prescribed candidate configurations.  Appendix~\ref{app:rebuttal-tuning} reports the full results.

%% file: 6_related_works.tex
\vspace{-0.28cm}
\section{Related Work}
\vspace{-0.35cm}

Exact unlearning removes designated forget data by ensuring that it never participates in training. Retraining from scratch on the retain set gives the gold standard. Sharding-based methods such as SISA~\cite{bourtoule2021machine} reduce deletion cost, but exact unlearning remains expensive at modern scale. This cost motivates approximate unlearning~\cite{golatkar2020eternal}.

Approximate unlearning keeps the pretrained model and updates it to reduce the influence of the forget set. This family can be divided into certified and empirical routes, depending on whether the update gives a formal indistinguishability guarantee with respect to exact retraining. Certified unlearning pursues this goal in parameter space and often converts unlearning error into a boundable indistinguishability gap with tools related to differential privacy~\cite{guo2019certified}. These guarantees usually require extra assumptions and calibrated noise injection, which can limit performance on large and complex models~\cite{mu2024rewind,chien2024certified,koloskova2025certified}. Since these conditions are often impractical, empirical approximate unlearning has become the default route. It supports flexible updates and can match retraining under empirical metrics, so it is widely used in large-scale settings such as large language models~\cite{yuan2024closer,entesari2025constrained}. Common designs include fine-tuning~\cite{graves2021amnesiac}, gradient ascent~\cite{thudi2022unrolling,zhang2024negative}, relabeling objectives~\cite{graves2021amnesiac,6can23chundawat}, and influence-based unlearning~\cite{wu2024scissorhands,golatkar2020eternal}. However, this flexibility leaves two practical gaps. Empirical approximate unlearning usually lacks a theory-backed upper bound on the gap to retraining, and its performance can change sharply with hyperparameters. These gaps motivate SALA, which stabilizes approximate unlearning and provides a computable upper bound on the per-sample gap to retraining.

%% file: acknowledgments.tex
\begin{ack}
This work was supported by the Ministry of Science and Technology of the People’s Republic of China (National Key Research and Development Programme, Grant No:  2025YFE0200100), the National Natural Science Foundation of China (Grant No. 62502422), Guangdong Overseas Distinguished Scholars Program (Grant No: MS202600058), and the Research Grants Council (Grant No:  15208923, 15224124, and 25207224), Hong Kong SAR, China.
\end{ack}

%% file: appendix.tex
\section{Algorithm}
\label{app:algorithm}
In Algorithm~\ref{alg:sala}, $z(x;\theta)$ denotes the shifted $z$-score in Eq.~\eqref{eq:zscore}; in Algorithm~\ref{alg:smse}, $D_A=A$ and $D_B=B$. The experimental offset clipping and scale stabilization follow Appendix~\ref{app:setup}.
\begin{algorithm}
	\caption{\textsc{SALA}: sample-grained likelihood alignment}
	\label{alg:sala}
	\begin{algorithmic}[1]
		\REQUIRE $\theta$, $D=D_\mathrm{r}\cup D_\mathrm{f}$; target distributions $(\mu_{\text{in/out}}(x),\sigma_{\text{in/out}}(x))$; unlearning algorithm $\mathcal{U}(\cdot;\texttt{arg})$;
		hyper-parameter candidates $\mathcal{K}$; $Z_{\text{cut}},\tau$; $\lambda_r,\lambda_f,\alpha$.
		\ENSURE $\theta_u$.
		
		\STATE Set $(\mu(x),\sigma(x))\gets(\mu_{\text{in}}(x),\sigma_{\text{in}}(x))$ for $x\in D_\mathrm{r}$; $(\mu_{\text{out}}(x),\sigma_{\text{out}}(x))$ for $x\in D_\mathrm{f}$
		\STATE Sample fixed $\epsilon(x)\sim\mathcal{N}(0,\mathrm{var}_\epsilon)$, set $\mu^r(x)\gets \mu(x)+\epsilon(x)\sigma(x)$
		\STATE Choose hard subset $D_\mathrm{hf}\subseteq D_\mathrm{f}$ (top-$p$ by hardness$\triangleq|\mu_\text{in}(x)-\mu_\text{out}(x)|$)
		\STATE Define $\mathrm{InDist}(S;\theta)=\mathbb{E}_{x\sim S}[\mathbf{1}(|z(x;\theta)|<Z_{\text{cut}})]$
		\STATE Store the pre-probe state $\theta^{0} \leftarrow \theta$
		\FORALL{$\mathrm{arg}\in\mathcal{K}$} 
		\STATE  Initialize a probe copy $\theta_{\mathrm{arg}} \leftarrow \theta^{0}$
		\STATE Run $T_{\mathrm{fp}}$ unlearning probe steps from $\theta_{\mathrm{arg}}$ with $\mathrm{arg}$
		\STATE $\text{score}_{\mathrm{fp}}(\mathrm{arg}) \leftarrow \sum_{S\in\{D_r,D_f,D_{hf}\}} \mathrm{InDist}(S;\theta_{\mathrm{arg}})$
		 \ENDFOR
		\STATE $\mathrm{arg}^{\star} \leftarrow \arg\max_{\mathrm{arg}\in \mathcal{K}} \text{score}_{\mathrm{fp}}(\mathrm{arg})$
		\STATE Reset $\theta \leftarrow \theta^{0}$
		\WHILE{$\exists S\in\{D_\mathrm{r},D_\mathrm{f},D_\mathrm{hf}\}$ s.t. $\mathrm{InDist}(S;\theta)<\tau$}
		\STATE sample minibatch $B\subseteq D$
		\STATE $L_{\text{SALA}}(\theta)\gets \frac{1}{|B|}\sum_{x\in B}\omega(x)\Big(\frac{s_\theta(x)-\mu^r(x)}{\sigma(x)}\Big)^2$
		\STATE where $\omega(x)=\lambda_r\mathbf{1}[x\in D_\mathrm{r}]+\lambda_f(1+\alpha\mathbf{1}[x\in D_\mathrm{hf}])\mathbf{1}[x\in D_\mathrm{f}]$
		\STATE $\theta \leftarrow U(\theta; \mathrm{arg}^{\star})$ using $(\text{unlearning loss} + L_{\mathrm{SALA}})$
		\ENDWHILE
		\STATE \textbf{return} $\theta_u\gets\theta$
	\end{algorithmic}
\end{algorithm}

\begin{algorithm}
	\caption{\textsc{SMSE}: split--merge segment ensemble estimation for member and non-member distributions}
	\label{alg:smse}
	\begin{algorithmic}[1]
		\REQUIRE Dataset $D$; training algorithm $\mathcal{A}$; warmup epoch $e_w$; merge finetune epoch $e_m$; checkpoints per branch $M$.
		\ENSURE $\{\hat\mu_{\text{in}}(x),\hat\sigma_{\text{in}}(x),\hat\mu_{\text{out}}(x),\hat\sigma_{\text{out}}(x)\}_{x\in D}$, $\theta$.
	
		\STATE Split $D$ into disjoint halves $(A,B)$; warmup $\theta_{\text{warm}}\gets\mathcal{A}(D;\text{epochs}=e_w)$
		\STATE Train branches: $\theta_A\gets\mathcal{A}(A;\text{init}=\theta_{\text{warm}})$, $\theta_B\gets\mathcal{A}(B;\text{init}=\theta_{\text{warm}})$
		\STATE Record late checkpoints $\{\theta_A^{(m)}\}_{m=1}^M$, $\{\theta_B^{(m)}\}_{m=1}^M$
		
		\FORALL{$x\in A$ (or $x\in B$)}
		\STATE let $\text{br}(x)=A$ if $x\in A$ else $B$, and $\overline{\text{br}}(x)$ be the other branch
		\STATE $u_m\gets s_{\theta_{\text{br}(x)}^{(m)}}(x)$,\; $v_m\gets s_{\theta_{\overline{\text{br}}(x)}^{(m)}}(x)$ for $m=1..M$
		\STATE $\hat\mu_{\text{in}}(x)\gets \frac{1}{M}\sum_m u_m$,\; $\hat\sigma_{\text{in}}^2(x)\gets \frac{1}{M}\sum_m (u_m-\hat\mu_{\text{in}}(x))^2$
		\STATE $\hat\mu_{\text{out}}(x)\gets \frac{1}{M}\sum_m v_m$,\; $\hat\sigma_{\text{out}}^2(x)\gets \frac{1}{M}\sum_m (v_m-\hat\mu_{\text{out}}(x))^2$
		\ENDFOR
		\STATE $\hat\sigma_{\text{in}}(x)\gets \sqrt{\hat\sigma_{\text{in}}^2(x)}$, \;
		$\hat\sigma_{\text{out}}(x)\gets \sqrt{\hat\sigma_{\text{out}}^2(x)}$ for all $x$
		\STATE merge: $\theta\gets \mathcal{A}(D;\text{init}=\text{FedAvg}(\theta_A, \theta_B),\text{epochs}=e_m)$
\STATE \hspace{1.35em} where $\text{FedAvg}(\theta_A, \theta_B)=\frac{|D_A|}{|D_A|+|D_B|}\theta_A+\frac{|D_B|}{|D_A|+|D_B|}\theta_B$

		\STATE \textbf{return} per-sample moments and $\theta$
	\end{algorithmic}
\end{algorithm}

\section{Proof of Theorem~\ref{thm:dno_bound}}
\label{app:dno_bound_proof}
\begin{proof}
	Step 1 (DNO term is total variation). For two densities $Q^{\mathrm{SALA}},Q$ on the same space,
	{\footnotesize \[
		\mathrm{DNO}=1-\int \min(Q^{\mathrm{SALA}},Q)=\frac{1}{2}\int |Q^{\mathrm{SALA}}-Q| \;=\; \mathrm{TV}(Q^{\mathrm{SALA}},Q).
		\]}
	
	Step 2 (TV bound by Bhattacharyya coefficient). The Hellinger distance satisfies
	$H^2(Q^{\mathrm{SALA}},Q)=1-\int\sqrt{Q^{\mathrm{SALA}}Q}=1-\mathrm{BC}(Q^{\mathrm{SALA}},Q)$
	and $\mathrm{TV}(Q^{\mathrm{SALA}},Q)\le \sqrt{2}\,H(Q^{\mathrm{SALA}},Q)$, hence
	\begin{equation}
		\label{eq:tv_bc}
		\footnotesize
		\mathrm{TV}(Q^{\mathrm{SALA}},Q)\le \sqrt{2\bigl(1-\mathrm{BC}(Q^{\mathrm{SALA}},Q)\bigr)}.
	\end{equation}
	
	Step 3 (closed form BC for Gaussians). For $Q=\mathcal{N}(\mu,\sigma^2)$ and
	$Q^{\mathrm{SALA}}=\mathcal{N}(\mu^{\mathrm{SALA}},(\sigma^{\mathrm{SALA}})^2)$,
	{\footnotesize\[
		\mathrm{BC}\bigl(Q^{\mathrm{SALA}},Q\bigr)
		=
		\sqrt{\frac{2\rho}{1+\rho^2}}\;
		\exp\!\left(-\frac{\Delta^2}{4(1+\rho^2)}\right),
		\]}
	with $\Delta,\rho$ defined above.
	
	Step 4 (InDist implies a feasible region for $(\Delta,\rho)$). Under $X\sim Q^{\mathrm{SALA}}$,
	{\footnotesize\[
		Y\triangleq \frac{X-\mu}{\sigma}\sim \mathcal{N}(\Delta,\rho^2).
		\]}
	Constraint~\eqref{eq:indist_constraint} becomes
	{\footnotesize\[
		\footnotesize
		\Phi\!\left(\frac{Z_{\mathrm{cut}}-\Delta}{\rho}\right)
		-\Phi\!\left(\frac{-Z_{\mathrm{cut}}-\Delta}{\rho}\right)\ \ge\ \tau.
		\]}
	A sufficient (closed-form) condition is obtained by symmetry:
	{\footnotesize\[
		\Phi\!\left(\frac{Z_{\mathrm{cut}}-|\Delta|}{\rho}\right)\ \ge\ \frac{1+\tau}{2},
		\]}
	which is equivalent to
	\begin{equation}
		\label{eq:closed_region}
		\footnotesize
		|\Delta|\ \le\ Z_{\mathrm{cut}}-\rho q,
		\qquad
		\rho\ \le\ \rho_{\max}=\frac{Z_{\mathrm{cut}}}{q},
	\end{equation}
	with $q=\Phi^{-1}((1+\tau)/2)$.
	
	Step 5 (worst case over the region). For fixed $\rho$, $\mathrm{BC}$ decreases with $|\Delta|$,
	so the smallest $\mathrm{BC}$ occurs at the boundary $|\Delta|=Z_{\mathrm{cut}}-\rho q$.
	Minimizing over $\rho\in[\rho_{\min},\rho_{\max}]$ yields~\eqref{eq:bcmin}.
	Plugging $\mathrm{BC}\ge \mathrm{BC}_{\min}$ into~\eqref{eq:tv_bc} proves~\eqref{eq:dno_x_ub}.
	
	If the same argument holds for both $Q_{\mathrm{in}}(\cdot\mid x)$ and $Q_{\mathrm{out}}(\cdot\mid x)$,
	each overlap gap is bounded by~\eqref{eq:dno_x_ub}, and their averaged DNO is bounded as well.
\end{proof}

\section{Proof of Corollary~\ref{cor:dno_bound_single}}
\label{app:dno_bound_proof_single}

By Markov's inequality,
\begin{equation}
	\footnotesize
	\mathbb P_{x\sim D_\mathrm{r}}
	\left(
	\left|
	\frac{s_\theta(x)-s_{\theta_0}(x)}{\sigma_{\mathrm{in}}(x)}
	\right|
	\ge Z_{\mathrm{cut}}
	\right)
	\le
	\frac{L_r(\theta)}{Z_{\mathrm{cut}}^2}
	\le
	1-\tau.
\end{equation}
Equivalently,
\begin{equation}
	\footnotesize
	\mathbb P_{x\sim D_\mathrm{r}}
	\left(
	\left|
	\frac{s_\theta(x)-s_{\theta_0}(x)}{\sigma_{\mathrm{in}}(x)}
	\right|
	<
	Z_{\mathrm{cut}}
	\right)
	\ge \tau.
\end{equation}
Hence, the retain-side condition required by the DNO analysis is recovered without explicitly applying SALA to $D_\mathrm{r}$. Therefore, the retain-side coverage requirement in Theorem~\ref{thm:dno_bound} can be equivalently replaced by the condition in~\eqref{eq:retain_consistency_loss}, and the same DNO upper bound continues to hold.

\section{Explicit Bound of the Estimation Error}
\label{app:smse_error_bound}
\paragraph{Bounding the estimation term $\mathrm{DNO}_{\mathrm{est}}$.} $\mathrm{DNO}_{\mathrm{est}}$  is bounded by the Bhattacharyya coefficient:
\begin{equation}
	\label{eq:tv_bc_est}
	\footnotesize
	\mathrm{DNO}_{\mathrm{est}}=\mathrm{TV}\!\bigl(Q,\hat Q\bigr)
	\ \le\
	\sqrt{2\bigl(1-\mathrm{BC}_{\min}(Q,\hat Q)\bigr)},
\end{equation}

Since $\mathrm{BC}$ decreases with $|\mu-\hat\mu|$, the worst case of $|\mu-\hat\mu|$is defined as $\varepsilon_\mu$. Let $\rho\triangleq \frac{\sigma}{\hat\sigma}$ and $\Delta_{\max}\triangleq \frac{\varepsilon_\mu}{\hat\sigma}$. Then $\rho\in[\rho_L,\rho_U]$ with $\rho_L=\frac{\sigma_L}{\hat\sigma}$ and $\rho_U=\frac{\sigma_U}{\hat\sigma}$, and
\begin{equation}
	\label{eq:bcmin_est_1d}
	\footnotesize
	\mathrm{BC}_{\min}(Q,\hat Q)
	=
	\min_{\rho\in[\rho_L,\rho_U]}
	\left[
	\sqrt{\frac{2\rho}{1+\rho^2}}\;
	\exp\!\left(
	-\frac{\Delta_{\max}^2}{4(1+\rho^2)}
	\right)
	\right].
\end{equation}
Combining~\eqref{eq:tv_bc_est}--\eqref{eq:bcmin_est_1d} gives an explicit bound for $\mathrm{DNO}_{\mathrm{est}}$.

\section{Supplementary Experimental Setups}
\label{app:setup}

\paragraph{Dataset and model training.}
Training follows common setups in prior unlearning work on three benchmarks of increasing scale: CIFAR-10 ($50{,}000$ training images) with ResNet-18, ImageNet-1000 ($1{,}281{,}167$ training images) with ResNet-50, and WikiText-103 ($1{,}165{,}029$ training sequences) with GPT-2. CIFAR-10 trains for $150$ epochs with SGD and learning rate $0.1$. ImageNet-1000 trains for $90$ epochs with SGD and learning rate $0.1$. WikiText-103 trains for $22$ epochs with AdamW and learning rate $5\times10^{-4}$.

\paragraph{Unlearning scenarios and evaluation protocol.}
We evaluate three unlearning scenarios. In random sample-wise unlearning, the forget set $D_\mathrm{f}$ contains individual training examples sampled across classes~\cite{bourtoule2021machine}. In class-wise unlearning, $D_\mathrm{f}$ contains all training samples from one or more target classes~\cite{fan2024salun,tarun2023fast}. In sequence unlearning, $D_\mathrm{f}$ contains contiguous target token sequences from the corpus~\cite{naderloui2025rectifying}. For each random sample-wise unlearning task, we construct task-specific held-out, forget, and retain subsets from the original training split. 

On CIFAR-10, we reserve 20\% of the training samples as the task-level held-out subset and 20\% as the forget subset, with the remaining 60\% used as the retain subset. 
On ImageNet-1000 and WikiText-103, we reserve 33\% of the training samples or sequences as the task-level held-out subset and 33\% as the forget subset, with the remaining 34\% used as the retain subset. 
These proportions provide observations in both membership roles over repeated tasks for estimating the per-sample score distributions.
For class-wise unlearning, the forget set contains one entire target class, and the retain set contains all remaining classes. DNO is computed over $64$ tasks on CIFAR-10 and $16$ tasks on ImageNet-1000 and WikiText-103. 

\paragraph{Baseline coverage.}
We compare SALA with direct update-based baselines and optimized approximate unlearning (AU) frameworks. The full set of $12$ baselines is evaluated on CIFAR-10. Due to memory constraints, ImageNet-1000 and WikiText-103 use the five representative baselines reported in Table~\ref{tab:baseline_comparison}. All baselines use tuned hyperparameters to maximize their own performance under the same evaluation protocol.

\paragraph{Unlearning baselines.}
The evaluated baselines are configured as follows.

\begin{itemize}
	\item \textbf{Fine-tuning (FT)}~\cite{1modelsparisty23jia}. 
	The pretrained model $f_{\theta}$ is further trained only on the retain set $D_\mathrm{r}$. Forgetting emerges from catastrophic forgetting because samples in $D_\mathrm{f}$ are no longer reinforced. On CIFAR-10, we use learning rate $0.04$ for $10$ epochs. On ImageNet-1000, we use learning rate $10^{-4}$ for $15$ epochs. On WikiText-103, we use AdamW with learning rate $5\times10^{-5}$ and max steps $10{,}000$.
	
	\item \textbf{Gradient Ascent (GA)}~\cite{thudi2022unrolling,graves2021amnesiac}. 
	GA approximates unlearning by ascending the loss on $D_\mathrm{f}$, which reverses part of the optimization signal induced by the deleted samples. On CIFAR-10, we use learning rate $0.04$ for $8$ epochs, with retain and forget weights $0.69$ and $0.31$, $\lambda=20$, and gradient clipping threshold $5$. On ImageNet-1000, we use learning rate $10^{-4}$ for $2$ epochs. On WikiText-103, we use AdamW with learning rate $5\times10^{-5}$ and max steps $10{,}000$, with $\alpha=0.85$, $\ell_2$ weight decay $0.02$, and forget threshold $\tau_\mathrm{f}=6.0$.
	
	\item \textbf{Random Label (RL)}~\cite{graves2021amnesiac,6can23chundawat}. 
	RL randomizes the labels of $D_\mathrm{f}$ and trains the model on corrupted targets, which disrupts the association between forget samples and their original labels. On CIFAR-10, we optimize the relabeling objective with learning rate $1.5\times10^{-4}$ for $8$ epochs. On ImageNet-1000, we use Adam with learning rate $5\times10^{-3}$ for $2$ epochs. On WikiText-103, we use AdamW with learning rate $5\times10^{-5}$ and max steps $10{,}000$, with amnesiac weight $0.35$.
	
	\item \textbf{Influence Unlearning (IU)}~\cite{1modelsparisty23jia}. 
	IU uses WoodFisher-style influence estimation to approximate how $D_\mathrm{f}$ contributes to the learned parameters, and then applies a compensatory parameter perturbation to remove this contribution from $\theta$. On CIFAR-10, we instantiate IU with the WoodFisher variant and set the WoodFisher coefficient to $\alpha=20\sim50$.
	
	\item \textbf{FisherForgetting (FF)}~\cite{golatkar2020eternal}. 
	FF perturbs parameters with Gaussian noise whose covariance is derived from the fourth root of the Fisher information matrix. This perturbation scrubs information associated with the forgotten data while approximately preserving retained behavior. On CIFAR-10, we set the noise-scale parameter to $\epsilon=7\times10^{-8}$.
	
	\item \textbf{MinNorm-OG}~\cite{jacob2025machine}. 
	MinNorm-OG solves for a minimum-norm parameter perturbation under an orthogonal-gradient constraint on $D_\mathrm{r}$, followed by descent steps on the retain loss. On CIFAR-10, the retain-loss descent uses learning rate $0.001$ for $5$ epochs, and the regularization weight is $\alpha=1.8$ in the minimum-norm perturbation subproblem.
	
	\item \textbf{SEMU}~\cite{marcin2025semu}. 
	SEMU uses truncated singular value decomposition of forget-set gradients to identify a low-dimensional critical subspace of weights, and restricts unlearning updates to this compact projection. On CIFAR-10, the fine-tuning stage uses learning rate $0.01$ for $3$ epochs, and the truncated-rank hyperparameter defaults to $5$.
	
	\item \textbf{$\ell_1$ sparsity}~\cite{1modelsparisty23jia}. 
	This method adds an $\ell_1$ penalty to retain-set fine-tuning, so the unlearning trajectory favors sparse parameter changes. On CIFAR-10, we use learning rate $0.026$ for $9$ epochs and disable data augmentation, with $\ell_1$ penalty weight $5\times10^{-5}$. On ImageNet-1000, we use learning rate $10^{-4}$ for $15$ epochs. On WikiText-103, we use AdamW with learning rate $5\times10^{-5}$ and max steps $10{,}000$, with $\ell_1$ penalty weight $2\times10^{-5}$.
	
	\item \textbf{SalUn}~\cite{fan2024salun}. 
	SalUn uses a gradient-based saliency mask to localize updates on parameters that are most relevant to the forgetting objective. On CIFAR-10, we run saliency-guided unlearning with learning rate $9\times10^{-4}$ for $14$ epochs and use a precomputed saliency mask with ratio $0.2$. On ImageNet-1000, we use Adam with learning rate $5\times10^{-3}$ for $2$ epochs. On WikiText-103, we use AdamW with learning rate $5\times10^{-5}$ and max steps $10{,}000$.
	
	\item \textbf{SCRUB}~\cite{kurmanji2023towards}. 
	SCRUB uses a teacher--student distillation objective. The student deviates from the teacher on $D_\mathrm{f}$ and stays close to the teacher on $D_\mathrm{r}$. On CIFAR-10, we run stochastic gradient ascent--descent with learning rate $1.5\times10^{-4}$ for $8$ epochs.
	
	\item \textbf{SFRon}~\cite{huang2024unified}. 
	SFRon decomposes gradient-based unlearning into weighted forgetting gradient ascent, retaining gradient descent, and saliency-guided weighting. It further uses remain-geometry information to modulate the update trajectory. On CIFAR-10, we use learning rate $0.0035$ for $5$ epochs and set the forgetting-update frequency to $1$.
	
	\item \textbf{Refined-Unlearning Meta-algorithm (RUM)}~\cite{zhao2024what}. 
	RUM first refines the forget set into more homogeneous subsets and then applies different unlearning strategies across these subsets. On CIFAR-10, we instantiate the pipeline with a fine-tuning stage at learning rate $0.1$ followed by a SalUn stage at learning rate $5\times10^{-4}$. We use $5$ fine-tuning epochs and $3$ SalUn epochs.
\end{itemize}

\paragraph{Supplementary setups of SALA and SMSE.}
We provide the default hyperparameters used for SALA and SMSE. Unless otherwise specified, the main experiments pair SALA with SalUn because SALA acts as an auxiliary objective for iterative unlearning. Additional base unlearning methods are evaluated in Appendix~\ref{app:other_backbones}.

\textbf{SALA setup.}
We evaluate three variants: SALA with shadow-model targets, SALA-SMSE with SMSE-estimated targets, and SALA w/ $D_\mathrm{f}$ only, which constrains only the forget set. Unless otherwise specified, we set $\lambda_f=1.0$, $\lambda_r=0.2$, and $\alpha=2.0$. We use $Z_{\mathrm{cut}}=2.0$, $\tau=0.95$, $\epsilon_{\mathrm{scale}}=1.0$, and $\epsilon_{\mathrm{clip}}=Z_{\mathrm{cut}}-0.2=1.8$. These settings provide stable guidance and keep the target perturbation within the theoretically motivated range. We ablate SALA weights in Appendix~\ref{sec:ablation_weights}, and we ablate $\tau$ and $Z_{\mathrm{cut}}$ in Appendix~\ref{sec:ablation_sala_threshold}.

\textbf{SMSE setup.}
SMSE allocates $20\%$ of the total epochs to warmup, $70\%$ to half-data training, and $10\%$ to fine-tuning after merging. We use segment ensemble size $M=30$. Checkpoints are recorded once per epoch on CIFAR-10 and ImageNet-1000, and eight times per epoch on WikiText-103. We ablate SMSE epoch allocation, segment ensemble size, and stochastic estimation method in Appendix~\ref{sec:ablation_smse_schedule}, Appendix~\ref{sec:ablation_smse_M}, and Appendix~\ref{sec:ablation_smse_estimation_method}.

\textbf{SMSE setup with $\hat{\sigma}$ stabilization.}
To tighten the adjusted DNO bound in~\eqref{eq:tv_triangle}, SMSE reduces the estimation term $\mathrm{DNO}_{\mathrm{est}}$ by stabilizing the estimated scale $\hat{\sigma}(x)$. The Gaussian-form analysis in Appendix~\ref{app:smse_error_bound} shows that $\mathrm{DNO}_{\mathrm{est}}$ increases when $\hat{\sigma}(x)$ deviates too far from a plausible range of the true scale $\sigma(x)$; see Eq.~\eqref{eq:bcmin_est_1d}. We therefore shrink each per-sample estimate toward a global scale and clip it to a safe interval:
\begin{equation}
	\label{eq:sigma_clip}
	\hat{\sigma}(x)\leftarrow
	\mathrm{clip}\!\Bigl(
	\lambda_{\sigma}\,\hat{\sigma}(x)
	+(1-\lambda_{\sigma})\,\bar{\sigma},\
	\sigma_{\min},\ \sigma_{\max}
	\Bigr),
\end{equation}
where $\lambda_{\sigma}\in[0,1]$ controls the shrinkage strength, $\bar{\sigma}$ is the global average of $\hat{\sigma}(x)$ over the training set, and $(\sigma_{\min},\sigma_{\max})$ prevent unrealistically small or large scales. In all experiments, we set the shrinkage coefficient to \(\lambda_\sigma=0.8\), and clip the stabilized scale to \([\sigma_{\min},\sigma_{\max}]=[1\mathrm{e}-2,5]\). This correction reduces variance in per-sample scale estimates and yields a tighter adjusted DNO estimate. We ablate $\hat{\sigma}$ stabilization in Appendix~\ref{app:hat_sigma}.

\paragraph{Implementation.}
CIFAR-10 runs on 2$\times$ RTX 4090 GPUs. ImageNet-1000 and WikiText-103 run on 8$\times$ RTX 4090 GPUs. All reported runtimes are measured under these hardware settings. 

\paragraph{Hyperparameter selection.}
Each SALA variant evaluates ten candidate configurations using the fast probe and selects by InDist. For SalUn, the learning rates are $\{10^{-4},4\times10^{-4},7\times10^{-4},10^{-3},1.5\times10^{-3},2.5\times10^{-3},4\times10^{-3},6\times10^{-3},8\times10^{-3},10^{-2}\}$. For $\ell_1$ sparsity, the pool combines $\{10^{-4},10^{-3},2.5\times10^{-3},4\times10^{-3},10^{-2}\}$ with regularization weights $\{5\times10^{-5},10^{-4}\}$. SCRUB uses the same five learning rates with teacher-KL weights $\{0.1,0.3\}$. Baselines use method-specific manual searches covering at least $10^{-4}$--$10^{-2}$ when learning rate is tuned, with approximately ten full runs per method and selection favoring the largest ToW at a practical runtime. SALA's search budget consists of short probes; baseline tuning uses full unlearning runs.

\section{Supplementary Experiments}
\begingroup

\input{supplemental_experiments}
\FloatBarrier
\endgroup

\Needspace{5\baselineskip}
\section{Limitations and Broader Impact}
\label{app:limitation}
\paragraph{Limitations.} 
The main limitation of SALA is its dependence on per-sample target distributions, namely $Q_{\mathrm{in}}(\cdot\mid x)$ and $Q_{\mathrm{out}}(\cdot\mid x)$. When memorization statistics or shadow-model estimates have already been collected for a task or dataset, SALA can directly reuse them. Otherwise, these target distributions must be obtained through the proposed SMSE procedure, which requires a split--merge training schedule and checkpoint ensembles during model development. Thus, SALA imposes requirements on the training pipeline and is less convenient for models that have already been trained without saving the necessary statistics, though SMSE reduces this preparation cost to roughly one structured training run.

\paragraph{Broader impact.} 
This additional training-time requirement is worthwhile because it enables approximate unlearning to be monitored and controlled at the individual-sample level. SALA helps reveal and reduce sample-level gap-to-retrain that can be hidden by dataset-level averaged metrics, thereby reducing privacy leakage after unlearning. More broadly, this work supports more reliable data deletion workflows under right-to-be-forgotten requirements. We expect SALA to help practitioners move from average-case unlearning evaluation towards privacy-preserving individuals and verifiable unlearning systems.

%% file: supplemental_experiments.tex
\raggedbottom
\setlength{\intextsep}{8pt plus 2pt minus 2pt}
\setlength{\textfloatsep}{12pt plus 2pt minus 2pt}
\setlength{\floatsep}{10pt plus 2pt minus 2pt}
\captionsetup[table]{skip=7pt}
\makeatletter
\setlength{\@fptop}{0pt}
\setlength{\@fpsep}{12pt}
\setlength{\@fpbot}{0pt plus 1fil}
\makeatother

\subsection{Empirical Validation of the Gaussian Approximation}
\label{app:rebuttal-gaussian}
To empirically evaluate the Gaussian approximation of the per-sample reference distributions in Section~\ref{sec:sala_framework}, we examine how closely the scores follow a Gaussian distribution under each membership role. We evaluate log-odds on CIFAR-10 and ImageNet-1000 and per-token cross-entropy on WikiText-103, using the models and training settings in Appendix~\ref{app:setup}. 

\noindent\textbf{Setup.} This experiment uses $64$, $32$, and $32$ models for CIFAR-10, ImageNet-1000, and WikiText-103, respectively. For each sample, we collect its scores across models separately for member and non-member roles, fit a Gaussian using the sample mean and standard deviation, and measure the fit using the Kolmogorov--Smirnov distance
\begin{equation}
D_{\mathrm{gauss}}=\sup_z\left|F_n(z)-\Phi\!\left(\frac{z-\hat\mu}{\hat\sigma}\right)\right|,
\end{equation}
where $F_n$ is the empirical CDF, $\Phi$ is the standard Gaussian CDF, $\hat\mu$ is the sample mean, and $\hat\sigma$ is the sample standard deviation with one degree-of-freedom correction. A smaller $D_{\mathrm{gauss}}$ indicates a closer fit. To interpret the distance at finite sample sizes, we also draw observations from a Gaussian distribution and apply the same fitting and distance computation.

\begin{table}[!htb]
\centering
\scriptsize
\setlength{\tabcolsep}{3.8pt}
\renewcommand{\arraystretch}{1.05}
\caption{Gaussian-fit evaluation of scalar scores. The last column gives the mean distance for the finite-sample Gaussian reference.}
\label{tab:rebuttal-gaussian}
\label{tab:rebuttal-score-choice}
\begin{tabular*}{\linewidth}{@{\extracolsep{\fill}}lllccc@{}}
\toprule
\textbf{Dataset} & \textbf{Score} & \textbf{Role} & \textbf{Mean $D_{\mathrm{gauss}}$} & \textbf{Median $D_{\mathrm{gauss}}$} & \shortstack{\textbf{Gaussian} \textbf{reference}} \\
\midrule
\multirow{6}{*}{CIFAR-10} & Log-odds & Member & \textbf{0.1098} & \textbf{0.1059} & \multirow{6}{*}{0.1139} \\
 & Log-odds & Non-member & \textbf{0.1162} & \textbf{0.1116} &  \\
 & Probability & Member & 0.2646 & 0.2606 &  \\
 & Probability & Non-member & 0.2919 & 0.2888 &  \\
 & Probability margin & Member & 0.2714 & 0.2680 &  \\
 & Probability margin & Non-member & 0.2890 & 0.2862 &  \\
\midrule
\multirow{6}{*}{ImageNet-1000} & Log-odds & Member & \textbf{0.1514} & \textbf{0.1457} & \multirow{6}{*}{0.1651} \\
 & Log-odds & Non-member & \textbf{0.1524} & \textbf{0.1461} &  \\
 & Probability & Member & 0.2326 & 0.2225 &  \\
 & Probability & Non-member & 0.2367 & 0.2269 &  \\
 & Probability margin & Member & 0.2330 & 0.2217 &  \\
 & Probability margin & Non-member & 0.2282 & 0.2157 &  \\
\midrule
\multirow{2}{*}{WikiText-103} & Cross-entropy & Member & \textbf{0.1629} & \textbf{0.1532} & \multirow{2}{*}{0.1651} \\
 & Cross-entropy & Non-member & \textbf{0.1690} & \textbf{0.1554} &  \\
\bottomrule
\end{tabular*}
\end{table}

\noindent\textbf{Results}. Table~\ref{tab:rebuttal-gaussian} compares the scores used by SALA with the finite-sample Gaussian references and alternative classification scores (\emph{i.e.}, the probability and the probability margin). It indicates that the log-odds and cross-entropy distances are close to their respective Gaussian references. These results empirically support approximating the per-sample log-odds and cross-entropy distributions as Gaussian.

\Needspace{4\baselineskip}
\subsection{Supplementary experimental results of Section~\ref{sec:ex_robust}}
\label{app:rebuttal-tuning}
\label{app:rebuttal-robustness}
To quantify the robustness observed in Section~\ref{sec:ex_robust}, Table~\ref{tab:rebuttal-robustness} reports the detailed results of $1-\mathrm{ToW}$ and DNO across the five settings. The SALA candidate pools and selection protocol are given in Appendix~\ref{app:setup}. For all three baselines, adding SALA or its variants reduces the standard deviation of both metrics, indicating more stable unlearning quality.

\begin{table}[!htb]
\centering
\scriptsize
\setlength{\tabcolsep}{3.5pt}
\renewcommand{\arraystretch}{1.05}
\caption{Robustness statistics for Section~\ref{sec:ex_robust}.}
\label{tab:rebuttal-robustness}
\begin{tabular*}{\linewidth}{@{\extracolsep{\fill}}llcccc@{}}
\toprule
\textbf{Method} & \textbf{Configuration} & \textbf{$1-\mathrm{ToW}$} & \textbf{95\% interval} & \textbf{DNO} & \textbf{95\% interval} \\
\midrule
\multirow{4}{*}{\textbf{SalUn}} & No SALA & 0.2740\scalebox{0.7}{$\!\pm\!$0.2409} & [-0.0251, 0.5731] & 0.9120\scalebox{0.7}{$\!\pm\!$0.0377} & [0.8652, 0.9588] \\
 & SALA & 0.0322\scalebox{0.7}{$\!\pm\!$0.0033} & [0.0280, 0.0364] & 0.1146\scalebox{0.7}{$\!\pm\!$0.0076} & [0.1051, 0.1241] \\
 & SALA-SMSE & 0.0218\scalebox{0.7}{$\!\pm\!$0.0029} & [0.0182, 0.0254] & 0.2060\scalebox{0.7}{$\!\pm\!$0.0119} & [0.1912, 0.2208] \\
 & SALA w/ $D_\mathrm{f}$ only & 0.0100\scalebox{0.7}{$\!\pm\!$0.0016} & [0.0080, 0.0120] & 0.2270\scalebox{0.7}{$\!\pm\!$0.0135} & [0.2102, 0.2438] \\
\midrule
\multirow{4}{*}{\textbf{SCRUB}} & No SALA & 0.3180\scalebox{0.7}{$\!\pm\!$0.1561} & [0.1242, 0.5118] & 0.6780\scalebox{0.7}{$\!\pm\!$0.1246} & [0.5233, 0.8327] \\
 & SALA & 0.0322\scalebox{0.7}{$\!\pm\!$0.0036} & [0.0278, 0.0366] & 0.1094\scalebox{0.7}{$\!\pm\!$0.0075} & [0.1000, 0.1188] \\
 & SALA-SMSE & 0.0150\scalebox{0.7}{$\!\pm\!$0.0022} & [0.0122, 0.0178] & 0.1940\scalebox{0.7}{$\!\pm\!$0.0160} & [0.1742, 0.2138] \\
 & SALA w/ $D_\mathrm{f}$ only & 0.0080\scalebox{0.7}{$\!\pm\!$0.0016} & [0.0060, 0.0100] & 0.2290\scalebox{0.7}{$\!\pm\!$0.0129} & [0.2129, 0.2451] \\
\midrule
\multirow{4}{*}{\textbf{$\ell_1$ sparsity}} & No SALA & 0.3020\scalebox{0.7}{$\!\pm\!$0.1771} & [0.0821, 0.5219] & 0.4920\scalebox{0.7}{$\!\pm\!$0.1721} & [0.2783, 0.7057] \\
 & SALA & 0.0322\scalebox{0.7}{$\!\pm\!$0.0030} & [0.0284, 0.0360] & 0.1178\scalebox{0.7}{$\!\pm\!$0.0073} & [0.1087, 0.1269] \\
 & SALA-SMSE & 0.0180\scalebox{0.7}{$\!\pm\!$0.0022} & [0.0152, 0.0208] & 0.2110\scalebox{0.7}{$\!\pm\!$0.0119} & [0.1962, 0.2258] \\
 & SALA w/ $D_\mathrm{f}$ only & 0.0090\scalebox{0.7}{$\!\pm\!$0.0016} & [0.0070, 0.0110] & 0.2340\scalebox{0.7}{$\!\pm\!$0.0129} & [0.2179, 0.2501] \\
\bottomrule
\end{tabular*}
\end{table}

\subsection{Ablation Study of SALA}
\label{sec:indist_ablation}

\Needspace{4\baselineskip}
\subsubsection{SALA weights}
\label{sec:ablation_weights}
SALA introduces three weighting hyperparameters: $\lambda_f$ (forget guidance weight), $\lambda_r$ (retain guidance weight), and $\alpha$ (extra emphasis on hard-forget samples). In the default pipeline, early stopping is triggered by the InDist threshold, so these weights do not change the stopping \emph{criterion} itself; instead, they mainly control (i) how quickly InDist reaches the threshold, and (ii) the trade-off between retaining and forgetting during the transient phase before termination. This ablation quantifies how sensitive the final outcome is to these weights and whether SALA remains stable across a broad, practical range.

\paragraph{Setup.}
A grid search is conducted with
$\lambda_f\in\{0.3,\,1.0,\,2.0\}$,
$\lambda_r\in\{0.2,\,0.5,\,1.0\}$, and
$\alpha\in\{0.0,\,1.0,\,2.0\}$.
For each $\alpha$, a $3{\times}3$ heatmap is plotted over $(\lambda_f,\lambda_r)$.
Three rows are reported: the first row is ToW (higher is better), the second row is DNO (lower is better), and the third row is runtime (lower is better). All other settings follow Section~\ref{sec:ex_setup}, and results are averaged over the same unlearning task suite.

\begin{figure}[!htb]
	\centering
	\subfloat[ToW($\alpha=0$)]{
	\includegraphics[width=0.27\textwidth]{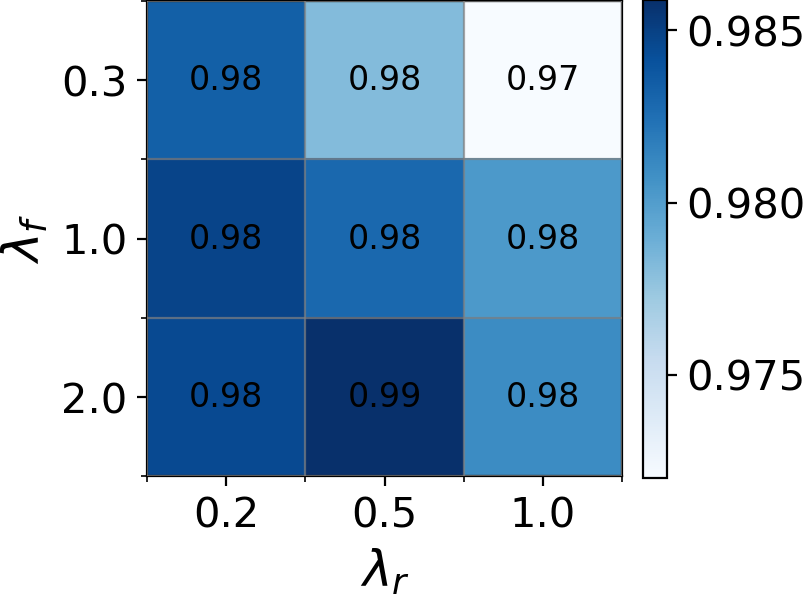}}
	\subfloat[ToW($\alpha=1.0$)]{
	\includegraphics[width=0.27\textwidth]{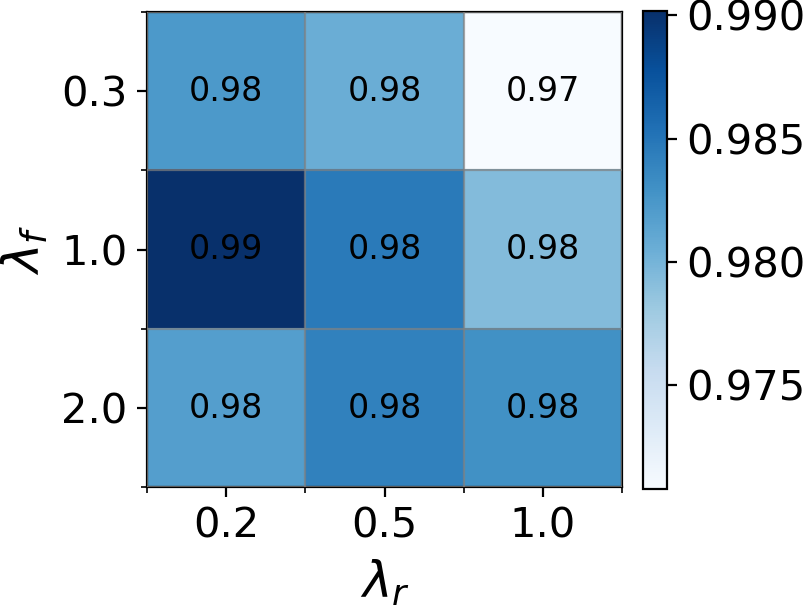}}
	\subfloat[ToW($\alpha=2.0$)]{
	\includegraphics[width=0.27\textwidth]{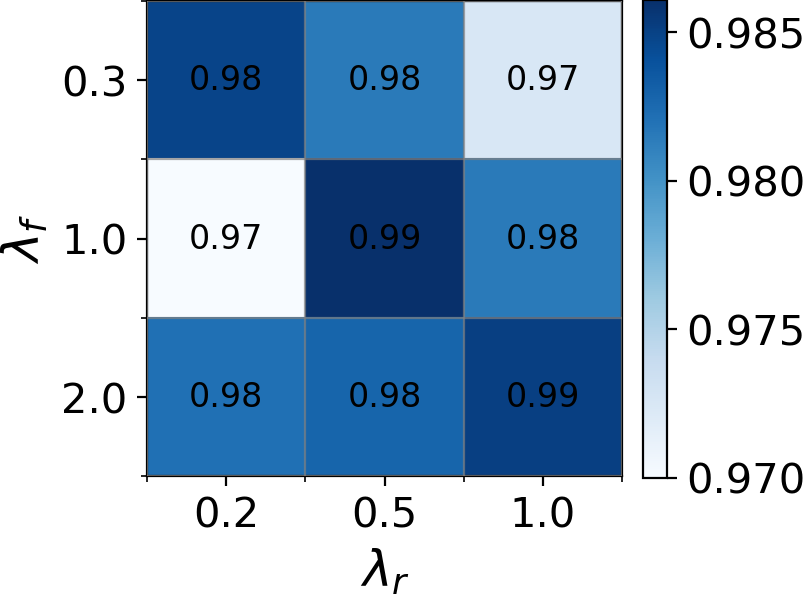}}\\
	\subfloat[DNO($\alpha=0$)]{
		\includegraphics[width=0.26325\textwidth]{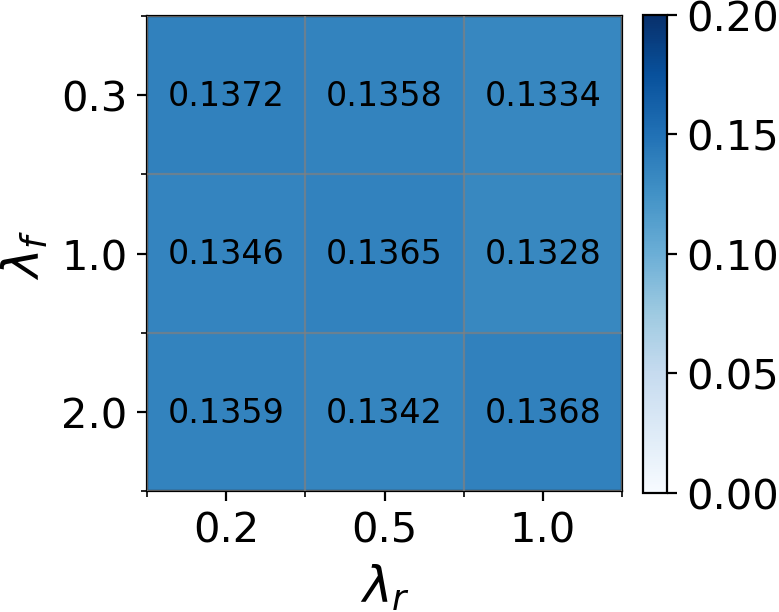}}
	\subfloat[DNO($\alpha=1.0$)]{
		\includegraphics[width=0.26325\textwidth]{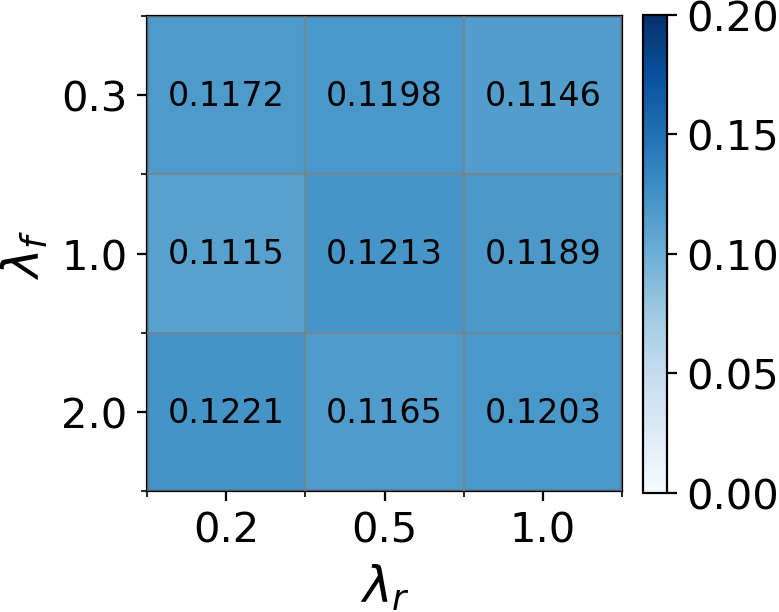}}
	\subfloat[DNO($\alpha=2.0$)]{
		\includegraphics[width=0.26325\textwidth]{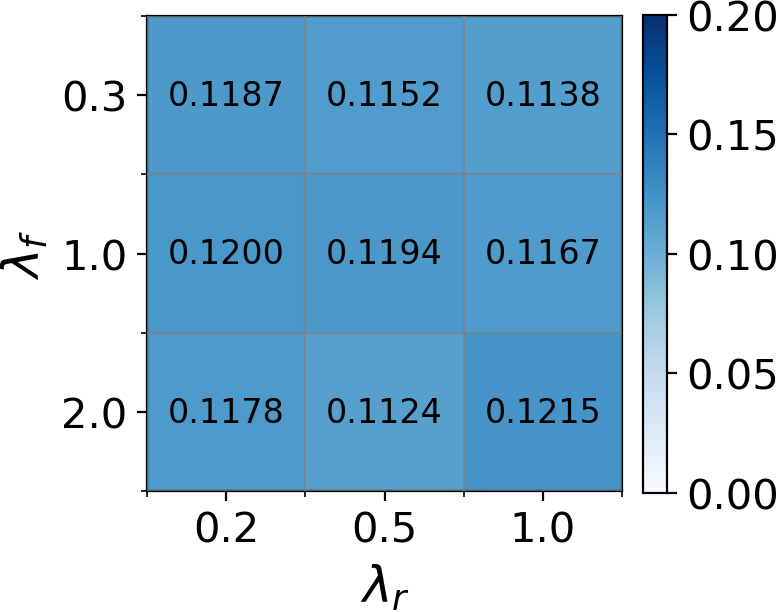}}\\
	\subfloat[Time($\alpha=0$)]{
		\includegraphics[width=0.25785\textwidth]{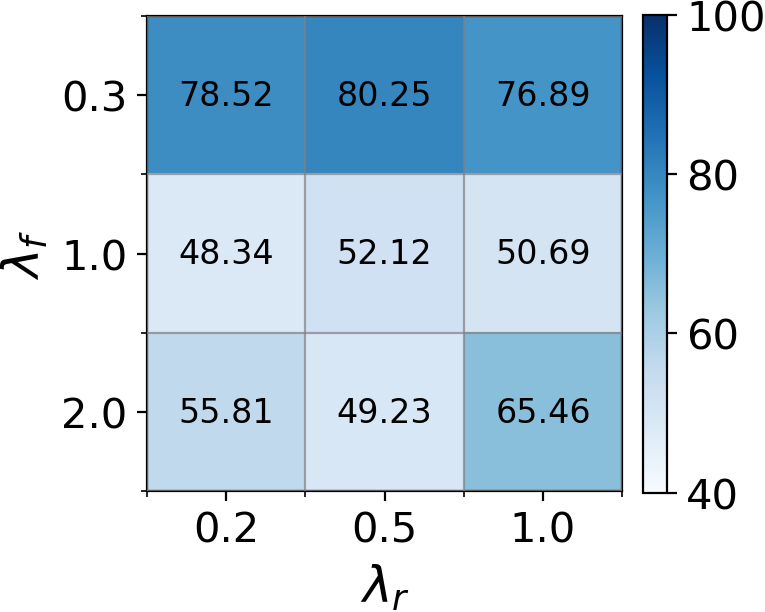}}
	\subfloat[Time($\alpha=1.0$)]{
		\includegraphics[width=0.25785\textwidth]{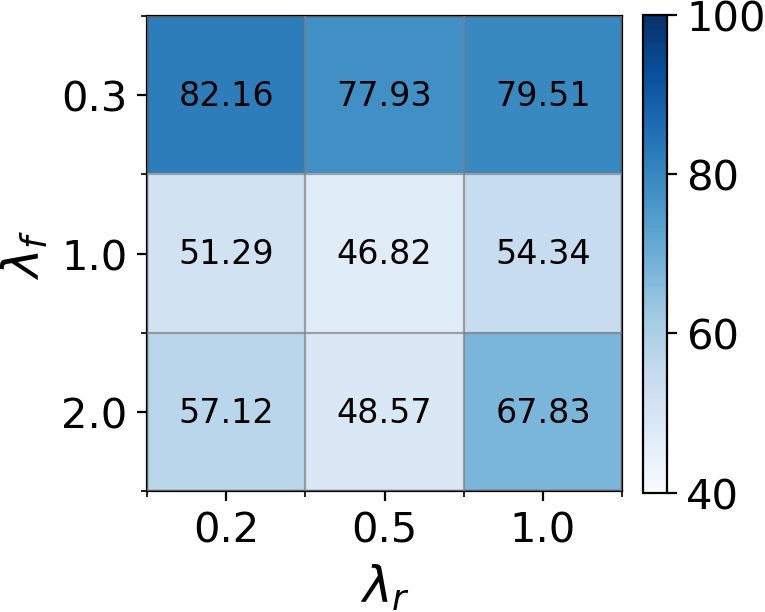}}
	\subfloat[Time($\alpha=2.0$)]{
		\includegraphics[width=0.25785\textwidth]{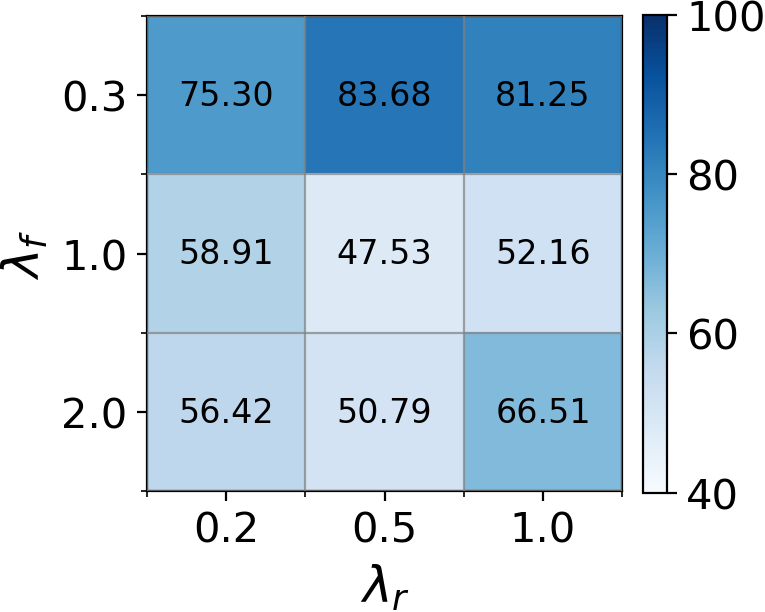}}
	\caption{Ablation on SALA weights. Columns correspond to $\alpha$. Within each panel, the heatmap axes are $\lambda_f$ (y-axis) and $\lambda_r$ (x-axis). Rows report $\mathrm{ToW}$, DNO, and runtime, respectively.}
	\label{fig:ablation_weights}
\end{figure}

\paragraph{Results.}
Figure~\ref{fig:ablation_weights} shows that SALA is stable across a broad range of weighting choices. Across all values of $\alpha$, ToW remains consistently high, mostly between 0.97 and 0.99, while DNO stays within a narrow range of approximately 0.11 to 0.14. This indicates that the final unlearning quality is not highly sensitive to moderate changes in $\lambda_f$, $\lambda_r$, or $\alpha$ once InDist-based stopping is used. The main variation appears in runtime mainly with $\lambda_f$ and $\lambda_r$: smaller weights tend to slow convergence to about 75 to 84 seconds, while stronger guidance reduces it to roughly 47 to 58 seconds.

\Needspace{4\baselineskip}
\subsubsection{Stop threshold $\tau$ and $Z_{\mathrm{cut}}$ of SALA}
\label{sec:ablation_sala_threshold}

We further examine the sensitivity of SALA to the InDist stopping threshold $\tau$ and the cutoff $Z_{\mathrm{cut}}$. Overall, Table~\ref{tab:param_sweep_z_threshold} shows that SALA remains stable across the tested stopping-threshold and cutoff settings. For the full SALA variant, DNO is consistently low at 0.12--0.13. The main effect of the sweep is on runtime rather than final unlearning quality. A smaller cutoff, $Z_{\mathrm{cut}}=1$, leads to substantially longer runtime, while $Z_{\mathrm{cut}}=2$ or $3$ reaches the stopping condition much faster. Increasing $\tau$ from 0.95 to 0.99 generally makes stopping stricter and increases runtime. The default setting, $Z_{\mathrm{cut}}=2$ and $\tau=0.95$, achieves a favorable balance, with low DNO, competitive leakage metrics, and short runtime.

\begin{table}[!htb]
	\centering
	\scriptsize
	\setlength{\tabcolsep}{3.11pt}
	\renewcommand{\arraystretch}{1.15}
	\caption{Parameter sweep over $Z_{\mathrm{cut}}$ and inDist threshold $\tau$ for SALA variants.}
	\begin{tabular}{lcccccccccc}
		\toprule
		\textbf{Method} &
		\boldmath$Z_{\mathrm{cut}}$ &
		\boldmath$\tau$ &
		\textbf{TA(\%)} &
		\textbf{FA(\%)} &
		\textbf{RA(\%)} &
		\textbf{ToW}$\uparrow$ &
		\textbf{DNO}$\downarrow$ &
		\textbf{MIA efficacy}(\%)$\downarrow$ &
		\textbf{RULI}(\%)$\downarrow$ &
		\textbf{Time}(s)$\downarrow$ \\
		\midrule
		\multirow{6}{*}{\textbf{SALA}}
		& 1 & 0.95 & 94.73\scalebox{0.7}{$\!\pm\!$0.23} &
		95.59\scalebox{0.7}{$\!\pm\!$0.20} &
		100.00\scalebox{0.7}{$\!\pm\!$0.00} &
		0.97\scalebox{0.7}{$\!\pm\!$0.01}  & 0.13 & 52.13\scalebox{0.7}{$\!\pm\!$0.57} & 52.88\scalebox{0.7}{$\!\pm\!$0.74} & 163.92\scalebox{0.7}{$\!\pm\!$14.67} \\
		
		& 1 & 0.99 & 94.73\scalebox{0.7}{$\!\pm\!$0.24} &
		94.69\scalebox{0.7}{$\!\pm\!$0.21} &
		100.00\scalebox{0.7}{$\!\pm\!$0.00} &
		0.98\scalebox{0.7}{$\!\pm\!$0.01}  & 0.13 & \textbf{52.09}\scalebox{0.7}{$\!\pm\!$0.73} & 53.55\scalebox{0.7}{$\!\pm\!$0.68} & 209.64\scalebox{0.7}{$\!\pm\!$9.61} \\ \cline{2-11}
		
		& 2 & 0.95 & 94.44\scalebox{0.7}{$\!\pm\!$0.28}&
		95.58\scalebox{0.7}{$\!\pm\!$0.21}&
		100.00\scalebox{0.7}{$\!\pm\!$0.00}&
		0.97\scalebox{0.7}{$\!\pm\!$0.01} &
		\textbf{0.12}&
		{52.42}\scalebox{0.7}{$\!\pm\!$0.16}&
		\textbf{52.15}\scalebox{0.7}{$\!\pm\!$1.30}&
		53.22\scalebox{0.7}{$\!\pm\!$3.21} \\
		
		& 2 & 0.99 & 94.59\scalebox{0.7}{$\!\pm\!$0.24} &
		95.15\scalebox{0.7}{$\!\pm\!$0.23} &
		100.00\scalebox{0.7}{$\!\pm\!$0.00} &
		0.98\scalebox{0.7}{$\!\pm\!$0.01}  & 0.12 & 52.23\scalebox{0.7}{$\!\pm\!$0.49} & 52.56\scalebox{0.7}{$\!\pm\!$0.53} & 101.37\scalebox{0.7}{$\!\pm\!$10.29} \\ \cline{2-11}
		
		& 3 & 0.95 & 94.57\scalebox{0.7}{$\!\pm\!$0.26} &
		96.33\scalebox{0.7}{$\!\pm\!$0.36} &
		100.00\scalebox{0.7}{$\!\pm\!$0.00} &
		0.95\scalebox{0.7}{$\!\pm\!$0.02}  & 0.13 & 53.52\scalebox{0.7}{$\!\pm\!$0.35} & 51.29\scalebox{0.7}{$\!\pm\!$0.72} & \textbf{47.67}\scalebox{0.7}{$\!\pm\!$0.52} \\
		
		& 3 & 0.99 & 94.54\scalebox{0.7}{$\!\pm\!$0.27} &
		95.82\scalebox{0.7}{$\!\pm\!$0.41} &
		100.00\scalebox{0.7}{$\!\pm\!$0.00} &
		0.96\scalebox{0.7}{$\!\pm\!$0.02}  & 0.13 & 53.44\scalebox{0.7}{$\!\pm\!$0.65} & 51.36\scalebox{0.7}{$\!\pm\!$0.43} & 55.83\scalebox{0.7}{$\!\pm\!$7.60} \\
		\midrule
		
		\multirow{6}{*}{\textbf{SALA+SMSE}}
		& 1 & 0.95 & 94.55\scalebox{0.7}{$\!\pm\!$0.24} &
		96.06\scalebox{0.7}{$\!\pm\!$0.21} &
		100.00\scalebox{0.7}{$\!\pm\!$0.00} &
		0.95\scalebox{0.7}{$\!\pm\!$0.03}  & 0.24 & 53.34\scalebox{0.7}{$\!\pm\!$0.46} & 56.48\scalebox{0.7}{$\!\pm\!$0.36} & 204.28\scalebox{0.7}{$\!\pm\!$10.30} \\
		
		& 1 & 0.99 & 94.56\scalebox{0.7}{$\!\pm\!$0.27} &
		96.06\scalebox{0.7}{$\!\pm\!$0.20} &
		100.00\scalebox{0.7}{$\!\pm\!$0.00} &
		0.95\scalebox{0.7}{$\!\pm\!$0.03}  & 0.24 & 53.32\scalebox{0.7}{$\!\pm\!$0.25} & 56.50\scalebox{0.7}{$\!\pm\!$0.75} & 212.64\scalebox{0.7}{$\!\pm\!$2.57} \\ \cline{2-11}
		
		& 2 & 0.95 & 94.25\scalebox{0.7}{$\!\pm\!$0.27}&
		95.13\scalebox{0.7}{$\!\pm\!$0.17}&
		99.97\scalebox{0.7}{$\!\pm\!$0.01}&
		0.98\scalebox{0.7}{$\!\pm\!$0.01} &
		0.21&
		54.28\scalebox{0.7}{$\!\pm\!$0.36}&
		55.97\scalebox{0.7}{$\!\pm\!$0.53}&
		56.47\scalebox{0.7}{$\!\pm\!$5.35}  \\
		
		& 2 & 0.99 & 94.35\scalebox{0.7}{$\!\pm\!$0.27} &
		95.63\scalebox{0.7}{$\!\pm\!$0.20} &
		99.99\scalebox{0.7}{$\!\pm\!$0.00} &
		0.97\scalebox{0.7}{$\!\pm\!$0.01}& 0.22 & 53.50\scalebox{0.7}{$\!\pm\!$0.26} & 55.83\scalebox{0.7}{$\!\pm\!$0.63} & 135.14\scalebox{0.7}{$\!\pm\!$9.56} \\ \cline{2-11}
		
		& 3 & 0.95 & 94.37\scalebox{0.7}{$\!\pm\!$0.28} &
		95.86\scalebox{0.7}{$\!\pm\!$0.27} &
		99.99\scalebox{0.7}{$\!\pm\!$0.00} &
		0.97\scalebox{0.7}{$\!\pm\!$0.01} & 0.28 & 54.15\scalebox{0.7}{$\!\pm\!$0.46} & 53.52\scalebox{0.7}{$\!\pm\!$0.72} & 48.52\scalebox{0.7}{$\!\pm\!$1.15} \\
		
		& 3 & 0.99 & 94.35\scalebox{0.7}{$\!\pm\!$0.26} &
		95.52\scalebox{0.7}{$\!\pm\!$0.21} &
		99.99\scalebox{0.7}{$\!\pm\!$0.01} &
		0.97\scalebox{0.7}{$\!\pm\!$0.01} & 0.25 & 53.96\scalebox{0.7}{$\!\pm\!$0.64} & 54.88\scalebox{0.7}{$\!\pm\!$0.83} & 74.81\scalebox{0.7}{$\!\pm\!$5.42} \\
		\midrule
		
		\multirow{6}{*}{\textbf{SALA w/ $D_\mathrm{f}$ only}}
		& 1 & 0.95 & 93.96\scalebox{0.7}{$\!\pm\!$0.31} &
		95.70\scalebox{0.7}{$\!\pm\!$0.24} &
		99.75\scalebox{0.7}{$\!\pm\!$0.11} &
		0.97\scalebox{0.7}{$\!\pm\!$0.01} & 0.27 & 53.80\scalebox{0.7}{$\!\pm\!$0.36} & 56.69\scalebox{0.7}{$\!\pm\!$1.13} & 139.11\scalebox{0.7}{$\!\pm\!$26.55} \\
		
		& 1 & 0.99 & 93.96\scalebox{0.7}{$\!\pm\!$0.28} &
		95.69\scalebox{0.7}{$\!\pm\!$0.27} &
		99.77\scalebox{0.7}{$\!\pm\!$0.10} &
		\textbf{0.99}\scalebox{0.7}{$\!\pm\!$0.01} & 0.26 & 53.78\scalebox{0.7}{$\!\pm\!$0.44} & 56.71\scalebox{0.7}{$\!\pm\!$0.72} & 135.68\scalebox{0.7}{$\!\pm\!$24.88} \\ \cline{2-11}
		
		& 2 & 0.95 & 93.75\scalebox{0.7}{$\!\pm\!$0.29}&
		94.49\scalebox{0.7}{$\!\pm\!$0.30}&
		99.75\scalebox{0.7}{$\!\pm\!$0.08}&
		\textbf{0.99}\scalebox{0.7}{$\!\pm\!$0.01} &
		0.23&
		53.37\scalebox{0.7}{$\!\pm\!$0.34}&
		56.75\scalebox{0.7}{$\!\pm\!$2.49}&
		59.35\scalebox{0.7}{$\!\pm\!$6.26} \\
		
		& 2 & 0.99 & 93.90\scalebox{0.7}{$\!\pm\!$0.26} &
		94.85\scalebox{0.7}{$\!\pm\!$0.63} &
		99.77\scalebox{0.7}{$\!\pm\!$0.11} &
		0.98\scalebox{0.7}{$\!\pm\!$0.01} & 0.24 & 53.79\scalebox{0.7}{$\!\pm\!$0.65} & 56.07\scalebox{0.7}{$\!\pm\!$0.43} & 129.29\scalebox{0.7}{$\!\pm\!$29.09} \\ \cline{2-11}
		
		& 3 & 0.95 & 93.92\scalebox{0.7}{$\!\pm\!$0.28} &
		95.37\scalebox{0.7}{$\!\pm\!$0.50} &
		99.85\scalebox{0.7}{$\!\pm\!$0.06} &
		0.98\scalebox{0.7}{$\!\pm\!$0.01} & 0.26 & 53.77 \scalebox{0.7}{$\!\pm\!$0.37}& 54.29\scalebox{0.7}{$\!\pm\!$0.82} & 74.74\scalebox{0.7}{$\!\pm\!$18.05} \\
		
		& 3 & 0.99 & 93.91\scalebox{0.7}{$\!\pm\!$0.31} &
		95.22\scalebox{0.7}{$\!\pm\!$0.47} &
		99.80\scalebox{0.7}{$\!\pm\!$0.09} &
		0.98\scalebox{0.7}{$\!\pm\!$0.01} & 0.24 & 53.76\scalebox{0.7}{$\!\pm\!$0.42} & 54.98\scalebox{0.7}{$\!\pm\!$0.75} & 118.27\scalebox{0.7}{$\!\pm\!$25.01} \\
		\bottomrule
	\end{tabular}
	\label{tab:param_sweep_z_threshold}
\end{table}

\Needspace{5\baselineskip}
The same trend holds for the relaxed SALA variants. SALA-SMSE remains stable across the sweep, with the default setting giving the lowest DNO among the SMSE configurations. Similarly, the $D_\mathrm{f}$-only variant releases the retain-side constraint and therefore produces moderately higher DNO, but it does not cause excessive DNO inflation across the tested settings.

\Needspace{4\baselineskip}
\subsection{Computational and storage overhead of SMSE}
\label{app:rebuttal-cost}
We quantify the computational and storage overhead of estimating the per-sample target distributions, comparing SMSE with the shadow-model estimator evaluated in Section~\ref{sec:estimation_accuracy}.

\paragraph{Storage overhead.}
Both estimators store four FP32 statistics per sample, $(\mu_{\mathrm{in}},\sigma_{\mathrm{in}},\mu_{\mathrm{out}},\sigma_{\mathrm{out}})$, adding $16N$ bytes for $N$ samples. Table~\ref{tab:rebuttal-storage} reports this persistent storage alongside the size of one trained model. During estimation, models or checkpoints can be processed sequentially.

\begin{table}[!htb]
\centering
\footnotesize
\setlength{\tabcolsep}{12pt}
\renewcommand{\arraystretch}{1.05}
\caption{Storage of the model and estimated distribution parameters (MiB).}
\label{tab:rebuttal-storage}
\begin{tabular}{@{}lcc@{}}
\toprule
\textbf{Dataset} & \textbf{Model} & \textbf{Distribution parameters} \\
\midrule
CIFAR-10 & 42.70 & 0.763 \\
ImageNet-1000 & 97.78 & 19.55 \\
WikiText-103 & 201.94 & 17.78 \\
\bottomrule
\end{tabular}
\end{table}

\Needspace{5\baselineskip}
\paragraph{Computational overhead.}
We measure computational overhead in two stages: training the models, and collecting their output scores to estimate the target-distribution parameters.
SMSE obtains the trained model and checkpoint ensemble within the $1\times$ training budget in Table~\ref{tab:estimation_accuracy}. Table~\ref{tab:rebuttal-training-cost} compares nominal training work and measured time relative to regular training on the same dataset and hardware. SMSE's training time remains close to regular training, while shadow-model estimation requires training multiple models.

After training, evaluating scores from $K$ models or checkpoints over $N$ samples requires $O(K\lceil N/B\rceil C_{\mathrm{fwd}}(B))$ work, where $B$ is the inference batch size and $C_{\mathrm{fwd}}(B)$ is the forward-pass cost per batch. Computing the distribution statistics then requires $O(NK)$ work. Table~\ref{tab:rebuttal-estimation-cost} separates these two components: forward inference dominates, while the statistical reduction takes seconds. For SMSE, this additional estimation time is $3.59\%$--$13.48\%$ of a regular training run, on top of the training time in Table~\ref{tab:rebuttal-training-cost}.

\begin{table}[!htb]
\centering
\scriptsize
\setlength{\tabcolsep}{3pt}
\renewcommand{\arraystretch}{1.05}
\caption{Training cost of shadow-model estimation and SMSE in Section~\ref{sec:estimation_accuracy}. Ratios in parentheses are relative to regular training.}
\label{tab:rebuttal-training-cost}
\begin{tabular*}{\linewidth}{@{\extracolsep{\fill}}lcccccc@{}}
\toprule
\multirow{2}{*}{\textbf{Method}} & \multicolumn{2}{c}{\textbf{CIFAR-10}} & \multicolumn{2}{c}{\textbf{ImageNet-1000}} & \multicolumn{2}{c}{\textbf{WikiText-103}} \\
\cmidrule(lr){2-3}\cmidrule(lr){4-5}\cmidrule(lr){6-7}
 & \textbf{Cost} & \textbf{Training time} & \textbf{Cost} & \textbf{Training time} & \textbf{Cost} & \textbf{Training time} \\
\midrule
Regular training & $1\times$ & 23.12 min & $1\times$ & 42.50 h & $1\times$ & 29.09 h \\
Shadow models & $32\times$ & 13.21 h ($34.28\times$) & $8\times$ & 307.73 h ($7.24\times$) & $8\times$ & 195.82 h ($6.73\times$) \\
SMSE & $1\times$ & 20.72 min ($0.90\times$) & $1\times$ & 45.72 h ($1.08\times$) & $1\times$ & 30.58 h ($1.05\times$) \\
\bottomrule
\end{tabular*}
\end{table}

\begin{table}[!htb]
\centering
\scriptsize
\setlength{\tabcolsep}{2.5pt}
\renewcommand{\arraystretch}{1.05}
\caption{Time to estimate distribution parameters from the trained models. Ratios are relative to regular training in Table~\ref{tab:rebuttal-training-cost}.}
\label{tab:rebuttal-estimation-cost}
\begin{tabular*}{\linewidth}{@{\extracolsep{\fill}}llcccc@{}}
\toprule
\textbf{Dataset} & \textbf{Method} & \textbf{Forward inference} & \textbf{Statistics} & \textbf{Total} & \textbf{Time ratio} \\
\midrule
\multirow{2}{*}{CIFAR-10} & Shadow models & 51.90 s & 1.10 s & 53.00 s & $0.0382\times$ \\
 & SMSE & 48.70 s & 1.12 s & 49.82 s & $0.0359\times$ \\
\midrule
\multirow{2}{*}{ImageNet-1000} & Shadow models & 1.72 h & 7.12 s & 1.72 h & $0.0405\times$ \\
 & SMSE & 5.72 h & 26.54 s & 5.73 h & $0.1348\times$ \\
\midrule
\multirow{2}{*}{WikiText-103} & Shadow models & 39.45 min & 6.43 s & 39.56 min & $0.0227\times$ \\
 & SMSE & 2.19 h & 24.10 s & 2.20 h & $0.0756\times$ \\
\bottomrule
\end{tabular*}
\end{table}

\Needspace{8\baselineskip}
\subsection{Ablation Study of SMSE}
\label{app:smse_ablation}

\Needspace{4\baselineskip}
\subsubsection{$\hat{\sigma}$ stabilization of SMSE}
\label{app:hat_sigma}
SMSE estimates per-sample standard deviations $\hat\sigma_{\mathrm{in}}(x)$ and $\hat\sigma_{\mathrm{out}}(x)$ from a limited ensemble, which can produce overly small or overly dispersed values for a subset of samples. This matters because SALA uses $\hat\sigma(x)$ to standardize scores and to form the InDist constraint; extreme $\hat\sigma(x)$ values can inflate the effective mismatch term and lead to a systematic DNO offset (Section~\ref{subsec:est_error_bound}). To mitigate this, SMSE applies a sigma stabilization step that shrinks per-sample $\hat\sigma(x)$ toward a global reference scale and clips it into a bounded range, preventing collapse and heavy tails. The ablation compares \textbf{SALA-SMSE} with and without sigma stabilization, keeping all other components fixed. Results are reported under the same unlearning task suite as the main experiments.

\paragraph{Results.} 
Table~\ref{tab:sigma_ablation} shows that sigma stabilization consistently improves sample-level alignment: it reduces DNO by about $0.01$ and RULI accuracy by about 0.49$\%$ on average, with negligible change to ToW. This supports the interpretation in Section~\ref{subsec:est_error_bound}: bounding $\hat\sigma(x)$ directly reduces the estimation-induced term in the DNO decomposition and tightens the achievable DNO under SALA-SMSE.

\begin{table*}[!htb]
	\centering
	\scriptsize
	\setlength{\tabcolsep}{1.86pt}
	\caption{Ablation study of SMSE sigma stabilization mechanism}
	\label{tab:sigma_ablation}
	\begin{tabular}{lcccccccccc}
		\toprule
		\textbf{Method} & 
		\textbf{Sigma Stable} &
			\textbf{TA}(\%)$\uparrow$ &
		\textbf{FA}(\%)$\uparrow$ &
		\textbf{RA}(\%)$\uparrow$ &
		\textbf{ToW}$\uparrow$ &
		\textbf{DNO}$\downarrow$ &
		\textbf{MIA efficacy}(\%)$\downarrow$ &
		\textbf{RULI}(\%)$\downarrow$ &
		\textbf{Time}(s)$\downarrow$ \\
		\hline
		SALA-SMSE + Salun&\ding{51}&
		94.25\scalebox{0.7}{$\!\pm\!$0.27}&
		95.13\scalebox{0.7}{$\!\pm\!$0.17}&
		99.97\scalebox{0.7}{$\!\pm\!$0.01}&
		0.98\scalebox{0.7}{$\!\pm\!$0.01} &
		0.21&
		54.28\scalebox{0.7}{$\!\pm\!$0.36}&
		55.97\scalebox{0.7}{$\!\pm\!$0.53}&
		56.47\scalebox{0.7}{$\!\pm\!$5.35}   \\
		SALA-SMSE + SCRUB &\ding{51}&
		94.01\scalebox{0.7}{$\!\pm\!$0.23}&
		94.03\scalebox{0.7}{$\!\pm\!$0.16}&
		99.97\scalebox{0.7}{$\!\pm\!$0.02}&
		{0.98}\scalebox{0.7}{$\!\pm\!$0.01}&
		0.23&
		54.43\scalebox{0.7}{$\!\pm\!$0.16}&
		55.25\scalebox{0.7}{$\!\pm\!$0.27}&
		46.45\scalebox{0.7}{$\!\pm\!$2.79}  \\
		SALA-SMSE + $\ell_1$&\ding{51}&
		93.13\scalebox{0.7}{$\!\pm\!$0.31}&
		93.52\scalebox{0.7}{$\!\pm\!$0.42}&
		99.99\scalebox{0.7}{$\!\pm\!$0.00}&
		\textbf{0.99}\scalebox{0.7}{$\!\pm\!$0.01}&
		\textbf{0.20}&
		\textbf{53.25}\scalebox{0.7}{$\!\pm\!$0.27}&
		\textbf{52.93}\scalebox{0.7}{$\!\pm\!$0.31}&
		125.52\scalebox{0.7}{$\!\pm\!$9.14}
		  \\  \hline
		SALA-SMSE  + Salun&\ding{55}&
        93.80\scalebox{0.7}{$\!\pm\!$0.19}&
        94.74\scalebox{0.7}{$\!\pm\!$0.22}&
        100.00\scalebox{0.7}{$\!\pm\!$0.00}&
        {0.97}\scalebox{0.7}{$\!\pm\!$0.01}&
        0.23&
        54.74\scalebox{0.7}{$\!\pm\!$0.25}&
        56.00\scalebox{0.7}{$\!\pm\!$0.31}&
        76.73\scalebox{0.7}{$\!\pm\!$7.15} \\
		SALA-SMSE  + SCRUB & \ding{55}&
        93.76\scalebox{0.7}{$\!\pm\!$0.21}&
        94.43\scalebox{0.7}{$\!\pm\!$0.35}&
        99.97\scalebox{0.7}{$\!\pm\!$0.02}&
        {0.97}\scalebox{0.7}{$\!\pm\!$0.00}&
        0.23&
        54.76\scalebox{0.7}{$\!\pm\!$0.19}&
        56.47\scalebox{0.7}{$\!\pm\!$0.21}&
        \textbf{44.35}\scalebox{0.7}{$\!\pm\!$2.21} \\
		SALA-SMSE  + $\ell_1$ sparsity&
        \ding{55}&
        93.92\scalebox{0.7}{$\!\pm\!$0.27}&
        94.14\scalebox{0.7}{$\!\pm\!$0.31}&
        99.99\scalebox{0.7}{$\!\pm\!$0.01}&
        {0.97}\scalebox{0.7}{$\!\pm\!$0.01}&
        0.23&
        54.41\scalebox{0.7}{$\!\pm\!$0.08}&
        53.16\scalebox{0.7}{$\!\pm\!$0.13}&
        108.70\scalebox{0.7}{$\!\pm\!$9.15}  \\ 
		\bottomrule
	\end{tabular}
\end{table*}

\Needspace{4\baselineskip}
\subsubsection{Epoch allocation of SMSE}
\label{sec:ablation_smse_schedule}
We further study how the epoch allocation in SMSE affects per-sample distribution estimation. In the main experiments, the training budget is divided into $20\%$ warmup, $70\%$ half-data training, and $10\%$ fine-tuning after merging. To examine the sensitivity of this schedule, we vary the warmup and merge allocations. Estimation quality is evaluated using the same metrics as in Table~\ref{tab:estimation_accuracy}, including Pearson correlation, MSE for the four estimated moments and $\mathrm{DNO}_\mathrm{est}$. We also report the final test accuracy of each trained model.

\begin{table}[!htb]
	\centering
	\scriptsize
	\setlength{\tabcolsep}{4.65pt}
	\renewcommand{\arraystretch}{1.05}
	\caption{Ablation on warmup and merge scheduling in SMSE on CIFAR-10.}
	\label{tab:smse_warmup_merge_ablation}
	\begin{tabular}{ll cccc cccc cc}
		\toprule
		\multirow{2}{*}{\makecell[c]{\textbf{Warmup Epoch}\\\textbf{Ratio} (\%)}} &
		\multirow{2}{*}{\makecell[c]{\textbf{Merge Epoch}\\\textbf{Ratio} (\%)}} &
		\multicolumn{2}{c}{$\boldsymbol{\mu_{\mathrm{in}}}$} &
		\multicolumn{2}{c}{$\boldsymbol{\mu_{\mathrm{out}}}$} &
		\multicolumn{2}{c}{$\boldsymbol{\sigma_{\mathrm{in}}}$} &
		\multicolumn{2}{c}{$\boldsymbol{\sigma_{\mathrm{out}}}$} &
		\multirow{2}{*}{$\mathrm{DNO}_\mathrm{est}\downarrow$} &
		\multirow{2}{*}{\textbf{Final Acc.}$\uparrow$} \\
		\cmidrule(lr){3-4} \cmidrule(lr){5-6} \cmidrule(lr){7-8} \cmidrule(lr){9-10}
		 & & \textbf{r}$\uparrow$ & \textbf{MSE}$\downarrow$
		& \textbf{r}$\uparrow$ & \textbf{MSE}$\downarrow$
		& \textbf{r}$\uparrow$ & \textbf{MSE}$\downarrow$
		& \textbf{r}$\uparrow$ & \textbf{MSE}$\downarrow$ && \\
		\midrule
			~0 & 10
		& 0.75 & 1.57
		& 0.90 & 2.34
		& {0.59} & \textbf{0.42}
		& \textbf{0.70} & \textbf{0.40}
		& 0.2464
		& \textbf{94.87}\\
		
			10 & 10
		& \textbf{0.79} & {1.27}
		& 0.91 & 1.95
		& 0.56 & 0.70
		& 0.60 & 0.57
		& 0.1999
		& 94.49 \\
		
		20 & 10 (Main setup)
		& \textbf{0.79} & 1.29
		& \textbf{0.92} & \textbf{1.90}
		& 0.58 & 0.67
		& 0.63 & 0.52
		& \textbf{0.1801}
		& 94.79 \\
		
		20 & 20
		& \textbf{0.79} & \textbf{0.91}
		& \textbf{0.92} & 2.07
		& \textbf{0.60} & 1.27
		& 0.65 & 0.79
		& 0.1822
		& 94.78 \\
		
		20 & ~0
		& 0.74 & 0.97
		& \textbf{0.92} & \textbf{1.90}
		& 0.58& 0.67
		&  0.64 & 0.53
		& 0.1832
		& 83.59 \\
		\bottomrule
	\end{tabular}

\end{table}

Table~\ref{tab:smse_warmup_merge_ablation} reports the results on CIFAR-10. Warmup mainly improves the calibration of per-sample moment estimates. Without warmup, SMSE obtains the worst $\mathrm{DNO}_{\mathrm{est}}$ of 0.2464 and lower Pearson correlation for $\mu_{\mathrm{in}}$, although the final accuracy remains high. Introducing a moderate warmup substantially improves estimation quality: $10\%$ warmup reduces $\mathrm{DNO}_{\mathrm{est}}$ to 0.1999, while the main $20\%$ warmup setting further reduces it to 0.1801.

The merge stage has a different role. With the warmup ratio fixed at $20\%$, removing fine-tuning after merging sharply reduces final accuracy from 94.79\% to 83.59\%, while leaving the estimation metrics relatively competitive. Thus, merging is primarily needed to recover the predictive utility of the final model rather than to improve moment estimation alone. Overall, the main schedule with $20\%$ warmup and $10\%$ merge achieves the lowest $\mathrm{DNO}_{\mathrm{est}}$ while maintaining high final accuracy, and is therefore used as the default SMSE configuration.

\Needspace{4\baselineskip}
\subsubsection{Segment ensemble size $M$ of SMSE}
\label{sec:ablation_smse_M}

We further ablate the segment ensemble size $M$, defined as the number of checkpoints per half-data training trajectory used to estimate per-sample score moments. In the main experiments, we set $M=30$. To examine sensitivity to this choice, we vary $M\in\{10,20,30,40\}$ and evaluate estimation quality using the same metrics as in Table~\ref{tab:estimation_accuracy}.

Table~\ref{tab:smse_M_ablation} reports the results on CIFAR-10. Increasing $M$ from 10 generally improves estimation stability, particularly for the variance terms: the MSE of $\sigma_{\mathrm{out}}$ decreases from 2.69 at $M=10$ to 0.52 at $M=30$. However, using a larger ensemble does not uniformly improve all metrics. Although $M=40$ achieves the best correlation for $\sigma_{\mathrm{in}}$, it worsens the MSE of all four moments compared with $M=30$. The main setting $M=30$ therefore provides the best overall balance, achieving the strongest mean estimation for $\mu_{\mathrm{out}}$ and the lowest MSE for both $\mu_{\mathrm{out}}$ and $\sigma_{\mathrm{out}}$.

\begin{table}[!htb]
	\centering
	\scriptsize
	\setlength{\tabcolsep}{7.0pt}
	\renewcommand{\arraystretch}{1.05}
	\caption{Ablation on the segment ensemble size $M$ in SMSE on CIFAR-10.}
	\label{tab:smse_M_ablation}
	\begin{tabular}{l cccc cccc}
		\toprule
		\multirow{2}{*}{$\boldsymbol{M}$} &
		\multicolumn{2}{c}{$\boldsymbol{\mu_{\mathrm{in}}}$} &
		\multicolumn{2}{c}{$\boldsymbol{\mu_{\mathrm{out}}}$} &
		\multicolumn{2}{c}{$\boldsymbol{\sigma_{\mathrm{in}}}$} &
		\multicolumn{2}{c}{$\boldsymbol{\sigma_{\mathrm{out}}}$} \\
		\cmidrule(lr){2-3} \cmidrule(lr){4-5} \cmidrule(lr){6-7} \cmidrule(lr){8-9}
		& \textbf{r}$\uparrow$ & \textbf{MSE}$\downarrow$
		& \textbf{r}$\uparrow$ & \textbf{MSE}$\downarrow$
		& \textbf{r}$\uparrow$ & \textbf{MSE}$\downarrow$
		& \textbf{r}$\uparrow$ & \textbf{MSE}$\downarrow$ \\
		\midrule
		10 & 0.77 & 1.30& 0.89 & 2.47 & 0.53 & 1.06 & \textbf{0.73} & 2.69 \\
		20 & 0.78 & \textbf{1.19} & 0.90 & 2.24 & 0.57 & \textbf{0.42} & 0.70 & 1.38 \\
		{30} (Main setup) & \textbf{0.79} & 1.29 & \textbf{0.92} & \textbf{1.90} & 0.58 & 0.67 & 0.63 & \textbf{0.52} \\
		40 & 0.77 & 1.70 & \textbf{0.92} & 2.08 & \textbf{0.60} & 0.85 & 0.65 & 0.58 \\
		\bottomrule
	\end{tabular}

\end{table}

\Needspace{4\baselineskip}
\subsubsection{Estimation methods of SMSE}
\label{sec:ablation_smse_estimation_method}
We examine alternative ways of constructing the stochastic ensemble used by SMSE for moment estimation. The default design uses a \textbf{checkpoint ensemble}, where checkpoints collected along the two half-data training trajectories are aggregated to estimate per-sample score moments. We compare this strategy with two lightweight alternatives that induce stochasticity from a single trained model, namely \textbf{Monte Carlo dropout} and repeated \textbf{stochastic data augmentation}.

\begin{table}[H]
	\centering
	\scriptsize
	\setlength{\tabcolsep}{3.75pt}
	\renewcommand{\arraystretch}{1.05}
	\caption{Ablation on different stochastic estimation methods in SMSE on CIFAR-10.}
	\label{tab:smse_estimation_method_ablation}
	\begin{tabular}{l cccc cccc}
		\toprule
		\multirow{2}{*}{\textbf{Method}} &
		\multicolumn{2}{c}{$\boldsymbol{\mu_{\mathrm{in}}}$} &
		\multicolumn{2}{c}{$\boldsymbol{\mu_{\mathrm{out}}}$} &
		\multicolumn{2}{c}{$\boldsymbol{\sigma_{\mathrm{in}}}$} &
		\multicolumn{2}{c}{$\boldsymbol{\sigma_{\mathrm{out}}}$} \\
		\cmidrule(lr){2-3} \cmidrule(lr){4-5} \cmidrule(lr){6-7} \cmidrule(lr){8-9}
		& \textbf{r}$\uparrow$ & \textbf{MSE}$\downarrow$
		& \textbf{r}$\uparrow$ & \textbf{MSE}$\downarrow$
		& \textbf{r}$\uparrow$ & \textbf{MSE}$\downarrow$
		& \textbf{r}$\uparrow$ & \textbf{MSE}$\downarrow$ \\
		\midrule
		{Checkpoint ensemble} (Main setup)
		& \textbf{0.79} & 1.29
		& \textbf{0.92} & \textbf{1.90}
		& 0.58 & 0.67
		& 0.64 & \textbf{0.52} \\
		Dropout
		& 0.72 & \textbf{1.18}
		& 0.89 & 2.55
		& -0.15 & \textbf{0.35}
		& -0.42 & 1.84 \\
		Data augmentation
		& 0.71 & 1.55
		& 0.91 & 2.48
		& \textbf{0.61} & 0.41
		& \textbf{0.76} & 1.00\\
		\bottomrule
	\end{tabular}

\end{table}

Table~\ref{tab:smse_estimation_method_ablation} reports the results on CIFAR-10. The checkpoint ensemble provides the most balanced estimation quality across the four target moments. It achieves the strongest correlations for both mean statistics, with $r=0.79$ for $\mu_{\mathrm{in}}$ and $r=0.92$ for $\mu_{\mathrm{out}}$, and also attains the lowest MSE for $\mu_{\mathrm{out}}$ and $\sigma_{\mathrm{out}}$. Data augmentation produces competitive variance correlations, but its mean estimation is weaker and its $\sigma_{\mathrm{out}}$ MSE remains higher than that of the checkpoint ensemble. Monte Carlo dropout achieves low MSE on some individual terms, but yields negative correlations for both $\sigma_{\mathrm{in}}$ and $\sigma_{\mathrm{out}}$, indicating that its induced randomness does not preserve the variance structure required by SMSE.

\subsection{Utility of Models Produced by SMSE}
\label{app:smse_utility}
\begin{table}[!htb]
	\centering
	\footnotesize
	\setlength{\tabcolsep}{6.8pt}
	\caption{Utility of models produced by SMSE.}
	\label{tab:smse_utility}
	\begin{tabular}{lcc}
		\toprule
		\textbf{Dataset} & \textbf{Standard Training} & \textbf{SMSE} \\
		\midrule
		CIFAR-10 (ACC$\uparrow$)      &
		94.68\scalebox{0.7}{$\pm$0.35} &
		 94.79\scalebox{0.7}{$\pm$0.13}\\
		ImageNet-1000 (ACC$\uparrow$) & 
		73.55\scalebox{0.7}{$\pm$0.13}&
		 73.37\scalebox{0.7}{$\pm$0.23} \\
		WikiText-103 (PPL$\downarrow$)& 
		53.36\scalebox{0.7}{$\pm$0.61}  & 
		54.35\scalebox{0.7}{$\pm$1.05} \\
		\bottomrule
	\end{tabular}

\end{table}

\Needspace{4\baselineskip}
\subsection{Experiments of SALA combined with various approximate unlearning algorithms}
\label{app:other_backbones}
\paragraph{Results.} 
Table~\ref{tab:sala} evaluates whether SALA can act as a stable plug-in mechanism for different approximate unlearning algorithms. Across SalUn, SCRUB, and $\ell_1$ sparsity unlearning, SALA consistently improves the corresponding base unlearning methods: DNO drops by 0.78, 0.23, and 0.21, respectively, while RULI is reduced by 40.23\%, 16.55\%, and 17.98\%. SALA also maintains nearly perfect ToW across all three base methods and keeps runtime far below full retraining.

When using only ${D}_\mathrm{f}$, DNO remains within 0.22 to 0.23, which is still lower than the corresponding base unlearning methods. Similarly, although SALA-SMSE uses an imperfect reference distribution with additional estimation error, its DNO remains within 0.20 to 0.23 and consistently improves over the baselines.

\begin{table*}[!htb]
	\centering
	\scriptsize
	\setlength{\tabcolsep}{3.80pt}
	\renewcommand{\arraystretch}{1.15}
	\caption{Comprehensive evaluation of SALA-assisted approximate unlearning methods.}
	\label{tab:sala}
	\begin{tabular}{lcccccccc}
		\toprule
		\textbf{Method} &
				\textbf{TA}(\%)$\uparrow$ &
			\textbf{FA}(\%)$\uparrow$ &
			\textbf{RA}(\%)$\uparrow$ &
			\textbf{ToW}$\uparrow$ &
			\textbf{DNO}$\downarrow$ &
			\textbf{MIA efficacy}(\%)$\downarrow$ &
			\textbf{RULI}(\%)$\downarrow$ &
			\textbf{Time}(s)$\downarrow$ \\ \hline
					Retrain& 
			93.84\scalebox{0.7}{$\!\pm\!$0.21}& 
			93.68\scalebox{0.7}{$\!\pm\!$0.28} &
			100.00\scalebox{0.7}{$\!\pm\!$0.01} &
			1.00\scalebox{0.7}{$\!\pm\!$0.00}& 
			0.07 &
			50.57\scalebox{0.7}{$\!\pm\!$0.25}& 
			50.00\scalebox{0.7}{$\!\pm\!$0.00}&
			1625.53\scalebox{0.7}{$\!\pm\!$5.33}\\
			\hline
		SalUn&92.36\scalebox{0.7}{$\!\pm\!$0.45} &
		95.29\scalebox{0.7}{$\!\pm\!$0.57} &
		98.61\scalebox{0.7}{$\!\pm\!$0.36} &
		0.95\scalebox{0.7}{$\!\pm\!$0.01}& 
		0.90 &
		51.87\scalebox{0.7}{$\!\pm\!$0.21}  &
		92.38\scalebox{0.7}{$\!\pm\!$0.28}  &
		221.92\scalebox{0.7}{$\!\pm\!$96.37} \\
		SCRUB&
		90.29\scalebox{0.7}{$\!\pm\!$2.31} &
		94.06\scalebox{0.7}{$\!\pm\!$3.15} &
		95.74\scalebox{0.7}{$\!\pm\!$3.00} &
		0.92\scalebox{0.7}{$\!\pm\!$0.03}  & 
		0.34 &
		52.23\scalebox{0.7}{$\!\pm\!$0.57}  & 
		68.67\scalebox{0.7}{$\!\pm\!$3.62}  & 
		101.93\scalebox{0.7}{$\!\pm\!$11.98} \\
			$\ell_1$ sparsity&
		88.18\scalebox{0.7}{$\!\pm\!$3.47} & 
		91.94\scalebox{0.7}{$\!\pm\!$4.48} & 
		95.80\scalebox{0.7}{$\!\pm\!$2.98} &
		0.88\scalebox{0.7}{$\!\pm\!$0.02}  & 
		0.32 &
		53.09\scalebox{0.7}{$\!\pm\!$2.72}  & 
		70.03\scalebox{0.7}{$\!\pm\!$5.24}  & 
		150.28\scalebox{0.7}{$\!\pm\!$28.45} \\
		\hline
		SALA + Salun&
		94.44\scalebox{0.7}{$\!\pm\!$0.28}&
		95.58\scalebox{0.7}{$\!\pm\!$0.21}&
		100.00\scalebox{0.7}{$\!\pm\!$0.00}&
		0.97\scalebox{0.7}{$\!\pm\!$0.02} &
		{0.12}&
		{52.42}\scalebox{0.7}{$\!\pm\!$0.16}&
		{52.15}\scalebox{0.7}{$\!\pm\!$1.30}&
		\textbf{53.22}\scalebox{0.7}{$\!\pm\!$3.21}\\
		SALA + SCRUB&
				94.35\scalebox{0.7}{$\!\pm\!$0.15}&
                94.03\scalebox{0.7}{$\!\pm\!$0.14}&
                99.99\scalebox{0.7}{$\!\pm\!$0.00}&
                \textbf{0.99}\scalebox{0.7}{$\!\pm\!$0.01}&
                \textbf{0.11}&
                {52.46}\scalebox{0.7}{$\!\pm\!$0.11}&
                {52.12}\scalebox{0.7}{$\!\pm\!$0.07}&
                50.31\scalebox{0.7}{$\!\pm\!$6.62}\\
		SALA + $\ell_1$&
				94.31\scalebox{0.7}{$\!\pm\!$0.19}&
                94.75\scalebox{0.7}{$\!\pm\!$0.27}&
                99.99\scalebox{0.7}{$\!\pm\!$0.01}&
                {0.98}\scalebox{0.7}{$\!\pm\!$0.01}&
                \textbf{0.11}&
                \textbf{51.23}\scalebox{0.7}{$\!\pm\!$0.02}&
                \textbf{52.05}\scalebox{0.7}{$\!\pm\!$0.14}&
                97.36\scalebox{0.7}{$\!\pm\!$4.14} \\  \hline
		SALA-SMSE + Salun&
        94.25\scalebox{0.7}{$\!\pm\!$0.27}&
        95.13\scalebox{0.7}{$\!\pm\!$0.17}&
        99.97\scalebox{0.7}{$\!\pm\!$0.01}&
        0.98\scalebox{0.7}{$\!\pm\!$0.01} &
        0.21&
        54.28\scalebox{0.7}{$\!\pm\!$0.36}&
        55.97\scalebox{0.7}{$\!\pm\!$0.53}&
        56.47\scalebox{0.7}{$\!\pm\!$5.35}   \\
		SALA-SMSE + SCRUB &
        95.01\scalebox{0.7}{$\!\pm\!$0.23}&
        94.03\scalebox{0.7}{$\!\pm\!$0.16}&
        99.97\scalebox{0.7}{$\!\pm\!$0.02}&
        {0.98}\scalebox{0.7}{$\!\pm\!$0.01}&
        0.23&
        54.43\scalebox{0.7}{$\!\pm\!$0.16}&
        55.25\scalebox{0.7}{$\!\pm\!$0.27}&
        46.45\scalebox{0.7}{$\!\pm\!$2.79}  \\
		SALA-SMSE + $\ell_1$&
        93.13\scalebox{0.7}{$\!\pm\!$0.31}&
        93.52\scalebox{0.7}{$\!\pm\!$0.42}&
        99.99\scalebox{0.7}{$\!\pm\!$0.00}&
        \textbf{0.99}\scalebox{0.7}{$\!\pm\!$0.01}&
        0.20&
        53.25\scalebox{0.7}{$\!\pm\!$0.27}&
        52.93\scalebox{0.7}{$\!\pm\!$0.31}&
        125.52\scalebox{0.7}{$\!\pm\!$9.14}
		 \\  \hline
		 SALA w/ ${D}_\mathrm{f}$ only + Salun&
		 	93.75\scalebox{0.7}{$\!\pm\!$0.29}&
		 	94.49\scalebox{0.7}{$\!\pm\!$0.30}&
		 	99.75\scalebox{0.7}{$\!\pm\!$0.08}&
		 	\textbf{0.99}\scalebox{0.7}{$\!\pm\!$0.01} &
		 	0.23&
		 	53.37\scalebox{0.7}{$\!\pm\!$0.34}&
		 	56.75\scalebox{0.7}{$\!\pm\!$2.49}&
		 	59.35\scalebox{0.7}{$\!\pm\!$6.26}  \\
		 	SALA w/ ${D}_\mathrm{f}$ only  + SCRUB &
		 93.41\scalebox{0.7}{$\!\pm\!$0.16}&
		 93.89\scalebox{0.7}{$\!\pm\!$0.34}&
		 99.97\scalebox{0.7}{$\!\pm\!$0.02}&
		 \textbf{0.99}\scalebox{0.7}{$\!\pm\!$0.01}&
		 0.23&
		 53.43\scalebox{0.7}{$\!\pm\!$0.53}&
		 55.25\scalebox{0.7}{$\!\pm\!$0.64}&
		 51.45\scalebox{0.7}{$\!\pm\!$5.24}  \\
		 	SALA w/ ${D}_\mathrm{f}$ only  + $\ell_1$& 
		 93.13\scalebox{0.7}{$\!\pm\!$0.53}&
		 93.52\scalebox{0.7}{$\!\pm\!$0.49}&
		 99.99\scalebox{0.7}{$\!\pm\!$0.00}&
		 \textbf{0.99}\scalebox{0.7}{$\!\pm\!$0.01}&
		 0.22&
		 51.25\scalebox{0.7}{$\!\pm\!$0.15}&
		 53.03\scalebox{0.7}{$\!\pm\!$0.28}&
		 115.52\scalebox{0.7}{$\!\pm\!$6.35}
		 \\ 
		 \bottomrule
	\end{tabular}
\end{table*}

\Needspace{4\baselineskip}
\subsection{Sensitivity to the forget-set size}
\label{app:rebuttal-forget-ratio}
\label{app:rebuttal-ratios}
The main sample-wise and sequence unlearning experiments use forget ratios of $20\%$ on CIFAR-10 and $33\%$ on ImageNet-1000 and WikiText-103, providing observations in both membership roles over repeated tasks. We additionally evaluate $5\%$, $10\%$, and $50\%$ on CIFAR-10, keeping the other settings fixed and using eight independent unlearning runs for each ratio. Table~\ref{tab:rebuttal-forget-utility} reports the same utility, efficacy, privacy, and runtime metrics as Table~\ref{tab:baseline_comparison}. Across the three ratios, SALA and its variants obtain lower DNO and RULI than SalUn, with comparable or higher ToW.

\begin{table}[!htb]
\centering
\scriptsize
\setlength{\tabcolsep}{2.5pt}
\renewcommand{\arraystretch}{1.05}
\caption{CIFAR-10 results under different forget ratios.}
\label{tab:rebuttal-forget-utility}
\label{tab:rebuttal-forget-efficacy}
\begin{tabular*}{\linewidth}{@{\extracolsep{\fill}}lcccccccc@{}}
\toprule
\textbf{Method} & \textbf{TA(\%)} & \textbf{FA(\%)} & \textbf{RA(\%)} & \textbf{ToW} & \textbf{DNO} & \textbf{MIA(\%)} & \textbf{RULI(\%)} & \textbf{Time(s)} \\
\midrule
\multicolumn{9}{c}{\textbf{Forget ratio: 5\%}} \\
\midrule
Retrain & 94.65\scalebox{0.7}{$\!\pm\!$0.37} & 94.33\scalebox{0.7}{$\!\pm\!$0.21} & 100.00\scalebox{0.7}{$\!\pm\!$0.00} & 1.00\scalebox{0.7}{$\!\pm\!$0.00} & 0.06 & 51.15\scalebox{0.7}{$\!\pm\!$0.08} & 50.23\scalebox{0.7}{$\!\pm\!$0.07} & 1457.39\scalebox{0.7}{$\!\pm\!$10.74} \\
SalUn & 94.45\scalebox{0.7}{$\!\pm\!$0.08} & 95.72\scalebox{0.7}{$\!\pm\!$0.14} & 100.00\scalebox{0.7}{$\!\pm\!$0.00} & 0.96\scalebox{0.7}{$\!\pm\!$0.01} & 0.56 & 53.15\scalebox{0.7}{$\!\pm\!$0.14} & 74.42\scalebox{0.7}{$\!\pm\!$0.57} & 94.45\scalebox{0.7}{$\!\pm\!$5.47} \\
SALA & 94.10\scalebox{0.7}{$\!\pm\!$0.24} & 93.90\scalebox{0.7}{$\!\pm\!$0.13} & 100.00\scalebox{0.7}{$\!\pm\!$0.00} & 0.99\scalebox{0.7}{$\!\pm\!$0.00} & 0.09 & 51.03\scalebox{0.7}{$\!\pm\!$0.11} & 52.14\scalebox{0.7}{$\!\pm\!$0.29} & 38.49\scalebox{0.7}{$\!\pm\!$1.52} \\
SALA-SMSE & 94.05\scalebox{0.7}{$\!\pm\!$0.13} & 94.21\scalebox{0.7}{$\!\pm\!$0.18} & 100.00\scalebox{0.7}{$\!\pm\!$0.00} & 0.97\scalebox{0.7}{$\!\pm\!$0.01} & 0.24 & 52.14\scalebox{0.7}{$\!\pm\!$0.12} & 59.15\scalebox{0.7}{$\!\pm\!$0.75} & 34.08\scalebox{0.7}{$\!\pm\!$2.47} \\
SALA w/ $D_\mathrm{f}$ only & 93.91\scalebox{0.7}{$\!\pm\!$0.15} & 94.33\scalebox{0.7}{$\!\pm\!$0.16} & 100.00\scalebox{0.7}{$\!\pm\!$0.00} & 0.97\scalebox{0.7}{$\!\pm\!$0.01} & 0.25 & 52.28\scalebox{0.7}{$\!\pm\!$0.14} & 59.87\scalebox{0.7}{$\!\pm\!$0.68} & 35.36\scalebox{0.7}{$\!\pm\!$2.21} \\
\midrule
\multicolumn{9}{c}{\textbf{Forget ratio: 10\%}} \\
\midrule
Retrain & 94.12\scalebox{0.7}{$\!\pm\!$0.12} & 94.02\scalebox{0.7}{$\!\pm\!$0.25} & 100.00\scalebox{0.7}{$\!\pm\!$0.00} & 1.00\scalebox{0.7}{$\!\pm\!$0.00} & 0.04 & 51.03\scalebox{0.7}{$\!\pm\!$0.09} & 50.10\scalebox{0.7}{$\!\pm\!$0.35} & 1572.27\scalebox{0.7}{$\!\pm\!$6.35} \\
SalUn & 94.41\scalebox{0.7}{$\!\pm\!$0.25} & 94.43\scalebox{0.7}{$\!\pm\!$0.11} & 100.00\scalebox{0.7}{$\!\pm\!$0.00} & 0.96\scalebox{0.7}{$\!\pm\!$0.01} & 0.76 & 54.35\scalebox{0.7}{$\!\pm\!$0.05} & 82.79\scalebox{0.7}{$\!\pm\!$0.28} & 115.72\scalebox{0.7}{$\!\pm\!$19.15} \\
SALA & 94.20\scalebox{0.7}{$\!\pm\!$0.15} & 93.78\scalebox{0.7}{$\!\pm\!$0.47} & 100.00\scalebox{0.7}{$\!\pm\!$0.00} & 0.98\scalebox{0.7}{$\!\pm\!$0.01} & 0.10 & 50.79\scalebox{0.7}{$\!\pm\!$0.35} & 51.95\scalebox{0.7}{$\!\pm\!$0.51} & 36.20\scalebox{0.7}{$\!\pm\!$1.13} \\
SALA-SMSE & 93.95\scalebox{0.7}{$\!\pm\!$0.12} & 93.92\scalebox{0.7}{$\!\pm\!$0.47} & 100.00\scalebox{0.7}{$\!\pm\!$0.00} & 0.98\scalebox{0.7}{$\!\pm\!$0.00} & 0.25 & 52.75\scalebox{0.7}{$\!\pm\!$0.41} & 58.37\scalebox{0.7}{$\!\pm\!$1.54} & 33.79\scalebox{0.7}{$\!\pm\!$1.63} \\
SALA w/ $D_\mathrm{f}$ only & 94.08\scalebox{0.7}{$\!\pm\!$0.14} & 94.26\scalebox{0.7}{$\!\pm\!$0.39} & 100.00\scalebox{0.7}{$\!\pm\!$0.00} & 0.98\scalebox{0.7}{$\!\pm\!$0.01} & 0.26 & 52.52\scalebox{0.7}{$\!\pm\!$0.36} & 57.91\scalebox{0.7}{$\!\pm\!$1.42} & 34.27\scalebox{0.7}{$\!\pm\!$1.81} \\
\midrule
\multicolumn{9}{c}{\textbf{Forget ratio: 50\%}} \\
\midrule
Retrain & 92.68\scalebox{0.7}{$\!\pm\!$0.25} & 92.53\scalebox{0.7}{$\!\pm\!$0.18} & 100.00\scalebox{0.7}{$\!\pm\!$0.00} & 1.00\scalebox{0.7}{$\!\pm\!$0.00} & 0.07 & 51.10\scalebox{0.7}{$\!\pm\!$0.05} & 50.25\scalebox{0.7}{$\!\pm\!$0.11} & 1625.53\scalebox{0.7}{$\!\pm\!$5.33} \\
SalUn & 89.88\scalebox{0.7}{$\!\pm\!$0.48} & 92.40\scalebox{0.7}{$\!\pm\!$0.54} & 98.05\scalebox{0.7}{$\!\pm\!$0.38} & 0.95\scalebox{0.7}{$\!\pm\!$0.00} & 0.91 & 52.35\scalebox{0.7}{$\!\pm\!$0.10} & 94.30\scalebox{0.7}{$\!\pm\!$2.22} & 221.92\scalebox{0.7}{$\!\pm\!$96.37} \\
SALA & 93.71\scalebox{0.7}{$\!\pm\!$0.23} & 94.25\scalebox{0.7}{$\!\pm\!$0.30} & 99.99\scalebox{0.7}{$\!\pm\!$0.01} & 0.97\scalebox{0.7}{$\!\pm\!$0.01} & 0.11 & 50.27\scalebox{0.7}{$\!\pm\!$0.04} & 52.34\scalebox{0.7}{$\!\pm\!$0.35} & 53.22\scalebox{0.7}{$\!\pm\!$3.21} \\
SALA-SMSE & 93.48\scalebox{0.7}{$\!\pm\!$0.20} & 93.49\scalebox{0.7}{$\!\pm\!$0.24} & 99.98\scalebox{0.7}{$\!\pm\!$0.02} & 0.97\scalebox{0.7}{$\!\pm\!$0.01} & 0.24 & 52.27\scalebox{0.7}{$\!\pm\!$0.13} & 59.96\scalebox{0.7}{$\!\pm\!$0.22} & 63.90\scalebox{0.7}{$\!\pm\!$3.59} \\
SALA w/ $D_\mathrm{f}$ only & 93.62\scalebox{0.7}{$\!\pm\!$0.18} & 93.31\scalebox{0.7}{$\!\pm\!$0.27} & 100.00\scalebox{0.7}{$\!\pm\!$0.00} & 0.96\scalebox{0.7}{$\!\pm\!$0.01} & 0.24 & 52.44\scalebox{0.7}{$\!\pm\!$0.15} & 59.71\scalebox{0.7}{$\!\pm\!$0.25} & 62.84\scalebox{0.7}{$\!\pm\!$3.21} \\
\bottomrule

\end{tabular*}
\end{table}

\Needspace{4\baselineskip}
\subsection{SALA on Class-wise Unlearning}
\label{sec:ex_classwise}

Main experiments focus on sample-wise unlearning (classification) and sequence unlearning (generation). This subsection extends SALA to \emph{class-wise unlearning} to test whether per-sample distributional targets remain effective when the forget set is structured and highly correlated. Class-wise unlearning is evaluated on the classification benchmarks by selecting one target class as the forget set $D_\mathrm{f}$ and treating all remaining classes as the retain set $D_\mathrm{r}$. SALA is applied on top of iterative baselines in the same way as in Section~\ref{sec:baseline_comparison}: forget-class samples are driven toward their $Q_{\mathrm{out}}(\cdot\mid x)$ targets, and retain-class samples are driven toward $Q_{\mathrm{in}}(\cdot\mid x)$ targets, with the same InDist-guided stopping rule. For class-wise unlearning, exact retraining yields FA = 0.00\%, making the original ToW undefined if FA is included. We therefore compute ToW using only TA and RA, and report FA separately as the forgetting measure on the target class. 

\begin{table}[!htb]
	\centering
	\scriptsize
	\setlength{\tabcolsep}{3.65pt}
	\caption{Comprehensive comparison of approximate unlearning methods in class-wise unlearning scenario.}
	\label{tab:sala_class_wise}
	\begin{tabular}{lccccccccc}
		\toprule
		\textbf{Method} & 
		\textbf{TA}(\%)$\uparrow$ &
		\textbf{FA}(\%)$\uparrow$ &
		\textbf{RA}(\%)$\uparrow$ &
		\textbf{ToW}$\uparrow$ &
		\textbf{DNO}$\downarrow$ &
		\textbf{MIA efficacy}(\%)$\downarrow$ &
		\textbf{RULI}(\%)$\downarrow$ &
		\textbf{Time}(s)$\downarrow$ \\ \hline
		Retrain&91.87\scalebox{0.7}{$\!\pm\!2.09$}&
		0.00\scalebox{0.7}{$\!\pm\!0.00$}&
		100.00\scalebox{0.7}{$\!\pm\!0.00$}&
		1.00\scalebox{0.7}{$\!\pm\!0.00$}&
		0.03&
		50.62\scalebox{0.7}{$\!\pm\!0.29$}&
		50.15\scalebox{0.7}{$\!\pm\!0.08$}&
		1627.24\scalebox{0.7}{$\!\pm\!12.10$}\\ \hline
		Salun&95.79\scalebox{0.7}{$\!\pm\!1.02$}&
		0.02\scalebox{0.7}{$\!\pm\!0.09$}&
		99.78\scalebox{0.7}{$\!\pm\!0.13$}&
		0.96\scalebox{0.7}{$\!\pm\!0.01$}&
		0.45&
		51.15\scalebox{0.7}{$\!\pm\!0.09$}&
		77.90\scalebox{0.7}{$\!\pm\!2.89$}&
		\textbf{23.43}\scalebox{0.7}{$\!\pm\!0.20$}\\
		SCRUB&94.43\scalebox{0.7}{$\!\pm\!1.15$}&
		0.00\scalebox{0.7}{$\!\pm\!0.00$}&
		97.42\scalebox{0.7}{$\!\pm\!0.16$}&
		0.95\scalebox{0.7}{$\!\pm\!0.01$}&
		0.39&
		51.36\scalebox{0.7}{$\!\pm\!0.05$}&
		69.49\scalebox{0.7}{$\!\pm\!1.14$}&
		73.39\scalebox{0.7}{$\!\pm\!7.47$} \\
		$\ell_1$ sparsity&90.57\scalebox{0.7}{$\!\pm\!1.69$}&
		0.00\scalebox{0.7}{$\!\pm\!0.00$}&
		99.92\scalebox{0.7}{$\!\pm\!0.04$}&
		\textbf{0.99}\scalebox{0.7}{$\!\pm\!0.00$}&
		0.27&
		50.35\scalebox{0.7}{$\!\pm\!0.04$}&
		66.45\scalebox{0.7}{$\!\pm\!2.12$}&
		92.92\scalebox{0.7}{$\!\pm\!4.21$}\\  \hline
		SALA + Salun&91.31\scalebox{0.7}{$\!\pm\!1.72$}&
		0.00\scalebox{0.7}{$\!\pm\!0.00$}&
		99.98\scalebox{0.7}{$\!\pm\!0.01$}&
		\textbf{0.99}\scalebox{0.7}{$\!\pm\!0.00$}&
		\textbf{0.06}&
		50.41\scalebox{0.7}{$\!\pm\!0.14$}&
		53.25\scalebox{0.7}{$\!\pm\!0.16$}&
		62.47\scalebox{0.7}{$\!\pm\!3.35$} \\
		SALA + SCRUB&91.16\scalebox{0.7}{$\!\pm\!2.15$}&
		0.00\scalebox{0.7}{$\!\pm\!0.00$}&
		99.99\scalebox{0.7}{$\!\pm\!0.00$}&
		\textbf{0.99}\scalebox{0.7}{$\!\pm\!0.01$}&
		0.09&
		50.92\scalebox{0.7}{$\!\pm\!0.25$}&
		52.76\scalebox{0.7}{$\!\pm\!0.24$}&
		64.35\scalebox{0.7}{$\!\pm\!2.15$} \\
		SALA + $\ell_1$ sparsity&90.69\scalebox{0.7}{$\!\pm\!1.72$}&
		0.00\scalebox{0.7}{$\!\pm\!0.00$}&
		100.00\scalebox{0.7}{$\!\pm\!0.00$}&
		\textbf{0.99}\scalebox{0.7}{$\!\pm\!0.00$}&
		0.11&
		\textbf{50.34}\scalebox{0.7}{$\!\pm\!0.10$}&
		\textbf{52.15}\scalebox{0.7}{$\!\pm\!0.16$}&
		74.57\scalebox{0.7}{$\!\pm\!6.41$} \\   \hline
		SALA-SMSE + Salun&91.35\scalebox{0.7}{$\!\pm\!1.43$}&
		0.00\scalebox{0.7}{$\!\pm\!0.00$}&
		100.00\scalebox{0.7}{$\!\pm\!0.00$}&
		\textbf{0.99}\scalebox{0.7}{$\!\pm\!0.00$}&
		0.27&
		50.54\scalebox{0.7}{$\!\pm\!0.16$}&
		60.43\scalebox{0.7}{$\!\pm\!0.65$}&
		69.62\scalebox{0.7}{$\!\pm\!4.47$}  \\
		SALA-SMSE + SCRUB&
		91.15\scalebox{0.7}{$\!\pm\!3.43$}&
		0.00\scalebox{0.7}{$\!\pm\!0.00$}&
		100.00\scalebox{0.7}{$\!\pm\!0.00$}&
		\textbf{0.99}\scalebox{0.7}{$\!\pm\!0.00$}&
		0.25&
		50.35\scalebox{0.7}{$\!\pm\!0.21$}&
		62.24\scalebox{0.7}{$\!\pm\!0.16$}&
		66.48\scalebox{0.7}{$\!\pm\!2.26$} \\
		SALA-SMSE + $\ell_1$ sparsity&
		92.23\scalebox{0.7}{$\!\pm\!1.47$}&
		0.00\scalebox{0.7}{$\!\pm\!0.00$}&
		99.99\scalebox{0.7}{$\!\pm\!0.00$}&
		\textbf{0.99}\scalebox{0.7}{$\!\pm\!0.01$}&
		0.27&
		50.31\scalebox{0.7}{$\!\pm\!0.19$}&
		64.45\scalebox{0.7}{$\!\pm\!1.31$}&
		70.92\scalebox{0.7}{$\!\pm\!4.43$} 
		\\ 
		\bottomrule
	\end{tabular}
\end{table}

\Needspace{4\baselineskip}
\paragraph{Why class-wise is different.}
Compared to random sample-wise deletion, class-wise deletion changes the distribution of the remaining data and induces a larger representation shift: many samples share class-level features, so forgetting one class can affect model outputs on neighboring classes. This makes dataset-level scores alone especially unreliable, and stresses whether SALA can maintain retain behavior while enforcing strong non-membership behavior on a coherent subset.

\paragraph{Results.}
As shown in Table~\ref{tab:sala_class_wise}, SALA remains compatible with class-wise unlearning and preserves the same qualitative benefits: it improves sample-level alignment on the forgotten class (lower DNO and lower RULI accuracy) while keeping ToW competitive.

\setlength{\intextsep}{8pt}
\subsection{Repeated deletion requests}
\label{app:rebuttal-sequential}
To evaluate unlearning quality and cumulative cost when target distributions are reused across deletion requests, we compare Retrain, SALA, and SALA-SMSE over eight sequential requests on CIFAR-10. Each request removes $1{,}000$ samples, totaling $8{,}000$ deletions. SALA and SALA-SMSE estimate their target distributions once and reuse them across requests; Retrain trains on the remaining retain set after each request.

\paragraph{Cumulative cost.}
After the common initial model training, Table~\ref{tab:rebuttal-sequential-cost} counts target estimation at $K=0$ and adds the time to process each completed request. Shadow-model estimation dominates SALA's cumulative time, which remains above repeated retraining after eight requests. With SMSE, the cumulative time for eight requests is $461.09$ s, compared with $9{,}854.16$ s for retraining.

\paragraph{Unlearning quality.}
Table~\ref{tab:rebuttal-sequential-performance} compares performance after the first and eighth requests. After eight requests, both variants maintain ToW near $0.965$ and retain accuracy above $99.5\%$. SALA achieves lower DNO, while SALA-SMSE maintains comparable ToW with substantially lower cumulative cost.

\begin{table}[!htb]
\centering
\scriptsize
\setlength{\tabcolsep}{1.5pt}
\renewcommand{\arraystretch}{1.05}
\caption{Cumulative time (seconds) over repeated deletion requests. $K=0$ records the target-estimation cost.}
\label{tab:rebuttal-sequential-cost}
\begin{tabular*}{\linewidth}{@{\extracolsep{\fill}}lccccccccc@{}}
\toprule
\textbf{Method} & \textbf{$K=0$} & \textbf{$K=1$} & \textbf{$K=2$} & \textbf{$K=3$} & \textbf{$K=4$} & \textbf{$K=5$} & \textbf{$K=6$} & \textbf{$K=7$} & \textbf{$K=8$} \\
\midrule
Retrain & 0.00 & 1,264.78 & 2,540.88 & 3,757.28 & 4,930.74 & 6,073.32 & 7,185.72 & 8,274.24 & 9,854.16 \\
SALA & 46,211.80 & 46,418.71 & 46,618.79 & 46,819.77 & 47,020.85 & 47,188.40 & 47,369.07 & 47,546.77 & 47,715.27 \\
SALA-SMSE & 49.82 & 110.61 & 166.30 & 222.20 & 281.67 & 329.76 & 376.10 & 420.77 & 461.09 \\
\bottomrule
\end{tabular*}
\end{table}

\begin{table}[!htb]
\centering
\scriptsize
\setlength{\tabcolsep}{3pt}
\renewcommand{\arraystretch}{1.05}
\caption{CIFAR-10 results after the first and eighth deletion requests.}
\label{tab:rebuttal-sequential-performance}
\begin{tabular*}{\linewidth}{@{\extracolsep{\fill}}llcccccc@{}}
\toprule
\textbf{$K$} & \textbf{Method} & \textbf{TA(\%)} & \textbf{FA(\%)} & \textbf{RA(\%)} & \textbf{ToW} & \textbf{DNO} & \textbf{RULI(\%)} \\
\midrule
\multirow{3}{*}{1} & Retrain & 94.57\scalebox{0.7}{$\!\pm\!$0.19} & 94.45\scalebox{0.7}{$\!\pm\!$0.72} & 100.000\scalebox{0.7}{$\!\pm\!$0.001} & 1.0000\scalebox{0.7}{$\!\pm\!$0.0000} & 0.0000\scalebox{0.7}{$\!\pm\!$0.0000} & 50.00\scalebox{0.7}{$\!\pm\!$0.00} \\
 & SALA & 94.44\scalebox{0.7}{$\!\pm\!$0.26} & 94.01\scalebox{0.7}{$\!\pm\!$0.98} & 99.88\scalebox{0.7}{$\!\pm\!$0.02} & 0.9871\scalebox{0.7}{$\!\pm\!$0.0057} & 0.1725\scalebox{0.7}{$\!\pm\!$0.0641} & 56.72\scalebox{0.7}{$\!\pm\!$4.13} \\
 & SALA-SMSE & 94.34\scalebox{0.7}{$\!\pm\!$0.29} & 94.54\scalebox{0.7}{$\!\pm\!$0.39} & 99.96\scalebox{0.7}{$\!\pm\!$0.01} & 0.9917\scalebox{0.7}{$\!\pm\!$0.0031} & 0.3469\scalebox{0.7}{$\!\pm\!$0.0850} & 59.44\scalebox{0.7}{$\!\pm\!$7.62} \\
\midrule
\multirow{3}{*}{8} & Retrain & 93.70\scalebox{0.7}{$\!\pm\!$0.25} & 93.77\scalebox{0.7}{$\!\pm\!$0.37} & 100.00\scalebox{0.7}{$\!\pm\!$0.00} & 1.0000\scalebox{0.7}{$\!\pm\!$0.0000} & 0.0000\scalebox{0.7}{$\!\pm\!$0.0000} & 50.00\scalebox{0.7}{$\!\pm\!$0.00} \\
 & SALA & 92.57\scalebox{0.7}{$\!\pm\!$1.51} & 94.73\scalebox{0.7}{$\!\pm\!$1.63} & 99.55\scalebox{0.7}{$\!\pm\!$0.43} & 0.9643\scalebox{0.7}{$\!\pm\!$0.0144} & 0.1486\scalebox{0.7}{$\!\pm\!$0.0791} & 56.93\scalebox{0.7}{$\!\pm\!$1.42} \\
 & SALA-SMSE & 94.33\scalebox{0.7}{$\!\pm\!$0.20} & 96.40\scalebox{0.7}{$\!\pm\!$0.21} & 99.97\scalebox{0.7}{$\!\pm\!$0.01} & 0.9650\scalebox{0.7}{$\!\pm\!$0.0052} & 0.2854\scalebox{0.7}{$\!\pm\!$0.0975} & 57.38\scalebox{0.7}{$\!\pm\!$0.99} \\
\bottomrule
\end{tabular*}
\end{table}

%% file: checklist.tex
\section*{NeurIPS Paper Checklist}
\begin{enumerate}

\Needspace{10\baselineskip}
\item {\bf Claims}
    \item[] Question: Do the main claims made in the abstract and introduction accurately reflect the paper's contributions and scope?
    \item[] Answer: \answerYes{} %
    \item[] Justification: The abstract and introduction explicitly state the paper's main claims: SALA is a sample-grained approximate unlearning framework, DNO measures per-sample gap-to-retrain, SALA provides a computable upper bound on this gap, and SMSE estimates the required per-sample target distributions.

\Needspace{10\baselineskip}
\item {\bf Limitations}
    \item[] Question: Does the paper discuss the limitations of the work performed by the authors?
    \item[] Answer: \answerYes{} %
    \item[] Justification: We discuss limitations in Appendix~\ref{app:limitation}. 

\Needspace{10\baselineskip}
\item {\bf Theory assumptions and proofs}
    \item[] Question: For each theoretical result, does the paper provide the full set of assumptions and a complete (and correct) proof?
    \item[] Answer: \answerYes{} %
    \item[] Justification:  The paper states the assumptions for each theoretical result, including the per-sample Gaussian score modeling, the InDist constraint, the scale feasibility condition, the retain-consistency condition for the forget-set-only variant, and the estimation-error decomposition. Theorem 3.1, Corollary 3.2, and Corollary 4.1 are accompanied by complete proofs or derivations in Appendices~\ref{app:dno_bound_proof},~\ref{app:dno_bound_proof_single}  and~\ref{app:smse_error_bound}.

    \item {\bf Experimental result reproducibility}
    \item[] Question: Does the paper fully disclose all the information needed to reproduce the main experimental results of the paper to the extent that it affects the main claims and/or conclusions of the paper (regardless of whether the code and data are provided or not)?
    \item[] Answer: \answerYes{}
    \item[] Justification: The paper discloses the datasets, models, unlearning scenarios, metrics, number of unlearning tasks, baseline coverage, and SALA/SMSE variants used to obtain the main experimental results. A complete code repository is provided with the implementation, configurations, and reproduction scripts needed to reproduce the main experiments.

\Needspace{10\baselineskip}
\item {\bf Open access to data and code}
    \item[] Question: Does the paper provide open access to the data and code, with sufficient instructions to faithfully reproduce the main experimental results, as described in supplemental material?
    \item[] Answer: \answerYes{} %
    \item[] Justification: We provide open-source code at the repository linked in Section~\ref{sec:ex}.

\Needspace{10\baselineskip}
\item {\bf Experimental setting/details}
    \item[] Question: Does the paper specify all the training and test details (e.g., data splits, hyperparameters, how they were chosen, type of optimizer) necessary to understand the results?
    \item[] Answer: \answerYes{} %
    \item[] Justification: The main text in Section~\ref{sec:ex_setup} describes the experimental setup, including datasets, models, unlearning scenarios, metrics, and evaluated variants, while Appendix~\ref{app:setup} provides detailed configurations.

\Needspace{10\baselineskip}
\item {\bf Experiment statistical significance}
    \item[] Question: Does the paper report error bars suitably and correctly defined or other appropriate information about the statistical significance of the experiments?
    \item[] Answer: \answerYes{} %
    \item[] Justification: We report std values in the experimental tables, and Appendix~\ref{app:setup} states that these values denote one standard deviation over independent unlearning tasks and random seeds. The same paragraph also specifies the number of independent tasks used for DNO computation on CIFAR-10, ImageNet-1000, and WikiText-103. 

\Needspace{10\baselineskip}
\item {\bf Experiments compute resources}
    \item[] Question: For each experiment, does the paper provide sufficient information on the computer resources (type of compute workers, memory, time of execution) needed to reproduce the experiments?
    \item[] Answer: \answerYes{} %
    \item[] Justification: Appendix~\ref{app:setup}, Implementation, reports the compute resources used for the experiments, including the GPU configurations for CIFAR-10, ImageNet-1000, and WikiText-103. It also states that all reported runtimes are measured under these hardware settings.
    
\Needspace{10\baselineskip}
\item {\bf Code of ethics}
    \item[] Question: Does the research conducted in the paper conform, in every respect, with the NeurIPS Code of Ethics \url{https://neurips.cc/public/EthicsGuidelines}?
    \item[] Answer: \answerYes{} %
    \item[] Justification: The research conforms to the NeurIPS Code of Ethics. It uses standard public benchmarks, does not involve human subjects or private user data, and is intended to improve the evaluation and privacy protection of machine unlearning methods.

\Needspace{10\baselineskip}
\item {\bf Broader impacts}
    \item[] Question: Does the paper discuss both potential positive societal impacts and negative societal impacts of the work performed?
    \item[] Answer: \answerYes{} %
    \item[] Justification: We discuss broader impacts  in Appendix~\ref{app:limitation}. 
    
\Needspace{10\baselineskip}
\item {\bf Safeguards}
    \item[] Question: Does the paper describe safeguards that have been put in place for responsible release of data or models that have a high risk for misuse (e.g., pre-trained language models, image generators, or scraped datasets)?
    \item[] Answer:\answerNA{} %
    \item[] Justification: \answerNA{}

\Needspace{10\baselineskip}
\item {\bf Licenses for existing assets}
    \item[] Question: Are the creators or original owners of assets (e.g., code, data, models), used in the paper, properly credited and are the license and terms of use explicitly mentioned and properly respected?
    \item[] Answer: \answerYes{} %
    \item[] Justification: We use standard existing assets, including public benchmark datasets, model architectures, and baseline methods, and properly credit their original creators through citations in the paper. The released repository further documents the assets used in the experiments and follows their corresponding terms of use.

\Needspace{10\baselineskip}
\item {\bf New assets}
    \item[] Question: Are new assets introduced in the paper well documented and is the documentation provided alongside the assets?
    \item[] Answer: \answerYes{} %
    \item[] Justification: The paper introduces new code assets for SALA and SMSE.

\Needspace{10\baselineskip}
\item {\bf Crowdsourcing and research with human subjects}
    \item[] Question: For crowdsourcing experiments and research with human subjects, does the paper include the full text of instructions given to participants and screenshots, if applicable, as well as details about compensation (if any)? 
    \item[] Answer: \answerNA{}%
    \item[] Justification: \answerNA{}

\Needspace{10\baselineskip}
\item {\bf Institutional review board (IRB) approvals or equivalent for research with human subjects}
    \item[] Question: Does the paper describe potential risks incurred by study participants, whether such risks were disclosed to the subjects, and whether Institutional Review Board (IRB) approvals (or an equivalent approval/review based on the requirements of your country or institution) were obtained?
    \item[] Answer: \answerNA{} %
    \item[] Justification: \answerNA{}

\Needspace{10\baselineskip}
\item {\bf Declaration of LLM usage}
    \item[] Question: Does the paper describe the usage of LLMs if it is an important, original, or non-standard component of the core methods in this research? Note that if the LLM is used only for writing, editing, or formatting purposes and does \emph{not} impact the core methodology, scientific rigor, or originality of the research, declaration is not required.
    \item[] Answer: \answerNA{} %
    \item[] Justification: \answerNA{}

\end{enumerate}

%% file: neurips_2026.bbl
\begin{thebibliography}{40}
\providecommand{\natexlab}[1]{#1}
\providecommand{\url}[1]{\texttt{#1}}
\expandafter\ifx\csname urlstyle\endcsname\relax
  \providecommand{\doi}[1]{doi: #1}\else
  \providecommand{\doi}{doi: \begingroup \urlstyle{rm}\Url}\fi

\bibitem[Block et~al.(2025)Block, Mokhtari, and Shakkottai]{jacob2025machine}
Jacob~L. Block, Aryan Mokhtari, and Sanjay Shakkottai.
\newblock Machine unlearning under overparameterization.
\newblock \emph{CoRR}, abs/2505.22601, 2025.
\newblock \doi{10.48550/ARXIV.2505.22601}.
\newblock URL \url{https://doi.org/10.48550/arXiv.2505.22601}.

\bibitem[Bourtoule et~al.(2021)Bourtoule, Chandrasekaran, Choquette-Choo, Jia,
  Travers, Zhang, Lie, and Papernot]{bourtoule2021machine}
Lucas Bourtoule, Varun Chandrasekaran, Christopher~A Choquette-Choo, Hengrui
  Jia, Adelin Travers, Baiwu Zhang, David Lie, and Nicolas Papernot.
\newblock Machine unlearning.
\newblock In \emph{2021 IEEE symposium on security and privacy (SP)}, pages
  141--159. IEEE, 2021.

\bibitem[Brophy et~al.(2023)Brophy, Hammoudeh, and Lowd]{brophy2023adapting}
Jonathan Brophy, Zayd Hammoudeh, and Daniel Lowd.
\newblock Adapting and evaluating influence-estimation methods for
  gradient-boosted decision trees.
\newblock \emph{Journal of Machine Learning Research}, 24\penalty0
  (154):\penalty0 1--48, 2023.

\bibitem[Cadet et~al.(2024)Cadet, Borovykh, Malekzadeh, Ahmadi-Abhari, and
  Haddadi]{cadet2024deep}
Xavier~F Cadet, Anastasia Borovykh, Mohammad Malekzadeh, Sara Ahmadi-Abhari,
  and Hamed Haddadi.
\newblock Deep unlearn: Benchmarking machine unlearning.
\newblock \emph{arXiv preprint arXiv:2410.01276}, 2024.

\bibitem[Cao and Yang(2015)]{cao2015forget}
Yinzhi Cao and Junfeng Yang.
\newblock Towards making systems forget with machine unlearning.
\newblock In \emph{2015 IEEE symposium on security and privacy}, pages
  463--480. IEEE, 2015.

\bibitem[Carlini et~al.(2022)Carlini, Chien, Nasr, Song, Terzis, and
  Tramer]{carlini2022membership}
Nicholas Carlini, Steve Chien, Milad Nasr, Shuang Song, Andreas Terzis, and
  Florian Tramer.
\newblock Membership inference attacks from first principles.
\newblock In \emph{2022 IEEE symposium on security and privacy (SP)}, pages
  1897--1914. IEEE, 2022.

\bibitem[Chien et~al.(2024)Chien, Wang, Chen, and Li]{chien2024certified}
Eli Chien, Haoyu Wang, Ziang Chen, and Pan Li.
\newblock Certified machine unlearning via noisy stochastic gradient descent.
\newblock \emph{Advances in Neural Information Processing Systems},
  37:\penalty0 38852--38887, 2024.

\bibitem[Chundawat et~al.(2023)Chundawat, Tarun, Mandal, and
  Kankanhalli]{6can23chundawat}
Vikram~S. Chundawat, Ayush~K. Tarun, Murari Mandal, and Mohan~S. Kankanhalli.
\newblock Can bad teaching induce forgetting? unlearning in deep networks using
  an incompetent teacher.
\newblock In Brian Williams, Yiling Chen, and Jennifer Neville, editors,
  \emph{Thirty-Seventh {AAAI} Conference on Artificial Intelligence, {AAAI}
  2023, Thirty-Fifth Conference on Innovative Applications of Artificial
  Intelligence, {IAAI} 2023, Thirteenth Symposium on Educational Advances in
  Artificial Intelligence, {EAAI} 2023, Washington, DC, USA, February 7-14,
  2023}, pages 7210--7217. {AAAI} Press, 2023.
\newblock \doi{10.1609/AAAI.V37I6.25879}.

\bibitem[Daxberger et~al.(2021)Daxberger, Kristiadi, Immer, Eschenhagen, Bauer,
  and Hennig]{daxberger2021laplace}
Erik Daxberger, Agustinus Kristiadi, Alexander Immer, Runa Eschenhagen,
  Matthias Bauer, and Philipp Hennig.
\newblock Laplace redux -- effortless bayesian deep learning.
\newblock In \emph{Advances in Neural Information Processing Systems},
  volume~34, 2021.

\bibitem[Entesari et~al.(2025)Entesari, Hatami, Khaziev, Ramakrishna, and
  Fazlyab]{entesari2025constrained}
Taha Entesari, Arman Hatami, Rinat Khaziev, Anil Ramakrishna, and Mahyar
  Fazlyab.
\newblock Constrained entropic unlearning: A primal-dual framework for large
  language models.
\newblock \emph{arXiv preprint arXiv:2506.05314}, 2025.

\bibitem[{European Parliament and Council of the European
  Union}(2016)]{gdpr2016}
{European Parliament and Council of the European Union}.
\newblock Regulation (eu) 2016/679 of the european parliament and of the
  council of 27 april 2016.
\newblock Official Journal of the European Union, 2016.

\bibitem[Fan et~al.(2024)Fan, Liu, Zhang, Wei, Wong, and Liu]{fan2024salun}
Chongyu Fan, Jiancheng Liu, Yihua Zhang, Dennis Wei, Eric Wong, and Sijia Liu.
\newblock Salun: Empowering machine unlearning via gradient-based weight
  saliency in both image classification and generation.
\newblock In \emph{International Conference on Learning Representations}, 2024.

\bibitem[Feldman and Zhang(2020)]{feldman2020neural}
Vitaly Feldman and Chiyuan Zhang.
\newblock What neural networks memorize and why: Discovering the long tail via
  influence estimation.
\newblock \emph{Advances in neural information processing systems},
  33:\penalty0 2881--2891, 2020.

\bibitem[Ginart et~al.(2019)Ginart, Guan, Valiant, and Zou]{ginart2019making}
Antonio Ginart, Melody Guan, Gregory Valiant, and James~Y Zou.
\newblock Making ai forget you: Data deletion in machine learning.
\newblock \emph{Advances in neural information processing systems}, 32, 2019.

\bibitem[Golatkar et~al.(2020)Golatkar, Achille, and
  Soatto]{golatkar2020eternal}
Aditya Golatkar, Alessandro Achille, and Stefano Soatto.
\newblock Eternal sunshine of the spotless net: Selective forgetting in deep
  networks.
\newblock In \emph{2020 {IEEE/CVF} Conference on Computer Vision and Pattern
  Recognition, {CVPR} 2020, Seattle, WA, USA, June 13-19, 2020}, pages
  9301--9309. Computer Vision Foundation / {IEEE}, 2020.
\newblock \doi{10.1109/CVPR42600.2020.00932}.

\bibitem[Graves et~al.(2021)Graves, Nagisetty, and Ganesh]{graves2021amnesiac}
Laura Graves, Vineel Nagisetty, and Vijay Ganesh.
\newblock Amnesiac machine learning.
\newblock In \emph{Proceedings of the AAAI Conference on Artificial
  Intelligence}, volume~35, pages 11516--11524, 2021.

\bibitem[Guo et~al.(2020)Guo, Goldstein, Hannun, and Van
  Der~Maaten]{guo2019certified}
Chuan Guo, Tom Goldstein, Awni Hannun, and Laurens Van Der~Maaten.
\newblock Certified data removal from machine learning models.
\newblock In \emph{Proceedings of the 37th International Conference on Machine
  Learning}, volume 119 of \emph{Proceedings of Machine Learning Research},
  pages 3832--3842, 2020.

\bibitem[Hayes et~al.(2025)Hayes, Shumailov, Triantafillou, Khalifa, and
  Papernot]{hayes2025inexact}
Jamie Hayes, Ilia Shumailov, Eleni Triantafillou, Amr Khalifa, and Nicolas
  Papernot.
\newblock Inexact unlearning needs more careful evaluations to avoid a false
  sense of privacy.
\newblock In \emph{2025 IEEE Conference on Secure and Trustworthy Machine
  Learning (SaTML)}, pages 497--519. IEEE, 2025.

\bibitem[He et~al.(2016)He, Zhang, Ren, and Sun]{he16deep}
Kaiming He, Xiangyu Zhang, Shaoqing Ren, and Jian Sun.
\newblock Deep residual learning for image recognition.
\newblock In \emph{2016 {IEEE} Conference on Computer Vision and Pattern
  Recognition, {CVPR} 2016, Las Vegas, NV, USA, June 27-30, 2016}, pages
  770--778. {IEEE} Computer Society, 2016.
\newblock \doi{10.1109/CVPR.2016.90}.

\bibitem[Huang et~al.(2024)Huang, Cheng, Zheng, Wang, He, Li, and
  Huang]{huang2024unified}
Zhehao Huang, Xinwen Cheng, JingHao Zheng, Haoran Wang, Zhengbao He, Tao Li,
  and Xiaolin Huang.
\newblock Unified gradient-based machine unlearning with remain geometry
  enhancement.
\newblock In Amir Globersons, Lester Mackey, Danielle Belgrave, Angela Fan,
  Ulrich Paquet, Jakub~M. Tomczak, and Cheng Zhang, editors, \emph{Advances in
  Neural Information Processing Systems 38: Annual Conference on Neural
  Information Processing Systems 2024, NeurIPS 2024, Vancouver, BC, Canada,
  December 10 - 15, 2024}, 2024.

\bibitem[Jia et~al.(2023)Jia, Liu, Ram, Yao, Liu, Liu, Sharma, and
  Liu]{1modelsparisty23jia}
Jinghan Jia, Jiancheng Liu, Parikshit Ram, Yuguang Yao, Gaowen Liu, Yang Liu,
  Pranay Sharma, and Sijia Liu.
\newblock Model sparsity can simplify machine unlearning.
\newblock In Alice Oh, Tristan Naumann, Amir Globerson, Kate Saenko, Moritz
  Hardt, and Sergey Levine, editors, \emph{Advances in Neural Information
  Processing Systems 36: Annual Conference on Neural Information Processing
  Systems 2023, NeurIPS 2023, New Orleans, LA, USA, December 10 - 16, 2023},
  2023.

\bibitem[Kailath(1967)]{kailath1967divergence}
Thomas Kailath.
\newblock The divergence and bhattacharyya distance measures in signal
  selection.
\newblock \emph{IEEE transactions on communication technology}, 15\penalty0
  (1):\penalty0 52--60, 1967.

\bibitem[Kairouz et~al.(2021)Kairouz, McMahan, Avent, Bellet, Bennis, Bhagoji,
  Bonawitz, Charles, Cormode, Cummings, et~al.]{kairouz2021advances}
Peter Kairouz, H~Brendan McMahan, Brendan Avent, Aur{\'e}lien Bellet, Mehdi
  Bennis, Arjun~Nitin Bhagoji, Kallista Bonawitz, Zachary Charles, Graham
  Cormode, Rachel Cummings, et~al.
\newblock Advances and open problems in federated learning.
\newblock \emph{Foundations and trends in machine learning}, 14\penalty0
  (1--2):\penalty0 1--210, 2021.

\bibitem[Koloskova et~al.(2025)Koloskova, Allouah, Jha, Guerraoui, and
  Koyejo]{koloskova2025certified}
Anastasia Koloskova, Youssef Allouah, Animesh Jha, Rachid Guerraoui, and Sanmi
  Koyejo.
\newblock Certified unlearning for neural networks.
\newblock \emph{arXiv preprint arXiv:2506.06985}, 2025.

\bibitem[Krizhevsky et~al.(2009)Krizhevsky, Hinton,
  et~al.]{krizhevsky2009learning}
Alex Krizhevsky, Geoffrey Hinton, et~al.
\newblock Learning multiple layers of features from tiny images.
\newblock Technical report, Citeseer, 2009.

\bibitem[Kurmanji et~al.(2023)Kurmanji, Triantafillou, Hayes, and
  Triantafillou]{kurmanji2023towards}
Meghdad Kurmanji, Peter Triantafillou, Jamie Hayes, and Eleni Triantafillou.
\newblock Towards unbounded machine unlearning.
\newblock \emph{Advances in neural information processing systems},
  36:\penalty0 1957--1987, 2023.

\bibitem[Li et~al.(2025)Li, Wei, and Thuraisingham]{li2025mubox}
Xiang Li, Wenqi Wei, and Bhavani Thuraisingham.
\newblock Mubox: A critical evaluation framework of deep machine unlearning
  [systematization of knowledge paper].
\newblock In \emph{Proceedings of the 30th ACM Symposium on Access Control
  Models and Technologies}, pages 175--188, 2025.

\bibitem[Merity et~al.(2016)Merity, Xiong, Bradbury, and
  Socher]{merity2016pointer}
Stephen Merity, Caiming Xiong, James Bradbury, and Richard Socher.
\newblock Pointer sentinel mixture models.
\newblock \emph{arXiv preprint arXiv:1609.07843}, 2016.

\bibitem[Mu and Klabjan(2024)]{mu2024rewind}
Siqiao Mu and Diego Klabjan.
\newblock Rewind-to-delete: Certified machine unlearning for nonconvex
  functions.
\newblock \emph{arXiv preprint arXiv:2409.09778}, 2024.

\bibitem[Naderloui et~al.(2025)Naderloui, Yan, Wang, Fu, Wang, Liu, and
  Hong]{naderloui2025rectifying}
Nima Naderloui, Shenao Yan, Binghui Wang, Jie Fu, Wendy~Hui Wang, Weiran Liu,
  and Yuan Hong.
\newblock Rectifying privacy and efficacy measurements in machine unlearning: A
  new inference attack perspective.
\newblock \emph{arXiv preprint arXiv:2506.13009}, 2025.

\bibitem[Radford et~al.(2019)Radford, Wu, Child, Luan, Amodei, Sutskever,
  et~al.]{radford2019language}
Alec Radford, Jeffrey Wu, Rewon Child, David Luan, Dario Amodei, Ilya
  Sutskever, et~al.
\newblock Language models are unsupervised multitask learners.
\newblock \emph{OpenAI blog}, 1\penalty0 (8):\penalty0 9, 2019.

\bibitem[Russakovsky et~al.(2015)Russakovsky, Deng, Su, Krause, Satheesh, Ma,
  Huang, Karpathy, Khosla, Bernstein, et~al.]{russakovsky2015imagenet}
Olga Russakovsky, Jia Deng, Hao Su, Jonathan Krause, Sanjeev Satheesh, Sean Ma,
  Zhiheng Huang, Andrej Karpathy, Aditya Khosla, Michael Bernstein, et~al.
\newblock Imagenet large scale visual recognition challenge.
\newblock \emph{International journal of computer vision}, 115\penalty0
  (3):\penalty0 211--252, 2015.

\bibitem[Sendera et~al.(2025)Sendera, Struski, Ksiazek, Musiol, Tabor, and
  Rymarczyk]{marcin2025semu}
Marcin Sendera, Lukasz Struski, Kamil Ksiazek, Kryspin Musiol, Jacek Tabor, and
  Dawid~Damian Rymarczyk.
\newblock {SEMU:} singular value decomposition for efficient machine
  unlearning.
\newblock In Aarti Singh, Maryam Fazel, Daniel Hsu, Simon Lacoste{-}Julien,
  Felix Berkenkamp, Tegan Maharaj, Kiri Wagstaff, and Jerry Zhu, editors,
  \emph{Forty-second International Conference on Machine Learning, {ICML} 2025,
  Vancouver, BC, Canada, July 13-19, 2025}, Proceedings of Machine Learning
  Research. {PMLR} / OpenReview.net, 2025.
\newblock URL \url{https://proceedings.mlr.press/v267/sendera25a.html}.

\bibitem[Sula et~al.(2024)Sula, Kumar, Hou, Wang, and Tourani]{sula2024silver}
Nexhi Sula, Abhinav Kumar, Jie Hou, Han Wang, and Reza Tourani.
\newblock Silver linings in the shadows: Harnessing membership inference for
  machine unlearning.
\newblock \emph{arXiv preprint arXiv:2407.00866}, 2024.

\bibitem[Tarun et~al.(2023)Tarun, Chundawat, Mandal, and
  Kankanhalli]{tarun2023fast}
Ayush~K Tarun, Vikram~S Chundawat, Murari Mandal, and Mohan Kankanhalli.
\newblock Fast yet effective machine unlearning.
\newblock \emph{IEEE Transactions on Neural Networks and Learning Systems},
  35\penalty0 (9):\penalty0 13046--13055, 2023.

\bibitem[Thudi et~al.(2022)Thudi, Deza, Chandrasekaran, and
  Papernot]{thudi2022unrolling}
Anvith Thudi, Gabriel Deza, Varun Chandrasekaran, and Nicolas Papernot.
\newblock Unrolling sgd: Understanding factors influencing machine unlearning.
\newblock In \emph{2022 IEEE 7th European Symposium on Security and Privacy
  (EuroS\&P)}, pages 303--319. IEEE, 2022.

\bibitem[Wu and Harandi(2024)]{wu2024scissorhands}
Jing Wu and Mehrtash Harandi.
\newblock Scissorhands: Scrub data influence via connection sensitivity in
  networks.
\newblock In \emph{European Conference on Computer Vision}, pages 367--384.
  Springer, 2024.

\bibitem[Yuan et~al.(2024)Yuan, Pang, Du, Chen, Zhang, and Lin]{yuan2024closer}
Xiaojian Yuan, Tianyu Pang, Chao Du, Kejiang Chen, Weiming Zhang, and Min Lin.
\newblock A closer look at machine unlearning for large language models.
\newblock \emph{arXiv preprint arXiv:2410.08109}, 2024.

\bibitem[Zhang et~al.(2024)Zhang, Lin, Bai, and Mei]{zhang2024negative}
Ruiqi Zhang, Licong Lin, Yu~Bai, and Song Mei.
\newblock Negative preference optimization: From catastrophic collapse to
  effective unlearning.
\newblock \emph{arXiv preprint arXiv:2404.05868}, 2024.

\bibitem[Zhao et~al.(2024)Zhao, Kurmanji, Barbulescu, Triantafillou, and
  Triantafillou]{zhao2024what}
Kairan Zhao, Meghdad Kurmanji, George{-}Octavian Barbulescu, Eleni
  Triantafillou, and Peter Triantafillou.
\newblock What makes unlearning hard and what to do about it.
\newblock In Amir Globersons, Lester Mackey, Danielle Belgrave, Angela Fan,
  Ulrich Paquet, Jakub~M. Tomczak, and Cheng Zhang, editors, \emph{Advances in
  Neural Information Processing Systems 38: Annual Conference on Neural
  Information Processing Systems 2024, NeurIPS 2024, Vancouver, BC, Canada,
  December 10 - 15, 2024}, 2024.

\end{thebibliography}
